\documentclass[preprint,12pt]{elsarticle}

\journal{Energy and AI}

\usepackage{graphicx}
\usepackage{amssymb}
\usepackage{xcolor}
\usepackage[linguistics]{forest}
\usepackage{adjustbox}
\usepackage{booktabs}
\usepackage{tabularx}
\usepackage[table]{xcolor}
\usepackage{float} 
\usepackage{tikz}
\usetikzlibrary{mindmap,trees}
\usetikzlibrary{mindmap,calc}
\usetikzlibrary{mindmap, backgrounds}

\usepackage{longtable}
\usepackage{amsmath}
\usepackage[hyphens]{url}

\usepackage{lmodern} 

\usepackage{multirow}
\usepackage{siunitx}
\usepackage{adjustbox}
\usepackage{amsthm} 
\usepackage{pgfplots}
\usepgfplotslibrary{fillbetween}
\usepgfplotslibrary{colorbrewer}
\pgfplotsset{
  compat=1.17,
  cycle list/Set1-8,
  grid style={gray!30},
  every axis/.append style={
    tick label style={font=\small},
    label style={font=\normalsize},
  }
}

\usepackage{pgf}
\usepackage{lmodern}

\everymath=\expandafter{\the\everymath\displaystyle}
\IfFileExists{scrextend.sty}{
\usepackage[fontsize=10.000000pt]{scrextend}
}{
\renewcommand{\normalsize}{\fontsize{10.000000}{12.000000}\selectfont}
\normalsize
}
  
\makeatletter\@ifpackageloaded{underscore}{}{\usepackage[strings]{underscore}}\makeatother

\usepackage{lineno}

\usepackage[utf8]{inputenc} 
\usepackage[T1]{fontenc}    
\usepackage{hyperref}       
\usepackage{url}            
\usepackage{booktabs}       
\usepackage{amsfonts}       
\usepackage{nicefrac}       
\usepackage{microtype}      
\usepackage{lipsum}		
\usepackage{graphicx}
\usepackage{natbib}
\usepackage{doi}

\usepackage[hyphens]{url}

\hypersetup{
    breaklinks=true,
    urlcolor=blue,
    pdfborder={0 0 0}
}

\usepackage{multirow}
\usepackage{siunitx}
\usepackage{adjustbox}

\usetikzlibrary{positioning}
\usetikzlibrary{decorations.pathreplacing, calligraphy}
\usetikzlibrary{patterns, patterns.meta} 

\usepackage{fontawesome5}
\usepackage{todonotes}
\usepackage{marginnote}
\usepackage{xcolor}
\usepackage{ulem}

\usepackage[headsepline]{scrlayer-scrpage}
\usepackage{xcolor}
\usepackage{everypage}  
\def\tempheadnote{}

\usepackage{rotating} 

\usepackage{xfp}

\newcommand{\val}[1]{%
  \ifnum\fpeval{#1 > 1000000}=1
    $>$\num{1e6}%
  \else
    \num{#1}%
  \fi
}

{\renewcommand{\arraystretch}{1.3}
}
\begin{document}
\hypersetup{breaklinks=true,urlcolor=blue,pdfborder={0 0 0}}

\clearpage
\makeatletter
\def\thefnote{\ifcase\c@fnote\or$\dagger$\else\arabic{fnote}\fi}
\makeatother

\begin{frontmatter}



\title{FETS Benchmark: \textbf{F}oundation Models Enable Scalable and Generalizable \textbf{E}nergy \textbf{T}ime \textbf{S}eries Forecasting}




\author[1,2,3]{Marco Obermeier\corref{corauthor}}
\ead{marco.obermeier@hswt.de}

\author[1]{Marco Pruckner}
\ead{marco.pruckner@uni-wuerzburg.de}

\author[2]{Florian Haselbeck\fnref{equal}} 
\ead{florian.haselbeck@hswt.de}

\author[3]{Andreas Zeiselmair\fnref{equal}} 
\ead{andreas.zeiselmair@hswt.de}

\affiliation[1]{
  organization={Julius-Maximilians-Universität Würzburg, 
Modeling and Simulation Lab},
  addressline={Am Hubland},
  city={Würzburg},
  postcode={97074},
  country={Germany}}


\affiliation[2]{
 organization={Weihenstephan-Triesdorf University of Applied Sciences, Smart Farming},
addressline={Am Staudengarten 1},
city={Freising},
postcode={85354},
country={Germany}
}

\affiliation[3]{
  organization={Weihenstephan-Triesdorf University of Applied Sciences, Digital Energy Transition},
  addressline={Hans-Carl-von-Carlowitz-Platz 3},
  city={Freising},
  postcode={85354},
  country={Germany}
  }

\cortext[corauthor]{Corresponding author.}

\fntext[equal]{These authors contributed equally as joint last authors.}

\begin{abstract}
Driven by the transition towards a climate-neutral energy system, accurate energy time series forecasting is critical for planning and operations. Yet, it remains a dataset-specific task, requiring comprehensive training data, limiting scalability, and resulting in high model development and maintenance effort. Recently, foundation models aiming to learn generalizable patterns via extensive pretraining have shown strong performance in multiple prediction tasks. Despite their success and strong potential in energy forecasting, a systematic, use-case-differentiated evaluation is still missing. We address this gap by presenting the \textbf{F}oundation Models in \textbf{E}nergy \textbf{T}ime \textbf{S}eries Forecasting (FETS) benchmark. We (1) provide a structured overview of energy forecasting use cases along three main dimensions, i.e., stakeholders, attributes, and data categories, (2) curate 54 datasets across 9 data categories, guided by typical stakeholder interests, and (3) benchmark foundation models against task-specific machine learning across different forecasting settings. In our benchmark study, covariate-informed zero-shot foundation models perform best in aggregate, with Chronos-2 attaining the lowest overall median NRMSE (0.472), closely followed by TiRex-2 (0.474). Both perform better than XGBoost (0.611) and random forest (0.696), although they were trained task-specifically on the full historic target data. Further analysis reveals a strong correlation between predictive performance and spectral entropy.  Performance saturates beyond a certain context length and improves with aggregation level, e.g., for national load, district heating, and power grid data. Overall, with the lowest median error, limited data requirements, and low inference and hardware demands, foundation models reduce development and maintenance effort, emerging as scalable and generalizable energy forecasting solutions.

\end{abstract}





\begin{keyword}
\small Time series foundation models \sep energy data \sep forecasting benchmark \sep machine learning



\end{keyword}

\end{frontmatter}


\newpage
\section{Introduction}
\label{sec:introduction}
Modern energy systems are undergoing a fundamental transformation, shifting from existing centralized structures with a small number of large-scale dispatchable generation units with predictable load patterns towards high shares of volatile renewable and decentralized generation. This integration requires higher flexibility and demand response, increasing system complexity. To address this, data-driven approaches are a promising solution for optimizing the operation and planning of energy infrastructure~\cite{Hossain2019BigData}. A key component of these is the anticipation of upcoming system states, e.g., from a market, grid, or system-stability perspective. Hence, accurate time series forecasting will play an essential role in the future of the digitalized energy system. For example, traders can optimize their market positions using precise load and generation forecasts, grid operators need reliable predictions to determine whether a power plant curtailment is necessary to maintain grid stability, and balancing authorities want to estimate their ancillary service needs~\cite{Rolnick2022Climate, Antonopoulos2020DSR}.

In recent years, machine-learning-based prediction models have emerged as state-of-the-art approaches across applications such as electricity load and price forecasting~\cite{Browell2025HEFTcom, Pu2025HybridStrategy}, district heating demand~\cite{Wahl2026HeatBenchmark}, and aggregated grid-level residential demand forecasting, surpassing
standard load profiles~\cite{Bayer2024DigitalTwin}. Despite their good predictive performance when optimized for specific energy-forecasting tasks, machine-learning-based approaches lead to several challenges. Many energy systems constitute critical infrastructure, making many energy time series data highly sensitive and thus limiting their availability~\cite{Donti2021ML}. Furthermore, current energy systems and markets are fractured in their organizational structures, driven by unbundling requirements, market regulations, and the increasing decentralization of assets, which leads to energy data silos that are not accessible for cross-application use cases. Additionally, many newly deployed energy assets, particularly in the rapid upscaling of renewable energy, frequently lack sufficiently long historical records required for training~\cite{Aslam2021Survey,Pan2010Transfer}. The global transformation of the energy system further reinforces the challenge of changing data distributions, often requiring recurring model updates~\cite{Ditzler2015Nonstationary,Haselbeck2021EVARS}. Beyond that, niche applications or forecasting for small-scale energy assets often do not justify the effort required by specialized models. These limitations motivate the development of forecasting approaches that need less task-specific training data, generalize across datasets, and remain stable under distribution shifts.

Time series foundation models (TSFMs) have recently gained traction in the machine-learning community. In particular, Chronos~\cite{Ansari2024Chronos} attracted attention in 2024 by adapting existing
language model architectures through minimal modifications -- requiring only tokenization via scaling and quantization -- and demonstrating strong zero-shot performance across diverse domains including finance, healthcare, nature, retail, mobility, and energy~\cite{Ansari2024Chronos,Ansari2025Chronos2}. Since then, general-purpose
benchmarks such as GIFT-Eval~\cite{Aksu2024GIFTEVAL} and fev-bench
~\cite{Shchur2025FEVBench} have adopted similar domain coverage to compare the growing number of TSFMs systematically.
 By learning generalizable representations from large and diverse pretraining datasets, these models often achieve competitive performance on previously unseen data in zero-shot settings, i.e., without any task-specific training. This broad applicability makes them particularly promising for energy time series forecasting. In the following, we use the shorter term \textit{energy forecasting} synonymously. A more detailed description of TSFM architectures and modeling approaches is provided in Section~\ref{sec:related_work} (Related Work).

Recent works have begun to explore TSFMs for specific energy use cases. Meyer et al.~\cite{Meyer2024Benchmarking} benchmark TSFMs for household load forecasting and find that zero-shot models can match trained-from-scratch Transformers, particularly with longer context. Yet, the study is limited to univariate settings without covariates or machine-learning comparison partners. Hornek et al.~\cite{Sartipi2025Benchmarking} evaluate TSFMs for electricity price forecasting across five European markets and report that task-specific machine-learning methods consistently outperform all tested TSFMs in zero-shot mode -- again without considering covariates or fine-tuning. Simeone~\cite{Simeone2026Benchmark} assesses TSFMs for energy load forecasting on a single dataset without covariate support and a comparison with machine-learning approaches. Ferdaus et al.~\cite{Ferdaus2025Foundation} provide a structured review of foundation models for clean energy forecasting but do not conduct a dedicated benchmark.

Despite the general success and potential benefits of TSFMs, multiple gaps in their application to energy data exist in current research. General-purpose benchmarks treat energy as a single category across many domains, lacking the granularity to determine in which energy use case TSFMs perform better or worse than task-specific models such as random forests or XGBoost. As shown above, recent domain-specific studies focus on individual energy use cases with narrow dataset coverage and do not include newer models, particularly those supporting covariate-informed forecasting, nor do they systematically compare against machine-learning baselines. In summary, there is a lack of comprehensive benchmarking of TSFMs on energy data that includes univariate and covariate approaches, zero-shot and task-specific fine-tuning settings, as well as a comparison with task-specifically trained and tuned state-of-the-art machine-learning methods. In this work, we address these research gaps by empirically investigating the potential of TSFMs to overcome the key challenges of energy data forecasting identified above. We therefore introduce the \textbf{FETS} (\textbf{F}oundation models in \textbf{E}nergy \textbf{T}ime \textbf{S}eries Forecasting) benchmark. Specifically, this paper makes the following contributions:
\begin{itemize}
    \item A stakeholder-driven taxonomy of energy-forecasting use cases along stakeholders, forecast attributes, and data categories (Figure~\ref{fig:energy_forecasts}, Table~\ref{tab:stakeholder_forecasts}).
    \item A collection of 54 energy datasets spanning nine energy-specific data categories.
    \item A benchmark of recent TSFMs in univariate, covariate-informed, and fine-tuned modes against task-specifically tuned XGBoost and random forest.
    \item Sensitivity analyses on context length, forecast horizon, forecastability, and aggregation level.
\end{itemize}
Our findings provide a comprehensive empirical assessment of foundation models for energy time series forecasting, including their strengths and limitations across diverse settings, and offer guidance for real-world energy applications.

The paper follows a hybrid structure combining the IMRaD convention~\cite{IMRAD} with the CRISP-DM methodology~\cite{CRISP-DM}. Section~\ref{sec:related_work} reviews related work, followed by energy-forecasting use cases (Section~\ref{sec:usecases}) and the materials and methods (Section~\ref{sec:materials and methods}) covering the dataset collection, evaluated models, and evaluation setup. Benchmark results are presented in Section~\ref{sec:results} and discussed separately in Section~\ref{sec:discussion}, before Section~\ref{sec:conclusion} concludes with key findings and future deployment directions.

\section{Related Work}
\label{sec:related_work}

In the energy domain, forecasting has traditionally been addressed using statistical approaches that account for temporal patterns, such as ARIMA, and machine-learning-based methods for tabular data, e.g., random forests and gradient boosting techniques~\cite{Antonopoulos2020DSR,Donti2021ML}. Comprehensive reviews of electricity price forecasting established early on that no single model class dominates: the heterogeneous statistical properties of energy time series -- spikes, mean reversion, calendar effects -- instead motivate dataset-specific tuning and ensemble strategies~\cite{Weron2014EPF}. As a consequence, practical energy-forecasting pipelines are typically built around domain-specific feature engineering, e.g., leveraging temperature and weather covariates, calendar and holiday indicators, lagged target values, and recency effects, used to task-specifically optimize comparatively small, well-regularized models. Lessons from the Global Energy Forecasting Competitions (GEFCom)~\cite{Hong2016Probabilistic} point in the same direction: in the wind and solar tracks, most highly ranked approaches used variants of gradient boosting and quantile regression, and careful data cleansing and feature handling characterized the winning entries. In line with these findings, gradient-boosted trees such as XGBoost~\cite{Chen2016XGBoost} dominate forecasting competitions: they were part of all winning solutions of the ASHRAE Great Energy Predictor III competition~\cite{Miller2020ASHRAE}, displaced deep-learning approaches among the top entries of the M5 competition~\cite{Makridakis2022M5,Januschowski2022Trees}, and prevail in the recent HEFTCom challenge~\cite{Browell2025HEFTcom,Pu2025HybridStrategy}. Furthermore, random forest~\cite{Breiman2001RandomForest} remains a robust, stable, and simple baseline widely adopted in practical applications due to its minimal tuning requirements and interpretability.

Large-scale empirical comparisons further report that such compact, task-specifically tuned tree ensembles remain highly competitive with task-specific deep-learning architectures while requiring substantially less hyperparameter tuning~\cite{ShwartzZiv2022Tabular}. The limited historical data typical for individual energy-forecasting tasks often leads to advantages for small, low-variance models~\cite{Fezzi2020SizeMatters} with strong inductive biases and hand-crafted features in comparison with larger, data-hungry architectures that learn their feature representations from the data itself, a pattern confirmed beyond the energy domain for small- to medium-sized tabular data in general~\cite{Grinsztajn2022Trees}. Exceptions exist where long histories are available: on six years of hourly Belgian day-ahead prices, a deep feed-forward network reached 12.3\% sMAPE versus 13.7--13.9\% for gradient-boosted trees and 14.7--15.3\% for random forests~\cite{Lago2018DLvsTrad}. Beyond such competition settings, no single model class dominates: deep-learning approaches lead in individual studies, such as on offshore wind power~\cite{Insunza2026Wind} or across hundreds of correlated retail series~\cite{Theodoridis2022Boosting}, although such results can reverse with the forecast horizon even within a single study~\cite{Wang2020Thermal}; the available data volume and tuning effort largely determine which model class prevails~\cite{Grinsztajn2022Trees,McElfresh2023GBDT}. Together, these findings establish XGBoost and random forest -- combined with energy-specific feature engineering -- as strong and widely applicable task-specific baselines for energy forecasting.

\subsection{Time Series Foundation Models}
TSFMs, as well as tabular foundation models, have emerged as a new class of models for time series and tabular data forecasting. TSFMs are typically large, sequential, deep-learning-based models pretrained on large volumes of diverse real-world and synthetic time series data. Similar to tokenization in natural language processing, an input sequence is divided into fixed-length patches (e.g.\ 32 timesteps), each transformed into a learned embedding vector. During pretraining, models learn sequential dependencies between these embeddings, often through masked prediction tasks, to learn generalizable patterns transferable to previously unseen data. Similar principles have recently been proposed for tabular foundation models, which also aim to learn transferable representations through pretraining on large, diverse datasets comprising both real-world and synthetic data. Hence, after pretraining, TSFMs and tabular foundation models enable downstream forecasting on unseen data, often achieving good performance in zero-shot and few-shot settings, respectively. While sharing similar pretraining strategies, current foundation models differ regarding their underlying architectural design and modeling assumptions. Several approaches use transformer-based sequential models~\cite{Vaswani2017Attention} that operate on patch embeddings of time series, e.g., Chronos-2~\cite{Ansari2025Chronos2}, TimesFM~\cite{Das2024TimesFM}, Moirai2~\cite{Liu2025Moirai2}, and the multivariate Toto-2.0~\cite{Khwaja2026Toto2}. Further state-of-the-art approaches include FlowState~\cite{Graf2025FlowState}, which uses state-space models~\cite{Gu2022S4,Smith2023S5}, and TiRex~\cite{Auer2025TiRex}, which builds upon an xLSTM-based architecture~\cite{Beck2024xLSTM}; its successor, TiRex-2~\cite{Podest2026TiRex2}, extends this architecture to covariate-informed and multivariate forecasting. For tabular data, TabPFN-TS~\cite{Grinsztajn2025TabPFN} employs a transformer-based prior-data fitted network, building on the TabPFN foundation model pretrained purely on synthetic data~\cite{Hollmann2025TabPFN} and extended to time series forecasting via temporal featurization~\cite{Hoo2025TabPFNTS}. A further distinction between these state-of-the-art approaches lies in the forecasting modes they support: univariate forecasting, predicting a single time series without additional information; covariate-informed forecasting, incorporating auxiliary features; and multivariate forecasting, jointly modeling multiple related time series. While FlowState and Toto-2.0 do not support covariate-informed forecasting, the other methods mentioned above support covariates, which might be beneficial for energy forecasting. Figure~\ref{fig:zeroshot_forecasting} shows a schematic example of univariate and covariate-informed zero-shot forecasting. In univariate mode, the model input consists of a segment of the target time series (the context), and the model output comprises a configurable set of forecast quantile trajectories (the horizon). In covariate-informed mode, the inputs are extended with covariates, such as weather forecasts or holiday indicators, enabling the model to capture dependencies between the target and covariates, e.g., via cross-attention mechanisms.
\begin{figure}[h]
	\centering
	\begin{tikzpicture}
		\begin{axis}[
			name=forecast,
			width=\columnwidth,
			height=4cm,
			xlabel={},
			ylabel={Value},
			xmin=0, xmax=192,
			ymin=-5, ymax=70,
			xtick={0,24,48,72,96,120,144,168,192},
			xticklabels={},
			ytick={0,20,40,60},
			grid=major,
			legend style={
				at={(0.46,1.05)},
				anchor=south,
				legend columns=4,
				font=\footnotesize,
				/tikz/every even column/.append style={column sep=0.3cm}
			},
			clip=false,
			]
			
			\addplot[
				color=blue,
				thick,
				domain=0:168,
				samples=300
			] {
				30 + 20*sin(deg(x/13.4))
			};
			\addlegendentry{Historical Data}
			
			\draw[dashed, gray, thick] (axis cs:168,0) -- (axis cs:168,70);
			\node[anchor=south, gray] at (axis cs:164,58) {\small Forecast Start};
			
			\fill[red!10] 
				(axis cs:168,{30 + 20*sin(deg(168/13.4)) - 0.3*(168-168) - 8 - 0.1*(168-168)})
				\foreach \x in {168,169,...,192} {
					-- (axis cs:\x,{30 + 20*sin(deg(\x/13.4)) - 0.3*(\x-168) - 8 - 0.1*(\x-168)})
				}
				\foreach \x in {192,191,...,168} {
					-- (axis cs:\x,{30 + 20*sin(deg(\x/13.4)) - 0.3*(\x-168) + 8 + 0.1*(\x-168)})
				}
				-- cycle;
			
			\addplot[
				forget plot,
				color=red,
				thin,
				dotted,
				domain=168:192,
				samples=100
			] {
				30 + 20*sin(deg(x/13.4)) - 0.3*(x-168) + 8 + 0.1*(x-168)
			};
			
			\addplot[
				forget plot,
				color=red,
				thin,
				dotted,
				domain=168:192,
				samples=100
			] {
				30 + 20*sin(deg(x/13.4)) - 0.3*(x-168) - 8 - 0.1*(x-168)
			};
			
			\addplot[
				color=red,
				thick,
				dashed,
				domain=168:192,
				samples=100
			] {
				30 + 20*sin(deg(x/13.4)) - 0.3*(x-168)
			};
			\addlegendentry{Model Forecast}
			
			\addlegendimage{area legend, fill=red!10, draw=none}
			\addlegendentry{Uncertainty (10\%-90\%)}
			
			\addplot[
				color=green!70!black,
				thick,
				domain=168:192,
				samples=100
			] {
				30 + 20*sin(deg(x/13.4)) - 0.3*(x-168) + 3*sin(deg((x-168)/3))
			};
			\addlegendentry{Actual Values}
			
            \draw [decorate,decoration={brace,amplitude=6pt,mirror},blue,semithick] 
            	(axis cs:1,-10) -- (axis cs:167,-10) 
            	node[midway,below=4pt,blue] {\footnotesize Context};
            
            \draw [decorate,decoration={brace,amplitude=6pt,mirror},red,semithick] 
            	(axis cs:169,-10) -- (axis cs:191,-10) 
            	node[midway,below=4pt,red] {\footnotesize Horizon};

		\end{axis}
		
		\begin{axis}[
			name=covariates,
			at={(forecast.below south west)},
			anchor=north west,
			yshift=-0.8cm,
			width=\columnwidth,
			height=3cm,
			xlabel={Time},
			ylabel={Covariates},
			xmin=0, xmax=192,
			ymin=0, ymax=1.2,
			xtick={0,24,48,72,96,120,144,168,192},
			xticklabels={0,1,2,3,4,5,6,7,8},
			ytick={0,0.5,1},
			grid=major,
			legend style={
				at={(0.47,1.08)},
				anchor=south,
				legend columns=2,
				font=\footnotesize,
				/tikz/every even column/.append style={column sep=0.5cm}
			},
			]
			
			\addplot[
				color=blue,
				thick,
				domain=0:192,
				samples=300
			] {
				0.5 + 0.3*sin(deg(x/15))
			};
			\addlegendentry{Covariate 1 (e.g., Temperature Forecast)}
			
			\addplot[
				color=blue!60,
				thick,
				dashed,
				domain=0:192,
				samples=300
			] {
				0.7 + 0.2*sin(deg(x/10)) + 0.1*cos(deg(x/20))
			};
			\addlegendentry{Covariate 2 (e.g., Wind Speed Forecast)}
			
			\draw[dashed, gray, thick] (axis cs:168,0) -- (axis cs:168,1.2);
			
		\end{axis}
		
	\end{tikzpicture}
	\caption{Schematic illustration of zero-shot time series forecasting in univariate and covariate modes: The upper panel shows historical data (7-day context), the model forecast with median (1-day horizon), the 10\%-90\% uncertainty interval, and actual values. The lower panel displays optional covariates that are used as additional input features in covariate mode.}
	\label{fig:zeroshot_forecasting}
\end{figure}

\subsection{Benchmarks and Energy-Specific Applications}
To compare the predictive performance of these foundation models, several general-purpose benchmarks, such as GIFT-Eval~\cite{Aksu2024GIFTEVAL} and fev-bench~\cite{Shchur2025FEVBench}, have been published. However, these benchmarks show limited coverage of the energy domain, with GIFT-Eval including only four energy-related datasets and fev-bench providing data on aggregated country and household loads, electric price forecasting, and solar as well as wind production forecasting~\cite{Zhou2024SDWPF}. Moreover, neither benchmark evaluates task-specifically tuned, covariate-informed tree ensembles: GIFT-Eval includes no tree-based baselines, and the fev-bench baselines (e.g., CatBoost, TFT) are evaluated without covariates. On GIFT-Eval, the leading pretrained models already rank ahead of the task-trained deep-learning baselines~\cite{Ansari2025Chronos2,Podest2026TiRex2}. A complementary direction is pursued by TS-Arena~\cite{Meyer2025TSArena}, a live forecasting platform with a pre-registration protocol that evaluates models on genuinely unseen future data to eliminate train--test contamination by design, with an initial application focus on the energy sector; however, it currently provides neither a systematic comparison against machine-learning baselines such as XGBoost or random forest, nor support for covariate-informed forecasting. As discussed in the introduction, recent works also explore TSFMs for energy data, focusing on individual use cases such as household load forecasting~\cite{Meyer2024Benchmarking}, electricity price forecasting~\cite{Sartipi2025Benchmarking}, univariate energy load forecasting~\cite{Simeone2026Benchmark}, and a review of clean energy forecasting~\cite{Ferdaus2025Foundation}. These contributions provide initial insights into the applicability of TSFMs in energy forecasting, but each focuses on a single energy use case and typically omits covariate-informed modes, fine-tuning, or comparisons with machine-learning baselines. Building on this, our \textbf{FETS} benchmark extends and complements prior work as follows:
\begin{itemize}
    \item Extending the general-purpose benchmarks GIFT-Eval~\cite{Aksu2024GIFTEVAL} and fev-bench~\cite{Shchur2025FEVBench} into the energy domain, we systematize the evaluation along stakeholder-driven energy use cases and assemble 54 datasets across nine energy-specific data categories.
    \item Building on energy-specific TSFM studies~\cite{Meyer2024Benchmarking,Sartipi2025Benchmarking,Simeone2026Benchmark}, which so far address a single use case in univariate zero-shot mode, we evaluate covariate-informed TSFMs such as Chronos-2~\cite{Ansari2025Chronos2} across diverse energy use cases under both zero-shot and fine-tuned conditions.
    \item In contrast to TS-Arena~\cite{Meyer2025TSArena}, which lacks machine-learning baselines, covariate support, and fine-tuning, we benchmark against task-specifically tuned XGBoost and random forest as the de facto state-of-the-art in practical energy forecasting.
\end{itemize}
To the best of our knowledge, our study provides the first systematic, use-case-differentiated comparison of where foundation models surpass task-specific machine learning-based approaches in energy forecasting, and where the latter remain the stronger choice.

\section{Forecasting Use Cases in the Energy Sector}
\label{sec:usecases}
Energy data spans a wide range of specific subdomains that the general-purpose benchmarks discussed above do not differentiate. To enable a more fine-grained analysis of TSFMs on energy data, we propose an energy-specific taxonomy that clusters datasets by representative fields of current forecasting applications, as depicted in Figure~\ref{fig:energy_forecasts}. The taxonomy is subdivided by stakeholders, data categories, and forecasting attributes. It spans a tree of involved parties, energy applications, and time series characteristics, whose leaves reflect the dimensions to be considered. Some leaf-level examples are specific to the European and German market, but the overall structure is broadly transferable. The forecast attributes shown extend those discussed in Ferdaus et al.~\cite{Ferdaus2025Foundation}.

The introduced \textbf{energy stakeholders} are derived from the ENTSO-E market role definitions~\cite{ENTSOE2022HRM} and abstracted to ensure transferability to energy markets beyond Europe, following the stakeholder categorization discussed in Antonopoulos et al.~\cite{Antonopoulos2020DSR}. 
\textbf{Data categories} reflect the energy sectors: electricity, heat, 
and mobility~\cite{BDEW2024Energieflussbild}, with specific requirements for the forecasting of respective time series. The electricity domain stands out for its central role in a fully carbon-free energy system~\cite{Rolnick2022Climate} in the future, spanning generation, load, markets, system services, and grid time series. Each forecast can be characterized by attributes such as horizon (long-term to real-time), grid level (transmission to low voltage), and spatial aggregation (national to individual asset). \textbf{Forecast attributes} introduce the technical dimension, covering temporal data resolution, forecasting horizons, external influences, addressed grid levels, and regional or type-level aggregation~\cite{Ferdaus2025Foundation}.

\vspace{-0.3cm}
\begin{figure}[htbp]
  \centering
  \begin{adjustbox}{width=\textwidth,center}
  \begin{forest}
    for tree={
      grow'=east,
      parent anchor=east, child anchor=west,
      edge={gray!60, line width=0.4pt},
      edge path={\noexpand\path[\forestoption{edge}]
        (!u.parent anchor) -- +(5pt,0) |- (.child anchor)\forestoption{edge label};},
      l sep=9mm, s sep=2mm,
      anchor=west, align=left,
      font=\scriptsize,
      rounded corners=2pt, draw=black!40, line width=0.4pt, inner sep=3pt,
    }
    [Energy\\forecasts, font=\scriptsize\bfseries, fill=gray!15, align=center
      [Stakeholders, font=\scriptsize\bfseries, fill=red!18
        [{System operator · Market operator · Grid operator\\Balance responsible · Trader · Retailer · Producer\\Consumer · Balancing service · \ldots}, fill=red!8]
      ]
      [Data\\categories, font=\scriptsize\bfseries, fill=green!22, align=center
        [Electricity, font=\scriptsize\bfseries, fill=green!14
          [Generation, font=\scriptsize\bfseries, fill=green!9
            [{Dispatchable: Gas · Battery · \ldots}, fill=green!6]
            [{Non-dispatchable: PV · Wind · \ldots}, fill=green!6]]
          [Load, font=\scriptsize\bfseries, fill=green!9
            [{Household · Industry · \ldots}, fill=green!6]]
          [Market, font=\scriptsize\bfseries, fill=green!9
            [{Day-ahead · Intraday · reBAP · \ldots}, fill=green!6]]
          [System, font=\scriptsize\bfseries, fill=green!9
            [{aFRR · mFRR · FCR · \ldots}, fill=green!6]]
          [Grid, font=\scriptsize\bfseries, fill=green!9
            [{Import · Export · \ldots}, fill=green!6]]
        ]
        [Heat, font=\scriptsize\bfseries, fill=green!14
          [{Space heat · Process heat · District heat · \ldots}, fill=green!6]]
        [Mobility, font=\scriptsize\bfseries, fill=green!14
          [{EV charging load · Plug-in/Plug-out time · \ldots}, fill=green!6]]
      ]
      [Forecast\\attributes, font=\scriptsize\bfseries, fill=blue!15, align=center
        [Resolution, font=\scriptsize\bfseries, fill=blue!10
          [{5\,min · 15\,min · 1\,h · \ldots}, fill=blue!6]]
        [Horizon, font=\scriptsize\bfseries, fill=blue!10
          [{Long · Mid · Short · Real-time}, fill=blue!6]]
        [Influences, font=\scriptsize\bfseries, fill=blue!10
          [{Weather · Price · Policy · Holidays · \ldots}, fill=blue!6]]
        [Grid level, font=\scriptsize\bfseries, fill=blue!10
          [{Transmission · HV · MV · LV}, fill=blue!6]]
        [Aggregation, font=\scriptsize\bfseries, fill=blue!10
          [{National · Cluster · Local region · Individual}, fill=blue!6]]
      ]
    ]
  \end{forest}
  \end{adjustbox}
  \vspace{-0.2cm}
  \caption{An overview of energy forecasts from different perspectives and dimensions. It constitutes a synthesis of an abstracted interpretation of ENTSO-E market role definitions \cite{ENTSOE2022HRM}, and forecast attributes derived from Ferdaus et al. \cite{Ferdaus2025Foundation}, common energy balances on a state or national level \cite{BDEW2024Energieflussbild}, with aggregated nodes summarizing consumption and production sectors, as well as own considerations.}
  \label{fig:energy_forecasts}
\end{figure}

\clearpage
Table~\ref{tab:stakeholder_forecasts} further links each energy system stakeholder to its typical data categories of interest. The tasks we consider focus on short-term forecasting applications, which are particularly relevant to operational planning and often face challenges such as non-stationarity, making potentially well-generalizing TSFMs a promising approach.

\begin{scriptsize}
\begin{longtable}{p{0.23\textwidth}p{0.33\textwidth}p{0.34\textwidth}}
\caption{Stakeholders' interests in energy forecasts with a focus on short-term applications} \label{tab:stakeholder_forecasts} \\
\hline
\textbf{Stakeholder} & \textbf{Role / task related to forecasts} & \textbf{Typical forecasted targets (data categories)} \\
\hline
\endfirsthead

\multicolumn{3}{c}%
{{\bfseries \tablename\ \thetable{} -- continued from previous page}} \\
\hline
\textbf{Stakeholder} & \textbf{Role / task related to forecasts} & \textbf{Typical forecasted targets (data categories)} \\
\hline
\endhead

\hline \multicolumn{3}{r}{{Continued on next page}} \\
\endfoot

\hline
\endlastfoot

System Operator &
Real-time grid balancing, frequency control, and system security (intraday to day-ahead). &
Aggregated \textbf{load}, \textbf{non-dispatchable} generation (wind, PV), \textbf{balancing} reserve activation (aFRR, mFRR), \textbf{grid} power flows. \\

\hline
Grid Operator &
Congestion forecasting to anticipate line/transformer overloads, enabling pre-emptive switching operations, redispatch requests, and real-time monitoring. &
Regional \textbf{load}, distributed \textbf{non-dispatchable} generation, \textbf{mobility} (EV peaks), \textbf{heat} (heat pumps), \textbf{grid} line flows, grid components/assets, i.e., network transformer load. \\

\hline
Balance Responsible Party &
Minimizing imbalance costs through portfolio balancing (15-min to day-ahead). &
Portfolio \textbf{load}, \textbf{non-dispatchable} generation, \textbf{balancing} energy prices (reBAP), imbalance volumes, balancing discrepancies, short-term/ID prices. \\

\hline
Trader &
Intraday and day-ahead trading, arbitrage, and position optimization. &
\textbf{Market} prices (day-ahead, intraday), \textbf{load} patterns, \textbf{non-dispatchable} generation, cross-border flows. \\

\hline
Producer &
Unit commitment and dispatch optimization (5min ahead to day-ahead). &
Plant-specific \textbf{non-dispatchable} generation (wind, PV), \textbf{market} prices, \textbf{balancing} service activation. \\

\hline
Energy Supplier &
Short-term procurement and balancing group management for retail customers. &
Customer \textbf{load}, \textbf{mobility} (EV charging), \textbf{heat} (heat pumps), \textbf{market} prices, distributed generation. \\

\hline
Market Operator &
Day-ahead and intraday market clearing and price formation. &
Aggregated \textbf{load}, \textbf{non-dispatchable} generation, cross-border capacity, \textbf{market} bid/offer curves. \\

\hline
Balancing Service Provider &
Bidding strategies for frequency restoration reserves (aFRR, mFRR). &
\textbf{Balancing} reserve activation, system frequency, \textbf{market} prices for reserve energy, flexible assets. \\

\hline
Consumer (industry / commercial) &
Demand response, peak shaving, and intraday procurement for flexible loads. &
Site \textbf{load}, on-site \textbf{non-dispatchable} generation (PV), process \textbf{heat}, \textbf{mobility} (fleet), \textbf{market} prices. \\

\hline
Consumer (residential with HEMS) &
Home energy management, optimal scheduling of flexible loads and storage. &
Household \textbf{load}, \textbf{heat} (heat pump), \textbf{mobility} (EV charging), rooftop PV, battery storage, \textbf{market} prices. \\

\hline
Metering Point Administrator / Data roles &
Real-time data validation and operational monitoring of metering infrastructure. &
Metered \textbf{load}, \textbf{mobility} (EV), \textbf{heat} (heat pumps), anomaly detection, data volume validation. \\

\end{longtable}
\end{scriptsize}

\clearpage
\section{Materials and Methods}
\label{sec:materials and methods}
In this work, we provide a systematic assessment of TSFMs on a comprehensive set of energy-forecasting data. Figure~\ref{fig:benchmark_architecture} summarizes the main components of our study. First, we consider representative datasets from different categories of the energy sector following the above-described taxonomy, allowing both general and task-specific conclusions. Second, we evaluate a wide range of prediction models, including TSFMs and machine-learning approaches, as well as across the TSFM forecasting modes (1) univariate zero-shot, (2) covariate zero-shot, and (3) task-specific fine-tuning. Third, we define a consistent evaluation framework that enables a fair comparison and further analysis of the behavior of the employed approaches. 

\begin{figure}[htbp]
\centering
\begin{tikzpicture}[
    node distance=0.3cm,
    mainbox/.style={rectangle, draw, very thick, rounded corners=5pt, minimum height=6.5cm, align=center, font=\bfseries\small},
    modebox/.style={rectangle, draw, thick, rounded corners=3pt, minimum height=1.6cm, text width=3.5cm, align=center, font=\scriptsize\bfseries, fill=white, text depth=1.3cm},
    databox/.style={rectangle, draw, thick, rounded corners=3pt, minimum height=0.85cm, text width=2.2cm, align=left, font=\scriptsize},
    minibox/.style={rectangle, draw, thick, rounded corners=2pt, minimum height=0.4cm, text width=1.3cm, align=center, font=\scriptsize, text depth=0.1cm},
    wideminibox/.style={rectangle, draw, thick, rounded corners=2pt, minimum height=0.4cm, text width=1.9cm, align=center, font=\scriptsize, text depth=0.1cm},
    metricbox/.style={rectangle, draw, thick, rounded corners=2pt, minimum height=0.6cm, text width=2.7cm, align=center, font=\scriptsize},
    covbox/.style={rectangle, draw, very thick, rounded corners=4pt, minimum height=1cm, text width=2.7cm, align=left, fill=blue!15, font=\scriptsize},
    settingsbar/.style={rectangle, draw, very thick, rounded corners=4pt, minimum height=1.2cm, text width=7.5cm, align=left, fill=gray!15, font=\scriptsize},
    arrow/.style={->, very thick, >=stealth}
]

\node[mainbox, minimum width=3cm, fill=blue!10] (data) at (0,0) {};
\node[above=0.05cm of data.north, font=\bfseries] {Data};

\node[mainbox, minimum width=4.2cm, minimum height=8.0cm, fill=red!8] (model) at (4.2,-0.75) {};
\node[above=0.05cm of model.north, font=\bfseries] {Models};

\node[mainbox, minimum width=3.2cm, fill=green!8, minimum height=8.0cm] (eval) at (8.8,-0.75) {};
\node[above=0.05cm of eval.north, font=\bfseries] {Evaluation};

\node[databox, fill=green!15] (load) at (0, 2.2) {\textbf{Non-Dispatchable Generation}\\PV\\};
\node[databox, fill=green!15] (gen) at (0, 1.0) {\textbf{Market Data}\\Day-ahead};
\node[databox, fill=orange!35] (heat) at (0, -0.1) {\textbf{Heat}\\ Heat Demand};
\node[databox, fill=purple!30] (mobility) at (0, -1.2) {\textbf{Mobility}\\ EV Load};
\node[databox, fill=gray!5] (market) at (0, -2.3) {\textbf{+5 more}\\\textbf{categories}\\54 series in total};

\node[modebox, fill=red!15, minimum height=2.4cm, text depth=2.1cm] (uni_mode) at (4.2, 1.50) {Univariate};
\node[minibox, fill=red!20] (uni_m1) at (3.35, 2.05) {Chronos-2};
\node[minibox, fill=red!20] (uni_m2) at (5.05, 2.05) {TiRex-2};
\node[minibox, fill=red!20] (uni_m3) at (3.35, 1.45) {TimesFM};
\node[minibox, fill=red!20] (uni_m4) at (5.05, 1.45) {FlowState};
\node[minibox, fill=red!20] (uni_m5) at (4.2, 0.85) {Toto-2.0};

\node[modebox, fill=red!25, minimum height=1.9cm, text depth=1.6cm] (cov_mode) at (4.2, -1.00) {Covariate};
\node[minibox, fill=red!30] (cov_m1) at (3.35, -0.70) {Chronos-2};
\node[minibox, fill=red!30] (cov_m2) at (5.05, -0.70) {TiRex-2};
\node[minibox, fill=red!30] (cov_m3) at (3.35, -1.30) {TimesFM};
\node[minibox, fill=red!30, text depth=0.3cm] (cov_m4) at (5.05, -1.40) {TabPFN-TS};

\node[modebox, fill=red!35, minimum height=1.9cm, text depth=1.6cm] (train_mode) at (4.2, -3.25) {Training/Fine-tuning};
\node[minibox, fill=red!40] (train_m1) at (3.35, -2.95) {Chronos-2};
\node[minibox, fill=red!40] (train_m2) at (5.05, -2.95) {XGBoost};
\node[wideminibox, fill=red!40] (train_m3) at (4.2, -3.55) {Random Forest};

\node[metricbox, fill=green!15, align=left] (e1) at (8.8, 2.26) {\textbf{Median NRMSE:}\\[1pt]
\begin{tabular}{@{}l@{\hspace{4pt}}r@{}}
Chronos-2 cov. & 0.472\\
TiRex-2 cov. & 0.474\\
Chronos-2 FT & 0.599\\
XGBoost train. & 0.611
\end{tabular}};
\node[metricbox, fill=green!20] (e2) at (8.8, 0.82) {Median NRMSE per\\Data Category};
\node[metricbox, fill=green!20] (e3) at (8.8, -0.12) {Significance:\\Friedman + Nemenyi};
\node[metricbox, fill=green!20] (e4) at (8.8, -1.23) {Predictability:\\Spectral Entropy\\vs. Performance};
\node[metricbox, fill=green!20] (e5) at (8.8, -2.50) {Sensitivity Analyses:\\Context, Horizon,\\Aggregation Level};
\node[metricbox, fill=green!20] (e6) at (8.8, -3.93) {Additional Metrics:\\MASE, nCRPS,\\Correlation, PICP,\\Sum Ratio};

\draw[arrow] (data.east) -- (data.east -| model.west);
\draw[arrow] ([yshift=0.75cm]eval.west -| model.east) -- ([yshift=0.75cm]eval.west);

\node[covbox] (cov) at (0, -4.5) {\textbf{Covariates:}\\Historical Weather\\Forecasts (Temp.,\\Wind, Radiation)\\Calendar\\(Hour, Day, Holiday)};
\draw[dashed, very thick] (cov.east) -| ++(0.3, 3) coordinate (covfork);

\draw[arrow, dashed] (covfork) |- (2.33, -1.00);

\draw[arrow, dashed] (covfork) |- (train_mode.west);

\draw[dashed, very thick, rounded corners=8pt]
    ([xshift=-0.3cm, yshift=0.7cm]data.north west) rectangle
    ([xshift=0.3cm, yshift=-1.5cm]eval.south east);

\node[settingsbar, text width=6.4cm] (settings) at (5.1, -6.25) {
    \textbf{\faCog \ Benchmark Settings:} \\
    Context Length: 672 -- 8000 (1w -- 12w @ 15min) \\
    Forecast Horizon: 96 -- 288 ~ (1d -- 3d ~~@ 15min) \\
    Rolling Windows: 35
};

\end{tikzpicture}
\caption{Benchmark architecture overview: Datasets are fed into three deployment modes with covariates as additional inputs. Models are evaluated across multiple perspectives with consistent benchmark settings.}
\label{fig:benchmark_architecture}
\end{figure}

\newpage
\subsection{Selected Datasets}
\label{sec:datasets}
Based on the data categories typically of interest to stakeholders (Table~\ref{tab:stakeholder_forecasts}), we selected a representative set of datasets for this study, as presented in Table~\ref{tab:datasets_comprehensive}. Our dataset selection follows four main criteria: (1) open availability, (2) recency with a preference for datasets covering time series data from 2022 onwards, (3) domain relevance based on citation count and market size, and (4) category coverage across all segments given in Table~\ref{tab:stakeholder_forecasts}. The resulting coverage of the collection is summarized in Appendix Table~\ref{tab:coverage_gaps}.
For each dataset and benchmark mode, a representative target subset is used (Appendix Table~\ref{tab:dataset_statistics}), yielding 54 (dataset, target) combinations, hereafter referred to as datasets. Each dataset is assigned to the data categories given in Figure~\ref{fig:energy_forecasts}. The general data section contains datasets that could be assigned to multiple data categories or comprise general feature sets.
Typical meteorological features include temperature (2\,m height), wind speed (at 80\,m, 120\,m, and 180\,m hub heights), shortwave radiation, and direct radiation. Temporal features comprise cyclic encodings of hour, minute, weekday, day, and month using sine and cosine transformations to preserve periodicity. Calendar features include binary indicators for weekends and public holidays derived from state-specific holiday calendars. Most datasets originate from the largest European electricity markets, with additional datasets from Asia and the United States.


\begin{scriptsize}
\begin{longtable}{p{3cm}p{4.2cm}p{3.3cm}p{1.2cm}}
\caption{Comprehensive overview of all datasets used in this study, categorized according to Figure~\ref{fig:energy_forecasts} and providing references besides a general description and outlining their usage in our comparative study.} \label{tab:datasets_comprehensive} \\
\hline
\textbf{Category \& Dataset} & \textbf{Description} & \textbf{Benchmark Usage} & \textbf{Ref.} \\
\hline
\addlinespace[2pt]
\endfirsthead

\multicolumn{4}{c}%
{{\bfseries \tablename\ \thetable{} -- continued from previous page}} \\
\hline
\textbf{Dataset} & \textbf{Description} & \textbf{Benchmark Usage} & \textbf{Ref.} \\
\hline
\endhead

\hline \multicolumn{4}{r}{{Continued on next page}} \\
\endfoot

\hline
\endlastfoot
\addlinespace[4pt]
\multicolumn{4}{l}{\textbf{General Data}} \\
\addlinespace[2pt]
\hline
\addlinespace[2pt]
Calendar Features & Holiday and temporal features (DE, FR, NL, UK, CN, CH, CA, Bavaria, ...) & All benchmark modes; required by TabPFN~\cite{Grinsztajn2025TabPFN,Hoo2025TabPFNTS} & \cite{murza_holidays_2025} \\
ENTSO-E Transparency & Load, generation, and market data for European countries & Country-level load, solar, wind, fossil generation, day-ahead spot market prices & \cite{entsoe_transparency_2025} \\
\hline

\multicolumn{4}{l}{\textbf{Mobility Data}} \\
\addlinespace[2pt]
\hline
\addlinespace[2pt]
Office EV Parking (NL) & EV charging at large office parking lot, Netherlands & Aggregated EV charging power forecasting & \cite{Bont2024} \\
Mobilithek & Public EV charging stations across Germany & Aggregated EV charging (50--25000 EVs) & \cite{Mobilithek2025} \\
UrbanEV & Urban EV charging demand dataset, Shenzhen, China & Urban-scale EV charging demand forecasting & \cite{Li2025} \\
UK Department for Transport & Domestic EV chargepoint analysis data, United Kingdom & Residential EV charging forecasting & \cite{DfT2025} \\
\hline

\multicolumn{4}{l}{\textbf{Heat Data}} \\
\hline
HEAPO & Heat pump smart meter data with inspection protocols & Heat pump electricity demand forecasting & \cite{brudermueller_heapo_2025} \\
Heat Grid Flensburg & Network data of the district heating system for the city of Flensburg from 2020--2024 & District heating forecasting with historical weatherprediction & \cite{freissmann_flensburg_2025} \\
\hline

\multicolumn{4}{l}{\textbf{Non-Dispatchable Generation}} \\
\hline
Energy Forecasting Competition & Hybrid wind and PV forecasting competition, UK & Wind park, PV cluster, and hybrid wind+PV forecasting & \cite{browell_hybrid_2025} \\
Hill of Towie Wind Farm & Onshore wind farm active power measurements, Scotland & 15-min wind farm power forecasting & \cite{clerc_hill_2025} \\
PV Hong Kong & High-resolution rooftop PV generation, 3-year, China & Single rooftop PV system forecasting & \cite{lin_high_2025} \\
DE/FR/DK Power Solar & Country-level solar generation for DE, FR, DK & National solar generation forecasting & \cite{entsoe_transparency_2025} \\
DE Power Wind & Onshore and offshore wind generation for Germany & National wind generation forecasting & \cite{entsoe_transparency_2025} \\
California CAISO & 5-min resolution aggregated load data for California (CAISO) & Aggregated solar forecasting (US market) & \cite{Kanter2025GridStatus,
CAISO2025OASIS} \\
\hline

\multicolumn{4}{l}{\textbf{Dispatchable Generation}} \\
\hline
ENTSO-E Germany & Fossil gas and hard coal generation for Germany & Dispatchable generation forecasting & \cite{entsoe_transparency_2025} \\
California CAISO & 5-min resolution aggregated generation data for California (CAISO) & Battery storage dispatch forecasting (US market) & \cite{Kanter2025GridStatus,
CAISO2025OASIS} \\
\hline

\multicolumn{4}{l}{\textbf{Load Data}} \\
\hline
KIT Company Load & Electricity consumption of 28 German companies (15-min) & Industrial and commercial load forecasting & \cite{huber_electricity_2023} \\
Industrial VEA Profiles & 5359 industrial load profiles & Industrial load forecasting & \cite{tiemann_industrial_2024} \\
DE/FR Power Load & Country-level load for Germany and France & National electricity demand forecasting & \cite{entsoe_transparency_2025} \\
California CAISO & 5-min resolution aggregated load data for California (CAISO) & Aggregated load forecasting (US market) & \cite{Kanter2025GridStatus, CAISO2025OASIS} \\
\hline

\multicolumn{4}{l}{\textbf{Residential Load}} \\
\hline
TSFM Household Benchmark & Household electricity load with foundation model benchmark & Short-term household load forecasting & \cite{Meyer2024Benchmarking} \\
HTW Berlin Households & Representative load profiles at 1-second resolution & Individual and aggregated household load (2--80 households) & \cite{schlemminger_dataset_2022} \\
Lower Saxony Households & Single-family house and heat pump load profiles & Individual household demand with/without PV & \cite{tjaden_repraesentative_nd} \\
\hline

\multicolumn{4}{l}{\textbf{Market Data}} \\
\hline
reBAP & balancing energy price across control areas for Germany & Balancing energy price forecasting & \cite{netztransparenz_rebap_2025} \\
Day-Ahead Prices & EPEX day-ahead auction prices (DE-LU) & Day-ahead electricity price forecasting & \cite{entsoe_transparency_2025} \\
Intraday Prices & Continuous intraday market prices & Intraday price forecasting & \cite{epex_intraday_2024} \\
California CAISO & Locational marginal prices (NP15, SP15, ZP26) & Electricity price forecasting (US market) & \cite{Kanter2025GridStatus, CAISO2025OASIS} \\
\hline

\multicolumn{4}{l}{\textbf{Grid Data}} \\
\hline
50Hertz Line Power & Transmission line power for high-voltage lines & High-voltage line power flow forecasting & \cite{50hertz_netzbelastung_2025} \\
Bayernwerk MV Grid & Medium voltage grid feed-in and consumption & MV grid load and feed-in forecasting & \cite{Bayernwerk2025} \\
LV Grid Feeders (200) & Real-world energy data of 200 low-voltage grid feeders with metadata & Low-voltage grid load forecasting & \cite{treutlein_2025_zenodo,treutlein_2026_preprint} \\
\hline

\multicolumn{4}{l}{\textbf{Balancing Services}} \\
\hline
aFRR Germany & Automatic frequency restoration reserve activation & aFRR activation forecasting (positive/negative) & \cite{netztransparenz_regelleistung_2025} \\
mFRR Germany & Manual frequency restoration reserve (TenneT TSO) & mFRR activation forecasting (positive/negative) & \cite{netztransparenz_regelleistung_2025} \\
NRV Saldo & Network control area balance for Germany & NRV balance forecasting & \cite{netztransparenz_regelleistung_2025} \\
\hline

\multicolumn{4}{l}{\textbf{Weather Data}} \\
\hline
Open-Meteo API & Archived and operational numerical weather prediction forecasts (e.g., DWD ICON-D2, ECMWF IFS) & Weather covariates for all forecasting tasks & \cite{openmeteo_2024,dwd_icon_d2,ecmwf_ifs_2024} \\

\end{longtable}
\end{scriptsize}

\clearpage
\subsection{Data Preparation}
\label{sec:data preparation}
Since 15-minute settlement intervals are the standard in most European and Asian energy markets, all time series are resampled to this resolution. California (CAISO) datasets constitute an exception and are retained at their native 5-minute resolution to reflect US market practices. Datasets with lower native resolution (for example, hourly data) are upsampled by duplicating values within each hour. For datasets with mixed resolutions, such as historic segments at hourly resolution and more recent segments at 15-minute resolution, the coarser segments are upsampled to the finer 15-minute resolution by forward-filling the first value of each interval. 
In addition, some time series exhibit varying resolutions with outliers or data gaps, resulting in irregular sampling intervals. Mobility datasets typically store individual charging events in a tabular format, recording start time, end time, energy delivered (in Wh or kWh), and, if available, maximum charging power per charging point. For datasets lacking explicit power specifications, a standard charging power of 11~kW was assumed, in line with the prevalent three-phase Type~2 AC charging standard in Europe~\cite{IEC62196_2022}. From these event-based records, continuous time series were constructed by converting start time, charging power, and energy delivered into aggregated load profiles at 15-minute resolution. Appendix Table~\ref{tab:dataset_statistics} provides a statistical overview of all datasets, including sample size ($N$), temporal resolution, coefficient of variation (CV), share of negative and positive values, and the Forecastability Index $\phi$, which is formally introduced and described in Section~\ref{sec:eval}. The collection spans a wide range of signal characteristics, with CV values typically ranging from $\approx 0.18$ for stable aggregated signals such as system load to values above $2$ for highly volatile signals such as balancing services ($\text{CV} \approx 1.7\text{--}12$) and individual residential loads ($\text{CV} \approx 0.9\text{--}2.4$). Extreme CV values occur for signals with a near-zero mean, such as low-voltage feeders combining local consumption and PV feed-in ($\text{CV} \approx 27$), where consumption and generation largely cancel out on average; in such cases, the CV becomes disproportionately large and no longer reflects the effective amplitude of the signal. More generally, a high CV indicates strong amplitude variation relative to the mean, but does not necessarily imply low predictability -- a distinction captured by $\phi$, which reflects structural regularity independently of the mean. Temporal resolutions are mostly 15 minutes, with a few datasets at 5~minutes, 30~minutes, or 1~hour. Negative values, which are typical for signed quantities such as grid balancing errors or battery charging power, occur primarily in balancing services, some grid feeders, and dispatchable battery profiles. All datasets, except for a small number subject to license restrictions, are released by the authors in preprocessed form, with per-dataset source links, access conditions, and license information documented in the published record.\footnote{\textbf{Data available at:} \href{https://doi.org/10.5281/zenodo.19418721}{10.5281/zenodo.19418721}~\cite{obermeier_dataset_fets_2026}.}

\textbf{Covariate engineering:} Similar to feature-based machine learning, covariate-informed TSFMs benefit when covariates provide predictive information, which requires domain-specific covariate engineering. Calendar features and historical weather forecasts serve as covariates if they are available and relevant for a dataset. Weather covariates are retrieved for dataset-specific coordinates from the Open-Meteo archive~\cite{openmeteo_2024} and constructed so that they contain only information available at inference time. Within the context window, archived historical forecasts are used, while the forecast horizon is covered by the most recent ECMWF IFS forecast run~\cite{ecmwf_ifs_2024} (issued at 00 or 12\,UTC) that was available at least nine hours before the forecast origin, mirroring operational dissemination delays.\footnote{For the few hourly datasets the longest horizon (288\,h) exceeds the 240\,h ECMWF single-run lead; the remaining steps are forward-filled.} This default setting therefore reflects a realistic operational deployment, without leakage of future observed weather not available at inference time. Figure~\ref{fig:covariate_schema} illustrates this construction schematically.\footnote{A few datasets instead come with their own covariates and do not use these coordinate-based Open-Meteo weather covariates: the HEFTCom wind/PV competition datasets, the Hill of Towie wind farm, and PV Hong Kong.} Alternative weather covariate sources -- archived historical forecasts, real-time numerical weather prediction (NWP) runs, and ERA5 reanalysis -- are compared in a dedicated sensitivity analysis (Appendix Table~\ref{tab:weather_sensitivity}).

\begin{figure}[h]
\centering
\begin{tikzpicture}[
    font=\small,
    block/.style={rectangle, draw, minimum height=0.52cm, inner sep=0pt},
    run/.style={rectangle, draw, minimum height=0.30cm, inner sep=0pt},
    lbl/.style={font=\footnotesize},
    issuedot/.style={circle, fill=black, inner sep=1.1pt},
]

\colorlet{colorTargetPast}{blue!15}
\colorlet{colorTargetForecast}{purple!25}
\colorlet{colorRunUsed}{orange!35}
\colorlet{colorRunUnused}{orange!10}
\colorlet{colorIfsUsed}{orange!35}
\colorlet{colorIfsUnused}{orange!10}
\colorlet{colorCalendar}{green!15}

\def\xForecastOrigin{5.6}        
\def\xEnd{9.2}                   
\def\xIfsIssue{4.9}              
\def\widthRunUsed{1.4}           
\def\widthRunTail{0.8}           

\def\yTargetRow{4.15}
\def\yRunLabel{3.55}
\def\yRunFirst{3.10}
\def\yRunStep{0.36}
\def\yCalendarRow{0.45}

\pgfmathsetmacro{\yIfsRow}{\yRunFirst - 4*\yRunStep}
\pgfmathsetmacro{\yTop}{\yTargetRow + 0.65}
\pgfmathsetmacro{\yBottom}{\yCalendarRow - 0.45}

\draw[decorate, decoration={calligraphic brace, amplitude=6pt, raise=2pt},
      line width=0.7pt, blue]
    (0, \yTop) -- (\xForecastOrigin, \yTop)
    node[midway, above=8pt, font=\footnotesize, blue] {Context $C$ (past)};
\draw[decorate, decoration={calligraphic brace, amplitude=6pt, raise=2pt},
      line width=0.7pt, red]
    (\xForecastOrigin, \yTop) -- (\xEnd, \yTop)
    node[midway, above=8pt, font=\footnotesize, red] {Horizon $H$ (future)};

\node[block, fill=colorTargetPast, minimum width=\xForecastOrigin cm, anchor=west]
    at (0, \yTargetRow) {};
\node[lbl] at ({\xForecastOrigin/2}, \yTargetRow) {Measured target};
\pgfmathsetmacro{\widthHorizonBlock}{\xEnd - \xForecastOrigin}
\node[block, fill=colorTargetForecast, dashed, minimum width=\widthHorizonBlock cm, anchor=west]
    at (\xForecastOrigin, \yTargetRow) {};
\node[lbl] at ({(\xForecastOrigin+\xEnd)/2}, \yTargetRow) {Forecast (output)};
\node[lbl, anchor=east, align=right] at (-0.15, \yTargetRow) {Target};

\node[lbl, anchor=west] at (0, \yRunLabel)
    {Historical weather forecasts (stitched runs)};

\foreach \i in {0,1,2,3}{
    \pgfmathsetmacro{\xIssue}{\i*\widthRunUsed}
    \pgfmathsetmacro{\yRun}{\yRunFirst - \i*\yRunStep}
    \node[run, fill=colorRunUsed, minimum width=\widthRunUsed cm, anchor=west]
        at (\xIssue, \yRun) {};
    \ifnum\i<3
        \node[run, fill=colorRunUnused, draw=gray!60, minimum width=\widthRunTail cm, anchor=west]
            at ({\xIssue+\widthRunUsed}, \yRun) {};
    \fi
    \node[issuedot] at (\xIssue, {\yRun+0.15}) {};
}

\pgfmathsetmacro{\widthIfsUnused}{\xForecastOrigin - \xIfsIssue}
\node[run, fill=colorIfsUnused, draw=gray!60, minimum width=\widthIfsUnused cm, anchor=west]
    at (\xIfsIssue, \yIfsRow) {};
\pgfmathsetmacro{\widthIfsUsed}{\xEnd - \xForecastOrigin}
\node[run, fill=colorIfsUsed, minimum width=\widthIfsUsed cm, anchor=west]
    at (\xForecastOrigin, \yIfsRow) {};
\node[lbl] at ({(\xForecastOrigin+\xEnd)/2}, \yIfsRow) {Most recent forecast run};
\node[issuedot] at (\xIfsIssue, {\yIfsRow+0.15}) {};
\node[lbl, font=\scriptsize, anchor=east] at ({\xIfsIssue-0.15}, \yIfsRow)
    {issued before $t_0$ (dissemination delay)};

\pgfmathsetmacro{\yWeatherLabel}{(\yRunFirst+\yIfsRow)/2}
\node[lbl, anchor=east, align=right] at (-0.15, \yWeatherLabel) {Weather\\covariates};

\node[block, fill=colorCalendar, minimum width=\xEnd cm, anchor=west]
    at (0, \yCalendarRow) {};
\node[lbl] at ({\xEnd/2}, \yCalendarRow) {Calendar features (known everywhere)};
\node[lbl, anchor=east, align=right] at (-0.15, \yCalendarRow) {Calendar\\covariates};

\draw[dashed, gray, thick] (\xForecastOrigin, {\yBottom+0.15}) -- (\xForecastOrigin, \yTop);

\pgfmathsetmacro{\yTimeAxis}{\yBottom - 0.5}
\draw[->, semithick, gray!80]
    (0, \yTimeAxis) -- ({\xEnd+0.3}, \yTimeAxis)
    node[right, font=\scriptsize, gray] {time};
\draw[gray!80, thin] (\xForecastOrigin, {\yTimeAxis+0.07}) -- (\xForecastOrigin, {\yTimeAxis-0.07});
\node[font=\scriptsize, below] at (\xForecastOrigin, \yTimeAxis) {$t_0$};

\pgfmathsetmacro{\yLegendRow}{\yTimeAxis - 0.85}
\node[issuedot] at (1.1, \yLegendRow) {};
\node[lbl, anchor=west] at (1.2, \yLegendRow) {issue time};
\node[run, fill=colorRunUsed, minimum width=0.42cm, anchor=west] at (3.3, \yLegendRow) {};
\node[lbl, anchor=west] at (3.95, \yLegendRow) {used};
\node[run, fill=colorRunUnused, draw=gray!60, minimum width=0.42cm, anchor=west] at (5.5, \yLegendRow) {};
\node[lbl, anchor=west] at (6.15, \yLegendRow) {superseded};

\end{tikzpicture}
\caption{Leakage-free covariate construction around the forecast origin $t_0$. Each weather forecast run is used from its issue time until superseded by the next run; the horizon is covered by the most recent run available at $t_0$.}
\label{fig:covariate_schema}
\end{figure}

\subsection{Selected Forecasting Models}\label{sec:selected_models}
\label{sec:models}
For our systematic assessment of TSFMs for energy data forecasting, we further select a diverse set of models (Table~\ref{tab:models_overview}). Only models with openly available weights and code are included. We further aim to include TSFMs that differ in their underlying architecture, i.e., transformer-, state-space-, and xLSTM-based approaches. Beyond that, the selected models should support evaluating univariate and covariate-informed settings, as well as task-specific fine-tuning, which is currently a limitation for many approaches. The multivariate mode is excluded due to its limited improvement over the univariate setup, as shown in the Chronos-2 paper by Ansari et al.~\cite{Ansari2025Chronos2} (Fig.~10a, p.~22). Finally, we consider performance on the general-purpose benchmarks GIFT-Eval~\cite{Aksu2024GIFTEVAL} and fev-bench~\cite{Shchur2025FEVBench} as a selection criterion. 

\begin{scriptsize}
\begin{longtable}{@{}p{0.14\textwidth}p{0.15\textwidth}p{0.08\textwidth}p{0.25\textwidth}p{0.26\textwidth}@{}}
\caption{Overview of the forecasting models used in this study, including references and the forecasting modes we employ.} 
\label{tab:models_overview} \\
\toprule
\textbf{Model} & \textbf{Type} & \textbf{Ref.} & \textbf{Mode} & \textbf{Additional Information} \\
\midrule
\endfirsthead

\multicolumn{5}{c}%
{{\tablename\ \thetable{} -- continued from previous page}} \\
\toprule
\textbf{Model} & \textbf{Type} & \textbf{Ref.} & \textbf{Mode} & \textbf{Additional Information} \\
\midrule
\endhead

\midrule
\multicolumn{5}{r}{{Continued on next page}} \\
\endfoot

\bottomrule
\endlastfoot

Chronos-2  & Foundation Model & \cite{Ansari2025Chronos2} & Univariate, Covariate, Fine-tuning & Version 2.2.2 \\
TimesFM & Foundation Model & \cite{Das2024TimesFM} & Univariate, Covariate &  Version 2.0.0, does not yet support fine-tuning \\
TiRex-2 & Foundation Model & \cite{Podest2026TiRex2} & Univariate, Covariate & Version 0.1.1, multivariate successor of the xLSTM-based TiRex~\cite{Auer2025TiRex}, does not yet support fine-tuning; the released checkpoint truncates inputs to the most recent 2{,}048 steps \\
FlowState & Foundation Model & \cite{Graf2025FlowState} & Univariate & Version -r1.1, does not yet support covariates and fine-tuning \\
Toto-2.0 & Foundation Model & \cite{Khwaja2026Toto2} & Univariate & Datadog Toto-2.0-2.5B, does not yet support covariates and fine-tuning \\
TabPFN-TS & Foundation Model & \cite{Grinsztajn2025TabPFN,Hoo2025TabPFNTS} & Covariate & TabPFN-TS with TabPFN-3.0 \cite{Grinsztajn2025TabPFN}, Version 8.0.7 (TabPFN) \\
Random Forest\footnote{Each quantile is fit as a separate random forest with the squared-error criterion (predicting the conditional mean), tuned per quantile; the spread between quantile levels stems only from these hyperparameter differences rather than from a quantile-regression method such as quantile regression forests. Its probabilistic scores should hence be read with caution.} & Bagging Ensemble & \cite{Breiman2001RandomForest} & Training & Version 26.2.0, Built with \textit{cuml-cu12}  \\
XGBoost & Gradient Boosting & \cite{Chen2016XGBoost} & Training & Version 3.1.1, recursive multi-step, \textit{xgboost python package}  \\
XGBoost-MO (multi-output)\footnote{In the standard \textit{xgboost} library, \texttt{multi\_strategy} is incompatible with the quantile objective. XGBoost-MO is therefore trained with squared error, so all predicted quantile levels coincide with the point forecast (no quantile spread); its probabilistic scores are reported for completeness only.} & Gradient Boosting & \cite{Chen2016XGBoost} & Training & Version 3.1.1, same \textit{xgboost} package with the built-in multi-output mode activated (\textit{multi\_strategy} option) \\
\end{longtable}
\end{scriptsize}

The selected models cover the architectural families introduced in Section~\ref{sec:related_work}: the transformer-based Chronos-2~\cite{Ansari2025Chronos2}, TimesFM~\cite{Das2024TimesFM}, TabPFN-TS~\cite{Hoo2025TabPFNTS}, and Toto-2.0~\cite{Khwaja2026Toto2}, the state-space model FlowState~\cite{Graf2025FlowState}, and the xLSTM-based TiRex-2~\cite{Podest2026TiRex2}. While FlowState and Toto-2.0 do not yet support covariates and are therefore evaluated only in the univariate setting, TimesFM, Chronos-2, and TiRex-2 are evaluated in both univariate and covariate-informed settings, whereas TabPFN-TS is evaluated only in covariate-informed mode. To assess the potential benefits of a domain-specific adaptation, we additionally fine-tune Chronos-2 on the respective training datasets, which the other architectures currently do not support at all or not open-source.
Chronos-2 is fine-tuned using the default hyperparameters given in the Chronos-2 technical documentation~\cite{Ansari2025Chronos2}, listed in Appendix Table~\ref{tab:model_specifications}, on the same fixed evaluation block as used in the univariate and covariate-informed modes. All remaining time steps of each dataset constitute the training block, whose length therefore varies per dataset. The training block is segmented into sliding windows of identical context length $C$ and forecast horizon $H$ for each experimental setting, with a 75\% overlap between windows.

A main question of our study is how TSFMs perform compared to state-of-the-art models in energy forecasting, which is why we include the machine-learning-based approaches XGBoost and random forest -- the dominant task-specific baselines in the energy domain as outlined in Section~\ref{sec:related_work}. We therefore focus the task-specific baselines on these tree ensembles, given their competition track record and low tuning requirements. Task-specific deep-learning and statistical architectures are not part of our experimental design; their performance is assessed by the general-purpose benchmarks discussed in Section~\ref{sec:related_work}, while deep learning itself remains part of our comparison through the TSFMs, which advance deep architectures beyond the per-series training considered here, via global training across many series, to large-scale pretraining on millions of series. Both are trained task-specifically, employing the subsequent pipeline:
\begin{enumerate}
    \item \textbf{Automatic lag selection}: Generation of candidate lagged features and selection of the most informative lags based on SHAP importance scores, a selection mechanism shown to outperform common feature-selection methods for gradient-boosted trees~\cite{Marcilio2020SHAPFS}.
    \item \textbf{Feature selection}: Retention of the top 50 features from lagged features and engineered features using SHAP values to reduce dimensionality,  prevent overfitting, and provide computational benefits.
    \item \textbf{Hyperparameter optimization}: Bayesian optimization with Optuna~\cite{Akiba2019Optuna}, using 250 trials per quantile and validation performance in terms of the mean squared error for the median and the pinball loss for the other quantiles as objective, a combination also employed in a top-ranked HEFTCom2024 solution~\cite{Pu2025HybridStrategy}. Appendix Table~\ref{tab:model_specifications} lists the hyperparameters considered and their ranges.
    \item \textbf{Final retraining}: Training of the final model on the full training set using the selected features and optimized hyperparameters.
    \item \textbf{Evaluation}: Evaluation employing recursive multi-step-ahead forecasting without retraining, as frequent retraining is often not required to maintain comparable forecast accuracy while reducing computational costs~\cite{Zanotti2025Retraining}, although distribution shifts may require model updates~\cite{Ditzler2015Nonstationary,Haselbeck2021EVARS}.
\end{enumerate}

Similar to fine-tuning Chronos-2, XGBoost and random forest are trained on the same training block using the hyperparameters (Appendix Table~\ref{tab:model_specifications}) and the evaluation setup (Figure~\ref{fig:rolling_evaluation}). For hyperparameter optimization, a subset of at most 5{,}000 time steps is drawn from the end of the training block, which is further split into a training (80\%) and a validation set (20\%). The final models are then retrained using the entire training block, with lag features constructed from a context window length $C$ to ensure comparability with the TSFMs. 

The complete benchmark pipeline -- including data loading, preprocessing, model wrappers, hyperparameter search, and evaluation -- is publicly available.\footnote{\textbf{Code available at:} \href{https://github.com/OberMarco/Energy_Benchmark_TSFM_pub}{github.com/OberMarco/Energy\_Benchmark\_TSFM\_pub}, archived on Zenodo: \href{https://doi.org/10.5281/zenodo.19808931}{10.5281/zenodo.19808931}~\cite{obermeier_code_fets_2026}.} Hardware details and computational-cost measurements are reported in \ref{sec:app_compute}.
\clearpage
\subsection{Evaluation}
\label{sec:eval}
\input{A13_schema_eval_train_split}

Figure~\ref{fig:rolling_evaluation} illustrates the rolling evaluation scheme applied uniformly across all datasets and model types: each dataset is split into a training/fine-tuning block of dataset-dependent length and a fixed evaluation block, on which all models are evaluated under identical conditions. All metrics are computed on each evaluation window and aggregated across windows.

We evaluate the forecasting performance using the Normalized Root Mean Squared Error (NRMSE), a standard metric in energy forecasting that enables comparison across time series with different scales and units. Using the mean absolute value of the observations \(\overline{|y|}\) defined as

\[
\overline{|y|} = \frac{1}{n}\sum_{i=1}^{n}|y_i|,
\]

where \(y_i\) denotes the observed values and \(n\) the number of time steps, the NRMSE for the predicted values \(\hat{y}_i\) can be determined as

\[
\text{NRMSE} = \frac{\;1\;}{\;\overline{|y|}\;} \sqrt{\frac{1}{n}\sum_{i=1}^{n}(y_i - \hat{y}_i)^2}.
\]

We normalize using the mean absolute value, since this is more robust for approximately stationary signals with near-zero means. Such time series, for instance, occur in battery storage dispatch profiles where charging and discharging power approximately cancel out over operating cycles. Furthermore, this normalization allows for an approximate interpretation as a percentage error, making NRMSE particularly intuitive: an NRMSE of 0.05 corresponds to roughly a 5\% deviation from typical signal levels. Further metrics, covering aspects not captured by the NRMSE, are summarized in Appendix Table~\ref{tab:metric_overview}.
\newline

To assess the model's performance across our diverse collection of datasets, we also aim to quantify the intrinsic predictability of each time series, independent of model choice. Analogous to Shannon's 1951 finding~\cite{Shannon1951Entropy} that English is predictable due to its low entropy (resulting from non-random letter frequencies), spectral entropy may quantify time series predictability by measuring deviation from uniform frequency distributions and therefore periodicity. We employ the spectral entropy-based Forecastability Index \(\phi\)~\cite{Kang2017Visualising}, a widely used metric for determining time series complexity, defined as

\[
\phi = 1 - \frac{\mathcal{H}}{\mathcal{H}_{\max}} \quad \text{with} \quad \mathcal{H} = -\sum_{k=1}^{m} p_k \log(p_k),
\]

where \(p_k = |F_k|^2 / \sum_{j=1}^{m}|F_j|^2\) represents the normalized power at frequency \(k\),  \(F_k\)~are the real Fast Fourier Transform (FFT) coefficients of the mean-centered series, and \(\mathcal{H}_{\max} = \log(m)\) is the maximum spectral entropy for \(m\) frequency components. Values range from 0 (random, unforecastable) to 1 (highly periodic, easily forecastable).
\newline

To assess whether performance differences between models are statistically significant, we follow the standard procedure for comparing multiple models over multiple datasets~\cite{Demsar2006Statistical}. We rank the models per (dataset, target) series by their median NRMSE, apply the Friedman test~\cite{Friedman1940Comparison} to the resulting ranks ($\alpha=0.05$), and assess pairwise differences with the Nemenyi post-hoc test~\cite{Nemenyi1963Distribution}: two models differ significantly if their average ranks deviate by more than the critical difference ($\mathrm{CD}$).

Our benchmark study is configured along three key dimensions:

\begin{description}
    \item[\faCog\ Context Length]
    Three context lengths are evaluated: 672, 2{,}000, and 8{,}000 steps. The minimum of 672~steps corresponds to one week at a 15-minute resolution -- the most prevalent resolution across the dataset collection. The maximum of 8{,}000~steps balances the upper context limits of TimesFM-2.5 (16{,}384~steps) and FlowState (${\approx}$16{,}000~steps at \texttt{scale\_factor}$\,{=}\,0.25$) against the shorter windows of Chronos-2 (8{,}192~steps) and TiRex-2 (2{,}048~steps)\footnote{This maximum is a checkpoint default (\texttt{context\_len} in the released model configuration) rather than an architectural limit -- the model is post-trained with longer contexts and, owing to its recurrent architecture, also supports streaming inference as new observations arrive; the accompanying ablation reports only a marginal average effect of this truncation~\cite{Podest2026TiRex2}.}; Chronos-2 is thus evaluated near its maximum. FlowState's SSM-based encoder has no architectural context limit and is bounded only by memory; its pretrained checkpoint defaults to 4{,}096~steps. TabPFN-TS is not constrained by a fixed context window but operates on the underlying TabPFN-3.0 model, for which a recommended maximum of 50{,}000~samples applies; both are evaluated at the same lengths for comparability. The intermediate length of 2{,}000~steps was identified as a favorable trade-off in a preliminary sensitivity analysis and approximates the maximum context length of TiRex-2.

    \item[\faCog\ Forecast Horizon]
        Horizons of 96, 192, and 288~steps are used, corresponding to
        1, 2, and 3~days at 15-minute resolution. These intervals cover
        typical short-term forecasting periods relevant to operational
        energy systems planning.

    \item[\faCog\ Rolling Windows]
        Evaluation is performed over 35 rolling windows with a step size
        of 132~steps. The number of windows reflects a trade-off between
        the length of the shortest dataset and total computation time,
        while ensuring coverage of diverse temporal conditions including
        varying times of hour, day, weekdays, and months. Full seasonal cycles
        cannot be adequately represented due to the limited length of
        several datasets. For a few longer series spanning a full year, selected
        across all data categories, Figure~\ref{fig:seasonality} shows that the
        season has no notable effect on the overall results, confirming that the
        evaluation period is representative.
\end{description}

\clearpage
\section{Results}
\label{sec:results}
We present our results in three parts: (1) an aggregated overview across all data categories and model configurations, (2) an analysis of the relationship between forecastability and prediction performance, and (3) a sensitivity analysis of the two top-performing models investigating the influence of context length, forecast horizon, and aggregation level. Detailed per-dataset results are provided in Appendix Tables~\ref{tab:univariate_benchmark}, \ref{tab:covariate_benchmark}, and~\ref{tab:training_benchmark}. Additionally, \ref{sec:app_compute} reports the computational costs: the TSFMs require hardware comparable to the tree-based baselines and are in part even faster in inference and training.

\subsection{Results Overview}
\label{sec:results:overview}

Table~\ref{tab:combined_benchmark_reduced} presents the aggregated results of all benchmark experiments. Figure~\ref{fig:results_overview} illustrates the underlying aggregation pipeline: experiment-level NRMSE values are aggregated per (dataset, target), per data category, and globally across all experiments to form the overall and best-count rows of Table~\ref{tab:combined_benchmark_reduced}.
\begin{figure}[htbp]
  \centering
\begin{adjustbox}{width=0.98\columnwidth, center}
\begin{tikzpicture}[
  node distance=0.5cm and 0.4cm,
  box/.style={
      rounded corners=2pt, draw=gray!60,
      minimum height=0.65cm, align=center,
      inner sep=2.5pt, font=\scriptsize
  },
  input/.style={box, fill=blue!15,   text width=1.7cm},
  exp/.style={  box, fill=orange!25, text width=2.8cm, font=\scriptsize\bfseries},
  agg/.style={  box, fill=green!25,  text width=4.2cm},
  outbox/.style={box, fill=gray!25,  text width=3.0cm, font=\scriptsize\bfseries},
  lbl/.style={font={\fontsize{6}{7}\selectfont\itshape}, text=gray!70!black},
  ref/.style={font={\fontsize{6}{7}\selectfont\itshape}, text=gray!70!black, align=left},
  >=stealth, line width=0.5pt,
]
\node[input] (m)    {Model};
\node[input, right=of m]    (mode) {Mode\\\fontsize{5.5}{6.5}\selectfont univariate / covariate / training};
\node[input, right=of mode] (d)    {Dataset $\times$ Target\\\fontsize{5.5}{6.5}\selectfont 54 series};
\node[input, right=of d]    (ctx)  {Context Length\\\fontsize{5.5}{6.5}\selectfont $\{672, 2000, 8000\}$};
\node[input, right=of ctx]  (hor)  {Horizon\\\fontsize{5.5}{6.5}\selectfont $\{96, 192, 288\}$};
\node[input, right=of hor]  (roll) {Rolling Windows\\\fontsize{5.5}{6.5}\selectfont $\approx 35$ shifts};

\node[exp, below=0.9cm of ctx, xshift=-0.5cm] (single)
    {Single Experiment\\\fontsize{5.5}{6.5}\selectfont\mdseries one NRMSE value};

\coordinate (bus) at ([yshift=-0.4cm]$(m.south)!0.5!(roll.south)$);
\foreach \n in {m,mode,d,ctx,hor,roll}
    \draw (\n.south) -- (\n.south |- bus);
\draw (m.south |- bus) -- (roll.south |- bus);
\draw[->] (single.north |- bus) -- (single.north);

\node[lbl, right=0.15cm of single, align=left]
    {$17{,}010$\\experiments per mode};

\node[agg, below=0.55cm of single] (ds)
    {Aggregate over rolling windows, context length, horizon\\
     \fontsize{5.5}{6.5}\selectfont per model, per (dataset, target)};
\draw[->] (single) -- (ds) node[midway, right, lbl] {Min, Q25, median, Q75, Max};

\node[ref, right=0.6cm of ds]
    {$\Rightarrow$ Tables~\ref{tab:univariate_benchmark},\\
     \ref{tab:covariate_benchmark},\\
     \ref{tab:training_benchmark}};

\node[agg, below=0.4cm of ds] (cat)
    {Aggregate over datasets within category\\
     \fontsize{5.5}{6.5}\selectfont per model, per data category};
\draw[->] (ds) -- (cat) node[midway, right, lbl] {median, (Q10, Q90)};

\node[ref, right=0.6cm of cat]
    {$\Rightarrow$ Table~\ref{tab:combined_benchmark_reduced}};

\node[outbox, below left=0.7cm and -0.5cm of cat]  (overall)
    {Overall Row\\\fontsize{5.5}{6.5}\selectfont\mdseries median across all experiments};
\node[outbox, below right=0.7cm and -0.5cm of cat] (count)
    {Best Count Row\\\fontsize{5.5}{6.5}\selectfont\mdseries wins per (dataset, target, context, horizon)};

\coordinate (split) at ([yshift=-0.3cm]cat.south);
\draw (cat.south) -- (split);
\draw[->] (split) -| (overall.north);
\draw[->] (split) -| (count.north);

\end{tikzpicture}
\end{adjustbox}
\vspace{-0.3cm}
\caption{Aggregation pipeline of the benchmark results, from individual experiments to the per-category, overall, and best-count rows of Table~\ref{tab:combined_benchmark_reduced}.}
\label{fig:results_overview}
\end{figure}

Each data-category column reports the median NRMSE with its (Q10, Q90) interval across all experiments within that category; the \textit{Overall} column aggregates across all experiments. The experiments cover different model modes and forecasting models as described in Section~\ref{sec:models}. The \textit{Best Count} column indicates the number of times a model achieves the best performance for a given experimental setting, defined by identical horizon length, context
length, and prediction target, while varying rolling windows and covariate configurations. Tied experiments and experiments with missing results are excluded from this count (note below each table). An orange background marks the best model within each benchmark mode, and a green background marks the overall best performer across all modes in each column.

We observe that covariate-informed approaches achieve the lowest overall median NRMSEs across all three benchmark modes across all 17{,}010 experiments (Table~\ref{tab:combined_benchmark_reduced}). Chronos-2 (covariate) achieves the lowest overall median NRMSE with 0.472, although the tree-based comparison partners were trained on the full history of each target and its covariates and benefited from SHAP-based feature selection and hyperparameter tuning, while the zero-shot models only had access to the unfiltered covariate candidates within their limited context window. Furthermore, the performance differences in terms of NRMSE are statistically significant: the Friedman test across all $N=54$ (dataset, target) series rejects equal model performance ($\chi^2=176.4$, $p=2.3\times10^{-31}$), and the Nemenyi post-hoc analysis given in Appendix Table~\ref{tab:significance_nrmse} assigns Chronos-2 (covariate) the best average rank of 3.63 as the only model that belongs exclusively to the top homogeneous group. It significantly outperforms all task-specifically trained machine-learning approaches, including XGBoost MO (8.82), XGBoost (9.56), and random forest (10.84), as well as TabPFN-TS (covariate), while not differing significantly from Chronos-2 (univariate), TiRex-2 (covariate), and FlowState (univariate). Computing the ranking using the normalized Continuous Ranked Probability Score (nCRPS) confirms this observation (Appendix Table~\ref{tab:significance_nrmse}, right block): Chronos-2 (covariate) again achieves the best average rank (3.56, $\chi^2=361.2$) as the only model belonging exclusively to the top homogeneous group.

Figure~\ref{fig:pairwise_rel_improvement} further compares the two best-performing models by median NRMSE, Chronos-2 (covariate) and TiRex-2 (covariate), highlighting their complementary strengths. The two models are separated by less than 0.5\% in overall median NRMSE (0.472 vs.\ 0.474). Chronos-2 (covariate) achieves a higher overall win rate across all paired experiments (57\% vs.\ 43\%) and shows the lower median NRMSE in seven of the nine data categories, while TiRex-2 (covariate) is slightly ahead for Balancing Services and Non-Dispatchable Generation.

Appendix Tables~\ref{tab:combined_benchmark_mase}--\ref{tab:combined_benchmark_picp} show similarly aggregated results for the further evaluation metrics, while Appendix Table~\ref{tab:combined_benchmark_settings} provides more detailed results for the nine forecast-horizon and context-length settings. Across these further metrics, the foundation models remain the leading group in all but one -- only the sum ratio favors the task-specifically tuned XGBoost. A seasonal analysis (Figure~\ref{fig:seasonality}) further shows that it holds across the year and is not specific to any individual season.

\begin{table}[htbp]
\centering
\caption{Results overview across all benchmark modes (NRMSE). Cells show the median NRMSE with (Q10, Q90) interval over all experiments in the respective data category. The Overall Median column is computed over each model's pooled set of all 17,010 experiments ((Dataset, Target) $\times$ Rolling Windows $\times$ Horizon $\times$ Context Length); Best Count shows wins across all benchmarks.}
\small
\renewcommand{\arraystretch}{1.5}
\begin{adjustbox}{max width=\textwidth}
\begin{tabular}{>{\raggedright\arraybackslash}m{2.6cm} c c c c c c c c c !{\vrule width 1pt} c | c}
\toprule
\textbf{Model} & \textbf{\shortstack{Balancing\\Services}} & \textbf{\shortstack{Dispatchable\\Generation}} & \textbf{\shortstack{Grid\\Data}} & \textbf{\shortstack{Heat\\Data}} & \textbf{Load} & \textbf{\shortstack{Market\\Data}} & \textbf{\shortstack{Mobility\\Data}} & \textbf{\shortstack{Non-Dispatchable\\Generation}} & \textbf{\shortstack{Residential\\Load}} & \shortstack{\textbf{Overall}\\\textbf{Median}} & \shortstack{\textbf{Best}\\\textbf{Count}} \\
\midrule
\multicolumn{12}{l}{\textit{Univariate}} \\
Toto-2.0 & \cellcolor{green!15}$\substack{\rule{0pt}{10pt}\textstyle\mathbf{2.42} \\[1pt] \scriptstyle\langle 1.25,\,4.73e{6}\rangle\rule[-5pt]{0pt}{5pt}}$ & \cellcolor{orange!10}$\substack{\rule{0pt}{10pt}\textstyle\mathbf{0.391} \\[1pt] \scriptstyle\langle 0.171,\,1.03\rangle\rule[-5pt]{0pt}{5pt}}$ & $\substack{\rule{0pt}{10pt}\textstyle 0.451 \\[1pt] \scriptstyle\langle 0.107,\,1.21\rangle\rule[-5pt]{0pt}{5pt}}$ & $\substack{\rule{0pt}{10pt}\textstyle 0.786 \\[1pt] \scriptstyle\langle 0.133,\,1.15\rangle\rule[-5pt]{0pt}{5pt}}$ & \cellcolor{green!15}$\substack{\rule{0pt}{10pt}\textstyle\mathbf{0.036} \\[1pt] \scriptstyle\langle 0.014,\,0.340\rangle\rule[-5pt]{0pt}{5pt}}$ & $\substack{\rule{0pt}{10pt}\textstyle 0.329 \\[1pt] \scriptstyle\langle 0.111,\,1.01\rangle\rule[-5pt]{0pt}{5pt}}$ & $\substack{\rule{0pt}{10pt}\textstyle 0.734 \\[1pt] \scriptstyle\langle 0.237,\,1.47\rangle\rule[-5pt]{0pt}{5pt}}$ & \cellcolor{orange!10}$\substack{\rule{0pt}{10pt}\textstyle\mathbf{0.536} \\[1pt] \scriptstyle\langle 0.177,\,1.42\rangle\rule[-5pt]{0pt}{5pt}}$ & $\substack{\rule{0pt}{10pt}\textstyle 0.800 \\[1pt] \scriptstyle\langle 0.372,\,1.35\rangle\rule[-5pt]{0pt}{5pt}}$ & \cellcolor{orange!10}$\substack{\rule{0pt}{10pt}\textstyle\mathbf{0.564} \\[1pt] \scriptstyle\langle 0.252,\,1.01\rangle\rule[-5pt]{0pt}{5pt}}$ & \textbf{1,264 (7\%)} \\
TimesFM & $\substack{\rule{0pt}{10pt}\textstyle 2.42 \\[1pt] \scriptstyle\langle 1.25,\,3.31e{7}\rangle\rule[-5pt]{0pt}{5pt}}$ & $\substack{\rule{0pt}{10pt}\textstyle 0.409 \\[1pt] \scriptstyle\langle 0.171,\,1.08\rangle\rule[-5pt]{0pt}{5pt}}$ & $\substack{\rule{0pt}{10pt}\textstyle 0.476 \\[1pt] \scriptstyle\langle 0.100,\,1.19\rangle\rule[-5pt]{0pt}{5pt}}$ & $\substack{\rule{0pt}{10pt}\textstyle 0.785 \\[1pt] \scriptstyle\langle 0.126,\,1.05\rangle\rule[-5pt]{0pt}{5pt}}$ & $\substack{\rule{0pt}{10pt}\textstyle 0.037 \\[1pt] \scriptstyle\langle 0.015,\,0.273\rangle\rule[-5pt]{0pt}{5pt}}$ & $\substack{\rule{0pt}{10pt}\textstyle 0.334 \\[1pt] \scriptstyle\langle 0.113,\,0.954\rangle\rule[-5pt]{0pt}{5pt}}$ & $\substack{\rule{0pt}{10pt}\textstyle 0.749 \\[1pt] \scriptstyle\langle 0.251,\,1.50\rangle\rule[-5pt]{0pt}{5pt}}$ & $\substack{\rule{0pt}{10pt}\textstyle 0.566 \\[1pt] \scriptstyle\langle 0.208,\,1.42\rangle\rule[-5pt]{0pt}{5pt}}$ & $\substack{\rule{0pt}{10pt}\textstyle 0.792 \\[1pt] \scriptstyle\langle 0.360,\,1.31\rangle\rule[-5pt]{0pt}{5pt}}$ & $\substack{\rule{0pt}{10pt}\textstyle 0.587 \\[1pt] \scriptstyle\langle 0.255,\,0.986\rangle\rule[-5pt]{0pt}{5pt}}$ & \textbf{1,000 (6\%)} \\
Chronos-2 & $\substack{\rule{0pt}{10pt}\textstyle 2.45 \\[1pt] \scriptstyle\langle 1.25,\,3.23e{7}\rangle\rule[-5pt]{0pt}{5pt}}$ & $\substack{\rule{0pt}{10pt}\textstyle 0.404 \\[1pt] \scriptstyle\langle 0.180,\,1.03\rangle\rule[-5pt]{0pt}{5pt}}$ & \cellcolor{orange!10}$\substack{\rule{0pt}{10pt}\textstyle\mathbf{0.441} \\[1pt] \scriptstyle\langle 0.105,\,1.20\rangle\rule[-5pt]{0pt}{5pt}}$ & \cellcolor{orange!10}$\substack{\rule{0pt}{10pt}\textstyle\mathbf{0.761} \\[1pt] \scriptstyle\langle 0.125,\,1.07\rangle\rule[-5pt]{0pt}{5pt}}$ & $\substack{\rule{0pt}{10pt}\textstyle 0.041 \\[1pt] \scriptstyle\langle 0.014,\,0.249\rangle\rule[-5pt]{0pt}{5pt}}$ & $\substack{\rule{0pt}{10pt}\textstyle 0.331 \\[1pt] \scriptstyle\langle 0.114,\,0.952\rangle\rule[-5pt]{0pt}{5pt}}$ & $\substack{\rule{0pt}{10pt}\textstyle 0.748 \\[1pt] \scriptstyle\langle 0.244,\,1.51\rangle\rule[-5pt]{0pt}{5pt}}$ & $\substack{\rule{0pt}{10pt}\textstyle 0.562 \\[1pt] \scriptstyle\langle 0.181,\,1.46\rangle\rule[-5pt]{0pt}{5pt}}$ & $\substack{\rule{0pt}{10pt}\textstyle 0.785 \\[1pt] \scriptstyle\langle 0.346,\,1.30\rangle\rule[-5pt]{0pt}{5pt}}$ & $\substack{\rule{0pt}{10pt}\textstyle 0.573 \\[1pt] \scriptstyle\langle 0.244,\,0.983\rangle\rule[-5pt]{0pt}{5pt}}$ & \textbf{910 (5\%)} \\
FlowState & $\substack{\rule{0pt}{10pt}\textstyle 2.44 \\[1pt] \scriptstyle\langle 1.26,\,7.95e{7}\rangle\rule[-5pt]{0pt}{5pt}}$ & $\substack{\rule{0pt}{10pt}\textstyle 0.523 \\[1pt] \scriptstyle\langle 0.213,\,1.32\rangle\rule[-5pt]{0pt}{5pt}}$ & $\substack{\rule{0pt}{10pt}\textstyle 0.471 \\[1pt] \scriptstyle\langle 0.096,\,1.19\rangle\rule[-5pt]{0pt}{5pt}}$ & $\substack{\rule{0pt}{10pt}\textstyle 0.774 \\[1pt] \scriptstyle\langle 0.132,\,1.14\rangle\rule[-5pt]{0pt}{5pt}}$ & $\substack{\rule{0pt}{10pt}\textstyle 0.050 \\[1pt] \scriptstyle\langle 0.017,\,0.261\rangle\rule[-5pt]{0pt}{5pt}}$ & $\substack{\rule{0pt}{10pt}\textstyle 0.365 \\[1pt] \scriptstyle\langle 0.137,\,1.06\rangle\rule[-5pt]{0pt}{5pt}}$ & \cellcolor{green!15}$\substack{\rule{0pt}{10pt}\textstyle\mathbf{0.694} \\[1pt] \scriptstyle\langle 0.240,\,1.29\rangle\rule[-5pt]{0pt}{5pt}}$ & $\substack{\rule{0pt}{10pt}\textstyle 0.616 \\[1pt] \scriptstyle\langle 0.234,\,1.67\rangle\rule[-5pt]{0pt}{5pt}}$ & \cellcolor{green!15}$\substack{\rule{0pt}{10pt}\textstyle\mathbf{0.775} \\[1pt] \scriptstyle\langle 0.345,\,1.27\rangle\rule[-5pt]{0pt}{5pt}}$ & $\substack{\rule{0pt}{10pt}\textstyle 0.611 \\[1pt] \scriptstyle\langle 0.265,\,1.04\rangle\rule[-5pt]{0pt}{5pt}}$ & \cellcolor{orange!10}\textbf{1,453 (9\%)} \\
TiRex-2 & $\substack{\rule{0pt}{10pt}\textstyle 2.46 \\[1pt] \scriptstyle\langle 1.25,\,2.29e{7}\rangle\rule[-5pt]{0pt}{5pt}}$ & $\substack{\rule{0pt}{10pt}\textstyle 0.402 \\[1pt] \scriptstyle\langle 0.181,\,1.10\rangle\rule[-5pt]{0pt}{5pt}}$ & $\substack{\rule{0pt}{10pt}\textstyle 0.481 \\[1pt] \scriptstyle\langle 0.105,\,1.19\rangle\rule[-5pt]{0pt}{5pt}}$ & $\substack{\rule{0pt}{10pt}\textstyle 0.776 \\[1pt] \scriptstyle\langle 0.129,\,1.08\rangle\rule[-5pt]{0pt}{5pt}}$ & $\substack{\rule{0pt}{10pt}\textstyle 0.043 \\[1pt] \scriptstyle\langle 0.017,\,0.227\rangle\rule[-5pt]{0pt}{5pt}}$ & \cellcolor{orange!10}$\substack{\rule{0pt}{10pt}\textstyle\mathbf{0.328} \\[1pt] \scriptstyle\langle 0.121,\,0.998\rangle\rule[-5pt]{0pt}{5pt}}$ & $\substack{\rule{0pt}{10pt}\textstyle 0.781 \\[1pt] \scriptstyle\langle 0.260,\,1.52\rangle\rule[-5pt]{0pt}{5pt}}$ & $\substack{\rule{0pt}{10pt}\textstyle 0.558 \\[1pt] \scriptstyle\langle 0.196,\,1.44\rangle\rule[-5pt]{0pt}{5pt}}$ & $\substack{\rule{0pt}{10pt}\textstyle 0.794 \\[1pt] \scriptstyle\langle 0.354,\,1.33\rangle\rule[-5pt]{0pt}{5pt}}$ & $\substack{\rule{0pt}{10pt}\textstyle 0.586 \\[1pt] \scriptstyle\langle 0.253,\,1.01\rangle\rule[-5pt]{0pt}{5pt}}$ & \textbf{652 (4\%)} \\
\midrule
\multicolumn{12}{l}{\textit{Covariate}} \\
Chronos-2 & $\substack{\rule{0pt}{10pt}\textstyle 2.45 \\[1pt] \scriptstyle\langle 1.25,\,4.84e{7}\rangle\rule[-5pt]{0pt}{5pt}}$ & \cellcolor{green!15}$\substack{\rule{0pt}{10pt}\textstyle\mathbf{0.355} \\[1pt] \scriptstyle\langle 0.172,\,0.933\rangle\rule[-5pt]{0pt}{5pt}}$ & \cellcolor{green!15}$\substack{\rule{0pt}{10pt}\textstyle\mathbf{0.379} \\[1pt] \scriptstyle\langle 0.102,\,1.08\rangle\rule[-5pt]{0pt}{5pt}}$ & \cellcolor{orange!10}$\substack{\rule{0pt}{10pt}\textstyle\mathbf{0.762} \\[1pt] \scriptstyle\langle 0.111,\,1.08\rangle\rule[-5pt]{0pt}{5pt}}$ & \cellcolor{orange!10}$\substack{\rule{0pt}{10pt}\textstyle\mathbf{0.038} \\[1pt] \scriptstyle\langle 0.015,\,0.287\rangle\rule[-5pt]{0pt}{5pt}}$ & \cellcolor{green!15}$\substack{\rule{0pt}{10pt}\textstyle\mathbf{0.308} \\[1pt] \scriptstyle\langle 0.110,\,0.908\rangle\rule[-5pt]{0pt}{5pt}}$ & \cellcolor{orange!10}$\substack{\rule{0pt}{10pt}\textstyle\mathbf{0.740} \\[1pt] \scriptstyle\langle 0.240,\,1.55\rangle\rule[-5pt]{0pt}{5pt}}$ & $\substack{\rule{0pt}{10pt}\textstyle 0.347 \\[1pt] \scriptstyle\langle 0.132,\,0.976\rangle\rule[-5pt]{0pt}{5pt}}$ & \cellcolor{orange!10}$\substack{\rule{0pt}{10pt}\textstyle\mathbf{0.791} \\[1pt] \scriptstyle\langle 0.351,\,1.30\rangle\rule[-5pt]{0pt}{5pt}}$ & \cellcolor{green!15}$\substack{\rule{0pt}{10pt}\textstyle\mathbf{0.472} \\[1pt] \scriptstyle\langle 0.218,\,0.902\rangle\rule[-5pt]{0pt}{5pt}}$ & \cellcolor{orange!10}\textbf{1,942 (11\%)} \\
TabPFN & $\substack{\rule{0pt}{10pt}\textstyle 2.47 \\[1pt] \scriptstyle\langle 1.25,\,1.47e{7}\rangle\rule[-5pt]{0pt}{5pt}}$ & $\substack{\rule{0pt}{10pt}\textstyle 0.408 \\[1pt] \scriptstyle\langle 0.207,\,1.04\rangle\rule[-5pt]{0pt}{5pt}}$ & $\substack{\rule{0pt}{10pt}\textstyle 0.397 \\[1pt] \scriptstyle\langle 0.116,\,1.03\rangle\rule[-5pt]{0pt}{5pt}}$ & $\substack{\rule{0pt}{10pt}\textstyle 0.855 \\[1pt] \scriptstyle\langle 0.127,\,1.20\rangle\rule[-5pt]{0pt}{5pt}}$ & $\substack{\rule{0pt}{10pt}\textstyle 0.088 \\[1pt] \scriptstyle\langle 0.029,\,0.284\rangle\rule[-5pt]{0pt}{5pt}}$ & $\substack{\rule{0pt}{10pt}\textstyle 0.335 \\[1pt] \scriptstyle\langle 0.130,\,0.980\rangle\rule[-5pt]{0pt}{5pt}}$ & $\substack{\rule{0pt}{10pt}\textstyle 0.744 \\[1pt] \scriptstyle\langle 0.258,\,1.38\rangle\rule[-5pt]{0pt}{5pt}}$ & $\substack{\rule{0pt}{10pt}\textstyle 0.390 \\[1pt] \scriptstyle\langle 0.186,\,1.06\rangle\rule[-5pt]{0pt}{5pt}}$ & $\substack{\rule{0pt}{10pt}\textstyle 0.808 \\[1pt] \scriptstyle\langle 0.369,\,1.34\rangle\rule[-5pt]{0pt}{5pt}}$ & $\substack{\rule{0pt}{10pt}\textstyle 0.491 \\[1pt] \scriptstyle\langle 0.249,\,0.934\rangle\rule[-5pt]{0pt}{5pt}}$ & \textbf{1,203 (7\%)} \\
TimesFM & $\substack{\rule{0pt}{10pt}\textstyle 3.24 \\[1pt] \scriptstyle\langle 1.41,\,8.69e{9}\rangle\rule[-5pt]{0pt}{5pt}}$ & $\substack{\rule{0pt}{10pt}\textstyle 0.535 \\[1pt] \scriptstyle\langle 0.246,\,1.60\rangle\rule[-5pt]{0pt}{5pt}}$ & $\substack{\rule{0pt}{10pt}\textstyle 0.503 \\[1pt] \scriptstyle\langle 0.134,\,1.35\rangle\rule[-5pt]{0pt}{5pt}}$ & $\substack{\rule{0pt}{10pt}\textstyle 0.812 \\[1pt] \scriptstyle\langle 0.145,\,1.27\rangle\rule[-5pt]{0pt}{5pt}}$ & $\substack{\rule{0pt}{10pt}\textstyle 0.106 \\[1pt] \scriptstyle\langle 0.040,\,0.413\rangle\rule[-5pt]{0pt}{5pt}}$ & $\substack{\rule{0pt}{10pt}\textstyle 0.432 \\[1pt] \scriptstyle\langle 0.165,\,1.49\rangle\rule[-5pt]{0pt}{5pt}}$ & $\substack{\rule{0pt}{10pt}\textstyle 0.878 \\[1pt] \scriptstyle\langle 0.293,\,2.17\rangle\rule[-5pt]{0pt}{5pt}}$ & $\substack{\rule{0pt}{10pt}\textstyle 0.412 \\[1pt] \scriptstyle\langle 0.198,\,1.17\rangle\rule[-5pt]{0pt}{5pt}}$ & $\substack{\rule{0pt}{10pt}\textstyle 0.848 \\[1pt] \scriptstyle\langle 0.387,\,1.70\rangle\rule[-5pt]{0pt}{5pt}}$ & $\substack{\rule{0pt}{10pt}\textstyle 0.593 \\[1pt] \scriptstyle\langle 0.287,\,1.09\rangle\rule[-5pt]{0pt}{5pt}}$ & \textbf{873 (5\%)} \\
TiRex-2 & \cellcolor{orange!10}$\substack{\rule{0pt}{10pt}\textstyle\mathbf{2.43} \\[1pt] \scriptstyle\langle 1.27,\,3.19e{7}\rangle\rule[-5pt]{0pt}{5pt}}$ & $\substack{\rule{0pt}{10pt}\textstyle 0.409 \\[1pt] \scriptstyle\langle 0.187,\,1.06\rangle\rule[-5pt]{0pt}{5pt}}$ & $\substack{\rule{0pt}{10pt}\textstyle 0.391 \\[1pt] \scriptstyle\langle 0.110,\,1.07\rangle\rule[-5pt]{0pt}{5pt}}$ & $\substack{\rule{0pt}{10pt}\textstyle 0.778 \\[1pt] \scriptstyle\langle 0.111,\,1.07\rangle\rule[-5pt]{0pt}{5pt}}$ & $\substack{\rule{0pt}{10pt}\textstyle 0.058 \\[1pt] \scriptstyle\langle 0.019,\,0.229\rangle\rule[-5pt]{0pt}{5pt}}$ & $\substack{\rule{0pt}{10pt}\textstyle 0.323 \\[1pt] \scriptstyle\langle 0.125,\,0.987\rangle\rule[-5pt]{0pt}{5pt}}$ & $\substack{\rule{0pt}{10pt}\textstyle 0.770 \\[1pt] \scriptstyle\langle 0.250,\,1.50\rangle\rule[-5pt]{0pt}{5pt}}$ & \cellcolor{green!15}$\substack{\rule{0pt}{10pt}\textstyle\mathbf{0.334} \\[1pt] \scriptstyle\langle 0.137,\,0.915\rangle\rule[-5pt]{0pt}{5pt}}$ & $\substack{\rule{0pt}{10pt}\textstyle 0.795 \\[1pt] \scriptstyle\langle 0.354,\,1.33\rangle\rule[-5pt]{0pt}{5pt}}$ & $\substack{\rule{0pt}{10pt}\textstyle 0.474 \\[1pt] \scriptstyle\langle 0.223,\,0.910\rangle\rule[-5pt]{0pt}{5pt}}$ & \textbf{1,497 (9\%)} \\
\midrule
\multicolumn{12}{l}{\textit{Training / Fine-tuning}} \\
Chronos-2 & \cellcolor{orange!10}$\substack{\rule{0pt}{10pt}\textstyle\mathbf{2.46} \\[1pt] \scriptstyle\langle 1.22,\,1.53e{8}\rangle\rule[-5pt]{0pt}{5pt}}$ & $\substack{\rule{0pt}{10pt}\textstyle 0.586 \\[1pt] \scriptstyle\langle 0.181,\,2.35\rangle\rule[-5pt]{0pt}{5pt}}$ & \cellcolor{orange!10}$\substack{\rule{0pt}{10pt}\textstyle\mathbf{0.385} \\[1pt] \scriptstyle\langle 0.123,\,1.14\rangle\rule[-5pt]{0pt}{5pt}}$ & \cellcolor{green!15}$\substack{\rule{0pt}{10pt}\textstyle\mathbf{0.727} \\[1pt] \scriptstyle\langle 0.123,\,0.955\rangle\rule[-5pt]{0pt}{5pt}}$ & $\substack{\rule{0pt}{10pt}\textstyle 0.140 \\[1pt] \scriptstyle\langle 0.026,\,0.554\rangle\rule[-5pt]{0pt}{5pt}}$ & \cellcolor{orange!10}$\substack{\rule{0pt}{10pt}\textstyle\mathbf{0.346} \\[1pt] \scriptstyle\langle 0.119,\,1.10\rangle\rule[-5pt]{0pt}{5pt}}$ & $\substack{\rule{0pt}{10pt}\textstyle 0.868 \\[1pt] \scriptstyle\langle 0.220,\,1.96\rangle\rule[-5pt]{0pt}{5pt}}$ & $\substack{\rule{0pt}{10pt}\textstyle 0.565 \\[1pt] \scriptstyle\langle 0.178,\,1.66\rangle\rule[-5pt]{0pt}{5pt}}$ & $\substack{\rule{0pt}{10pt}\textstyle 0.786 \\[1pt] \scriptstyle\langle 0.351,\,1.34\rangle\rule[-5pt]{0pt}{5pt}}$ & \cellcolor{orange!10}$\substack{\rule{0pt}{10pt}\textstyle\mathbf{0.599} \\[1pt] \scriptstyle\langle 0.252,\,1.05\rangle\rule[-5pt]{0pt}{5pt}}$ & \cellcolor{green!15}\textbf{2,995 (18\%)} \\
RandomForest & $\substack{\rule{0pt}{10pt}\textstyle 3.58 \\[1pt] \scriptstyle\langle 1.25,\,1.25e{11}\rangle\rule[-5pt]{0pt}{5pt}}$ & $\substack{\rule{0pt}{10pt}\textstyle 0.711 \\[1pt] \scriptstyle\langle 0.291,\,1.75\rangle\rule[-5pt]{0pt}{5pt}}$ & $\substack{\rule{0pt}{10pt}\textstyle 0.534 \\[1pt] \scriptstyle\langle 0.175,\,1.17\rangle\rule[-5pt]{0pt}{5pt}}$ & $\substack{\rule{0pt}{10pt}\textstyle 0.907 \\[1pt] \scriptstyle\langle 0.141,\,1.27\rangle\rule[-5pt]{0pt}{5pt}}$ & $\substack{\rule{0pt}{10pt}\textstyle 0.108 \\[1pt] \scriptstyle\langle 0.031,\,0.350\rangle\rule[-5pt]{0pt}{5pt}}$ & $\substack{\rule{0pt}{10pt}\textstyle 0.456 \\[1pt] \scriptstyle\langle 0.148,\,1.33\rangle\rule[-5pt]{0pt}{5pt}}$ & $\substack{\rule{0pt}{10pt}\textstyle 0.906 \\[1pt] \scriptstyle\langle 0.332,\,1.92\rangle\rule[-5pt]{0pt}{5pt}}$ & $\substack{\rule{0pt}{10pt}\textstyle 0.603 \\[1pt] \scriptstyle\langle 0.266,\,1.42\rangle\rule[-5pt]{0pt}{5pt}}$ & $\substack{\rule{0pt}{10pt}\textstyle 1.00 \\[1pt] \scriptstyle\langle 0.413,\,1.97\rangle\rule[-5pt]{0pt}{5pt}}$ & $\substack{\rule{0pt}{10pt}\textstyle 0.696 \\[1pt] \scriptstyle\langle 0.339,\,1.17\rangle\rule[-5pt]{0pt}{5pt}}$ & \textbf{557 (3\%)} \\
XGBoost & $\substack{\rule{0pt}{10pt}\textstyle 3.71 \\[1pt] \scriptstyle\langle 1.25,\,1.51e{9}\rangle\rule[-5pt]{0pt}{5pt}}$ & $\substack{\rule{0pt}{10pt}\textstyle 0.494 \\[1pt] \scriptstyle\langle 0.233,\,1.08\rangle\rule[-5pt]{0pt}{5pt}}$ & $\substack{\rule{0pt}{10pt}\textstyle 0.460 \\[1pt] \scriptstyle\langle 0.132,\,1.06\rangle\rule[-5pt]{0pt}{5pt}}$ & $\substack{\rule{0pt}{10pt}\textstyle 0.916 \\[1pt] \scriptstyle\langle 0.123,\,1.37\rangle\rule[-5pt]{0pt}{5pt}}$ & $\substack{\rule{0pt}{10pt}\textstyle 0.063 \\[1pt] \scriptstyle\langle 0.028,\,0.326\rangle\rule[-5pt]{0pt}{5pt}}$ & $\substack{\rule{0pt}{10pt}\textstyle 0.462 \\[1pt] \scriptstyle\langle 0.171,\,1.85\rangle\rule[-5pt]{0pt}{5pt}}$ & \cellcolor{orange!10}$\substack{\rule{0pt}{10pt}\textstyle\mathbf{0.838} \\[1pt] \scriptstyle\langle 0.354,\,1.57\rangle\rule[-5pt]{0pt}{5pt}}$ & \cellcolor{orange!10}$\substack{\rule{0pt}{10pt}\textstyle\mathbf{0.508} \\[1pt] \scriptstyle\langle 0.215,\,1.18\rangle\rule[-5pt]{0pt}{5pt}}$ & $\substack{\rule{0pt}{10pt}\textstyle 0.985 \\[1pt] \scriptstyle\langle 0.402,\,2.20\rangle\rule[-5pt]{0pt}{5pt}}$ & $\substack{\rule{0pt}{10pt}\textstyle 0.611 \\[1pt] \scriptstyle\langle 0.307,\,1.11\rangle\rule[-5pt]{0pt}{5pt}}$ & \textbf{1,047 (6\%)} \\
XGBoost-MO & $\substack{\rule{0pt}{10pt}\textstyle 2.55 \\[1pt] \scriptstyle\langle 1.26,\,3.80e{9}\rangle\rule[-5pt]{0pt}{5pt}}$ & \cellcolor{orange!10}$\substack{\rule{0pt}{10pt}\textstyle\mathbf{0.406} \\[1pt] \scriptstyle\langle 0.210,\,1.00\rangle\rule[-5pt]{0pt}{5pt}}$ & $\substack{\rule{0pt}{10pt}\textstyle 0.510 \\[1pt] \scriptstyle\langle 0.134,\,1.23\rangle\rule[-5pt]{0pt}{5pt}}$ & $\substack{\rule{0pt}{10pt}\textstyle 0.892 \\[1pt] \scriptstyle\langle 0.148,\,1.14\rangle\rule[-5pt]{0pt}{5pt}}$ & \cellcolor{orange!10}$\substack{\rule{0pt}{10pt}\textstyle\mathbf{0.052} \\[1pt] \scriptstyle\langle 0.023,\,0.298\rangle\rule[-5pt]{0pt}{5pt}}$ & $\substack{\rule{0pt}{10pt}\textstyle 0.465 \\[1pt] \scriptstyle\langle 0.205,\,1.36\rangle\rule[-5pt]{0pt}{5pt}}$ & $\substack{\rule{0pt}{10pt}\textstyle 0.845 \\[1pt] \scriptstyle\langle 0.323,\,1.34\rangle\rule[-5pt]{0pt}{5pt}}$ & $\substack{\rule{0pt}{10pt}\textstyle 0.596 \\[1pt] \scriptstyle\langle 0.273,\,1.62\rangle\rule[-5pt]{0pt}{5pt}}$ & \cellcolor{orange!10}$\substack{\rule{0pt}{10pt}\textstyle\mathbf{0.783} \\[1pt] \scriptstyle\langle 0.369,\,1.27\rangle\rule[-5pt]{0pt}{5pt}}$ & $\substack{\rule{0pt}{10pt}\textstyle 0.644 \\[1pt] \scriptstyle\langle 0.327,\,1.07\rangle\rule[-5pt]{0pt}{5pt}}$ & \textbf{1,609 (9\%)} \\
\bottomrule
\end{tabular}
\end{adjustbox}
\vspace{4pt}
\noindent\begin{tikzpicture}
\node[inner sep=3pt, minimum width=1.2em, minimum height=1.2em, fill=orange!10, draw=gray!50] at (0.65\textwidth,0) (A) {};
\node[right=4pt of A, anchor=west] {\small Mode Winner};
\node[inner sep=3pt, minimum width=1.2em, minimum height=1.2em, fill=green!15, draw=gray!50] at (0.9\textwidth,0) (B) {};
\node[right=4pt of B, anchor=west] {\small Category Winner / Overall Winner};
\end{tikzpicture}
\par\vspace{4pt}
\begin{minipage}{\textwidth}\footnotesize\emph{Note:} For Best Count, All = Total - Ties = 17,010 - 8 = 17,002.\end{minipage}
\label{tab:combined_benchmark_reduced}
\end{table}

\begin{figure}[h]
    \centering
    \resizebox{1.01\textwidth}{!}{%
    \input{A_plot_pairwise_rel_improvement.pgf}
    }
    \caption{Pairwise relative NRMSE comparison between TiRex-2 (covariate) and Chronos-2 (covariate). Bars show $(Chronos$-$2 - TiRex$-$2) / Chronos$-$2 \times 100\%$; positive values (amber) favor TiRex-2, negative values (blue) favor Chronos-2. Lines (right axis) show per-category win rates with absolute counts.}
    \label{fig:pairwise_rel_improvement}
\end{figure}

\clearpage
\subsection{Results Analysis}
\label{sec:results_analysis}
We analyze the correlation between the Forecastability Index $\phi$ (Section~\ref{sec:eval}) and the forecasting performance, defined as $1 - NRMSE$, with NRMSE values above 1 clipped to 0. For each (dataset, target) combination and model mode, we computed the median across all results (rolling windows, context lengths, and forecast horizons). Figure~\ref{fig:forecastability_xy} shows Chronos-2 -- the only model that offers all three forecasting modes and also achieves the lowest overall median NRMSE -- and reveals a strong correlation between forecasting performance and forecastability ($r=0.687$, Spearman $\rho=0.700$, both $p<0.001$). Datasets with higher intrinsic stability, such as aggregated country loads, grid loads, and heat demands, exhibit both high forecastability and strong forecast scores, suggesting that aggregation smooths stochastic variability, whereas volatile series such as single residential loads or balancing services cluster in the lower-left region. Surprisingly, some cases with high forecastability show near-zero forecast scores after fine-tuning; possible explanations are discussed in Section~\ref{sec:discussion}. To verify that this relationship is not an artifact of a single model, Figure~\ref{fig:forecastability_two_models} repeats the analysis for the two best-performing configurations, Chronos-2 and TiRex-2, in covariate mode: the correlation is even more pronounced ($r=0.798$, $\rho=0.805$, both $p<0.001$) and nearly identical for both models ($r=0.799$ and $r=0.796$). The relationship thus holds across two independently developed architectures and reflects properties of the data rather than of a specific model. The Forecastability Index $\phi$ may thus serve as an indicator of the predictability of a data category, although not as an absolute measure: a high $\phi$ does not guarantee a correspondingly high forecast score.

\begin{figure}[h]
    \centering
    \resizebox{1.01\textwidth}{!}{%
    \input{A8_plot_forecastability_half_circle}
    }
    \caption{Forecastability Index $\phi$ versus forecast score (1-NRMSE) for Chronos-2 across its three forecasting modes and all experimental settings ($r = 0.687$, $\rho = 0.700$, both $p < 0.001$).}
    \label{fig:forecastability_xy}
\end{figure}

\begin{figure}[h]
    \centering
    \resizebox{1.01\textwidth}{!}{%
    \input{A8_plot_forecastability_two_models}
    }
    \caption{Forecastability Index $\phi$ versus forecast score (1-NRMSE) for Chronos-2 and TiRex-2 in covariate mode ($r = 0.798$, $\rho = 0.805$, both $p < 0.001$).}
    \label{fig:forecastability_two_models}
\end{figure}

\clearpage

\subsection{Sensitivity Analyses}

We further examine the influence of three key parameters: the context length, the forecast horizon, and the data aggregation level. All three analyses cover the two best-performing models, Chronos-2 and TiRex-2 (both covariate-informed). All experiments maintain the same configurations as the main benchmark settings (benchmark overview in Section~\ref{sec:materials and methods}, evaluation scheme in Section~\ref{sec:eval}) to ensure consistency and comparability.

\begin{figure}[h]
    \centering
    \includegraphics[width=1.01\textwidth]{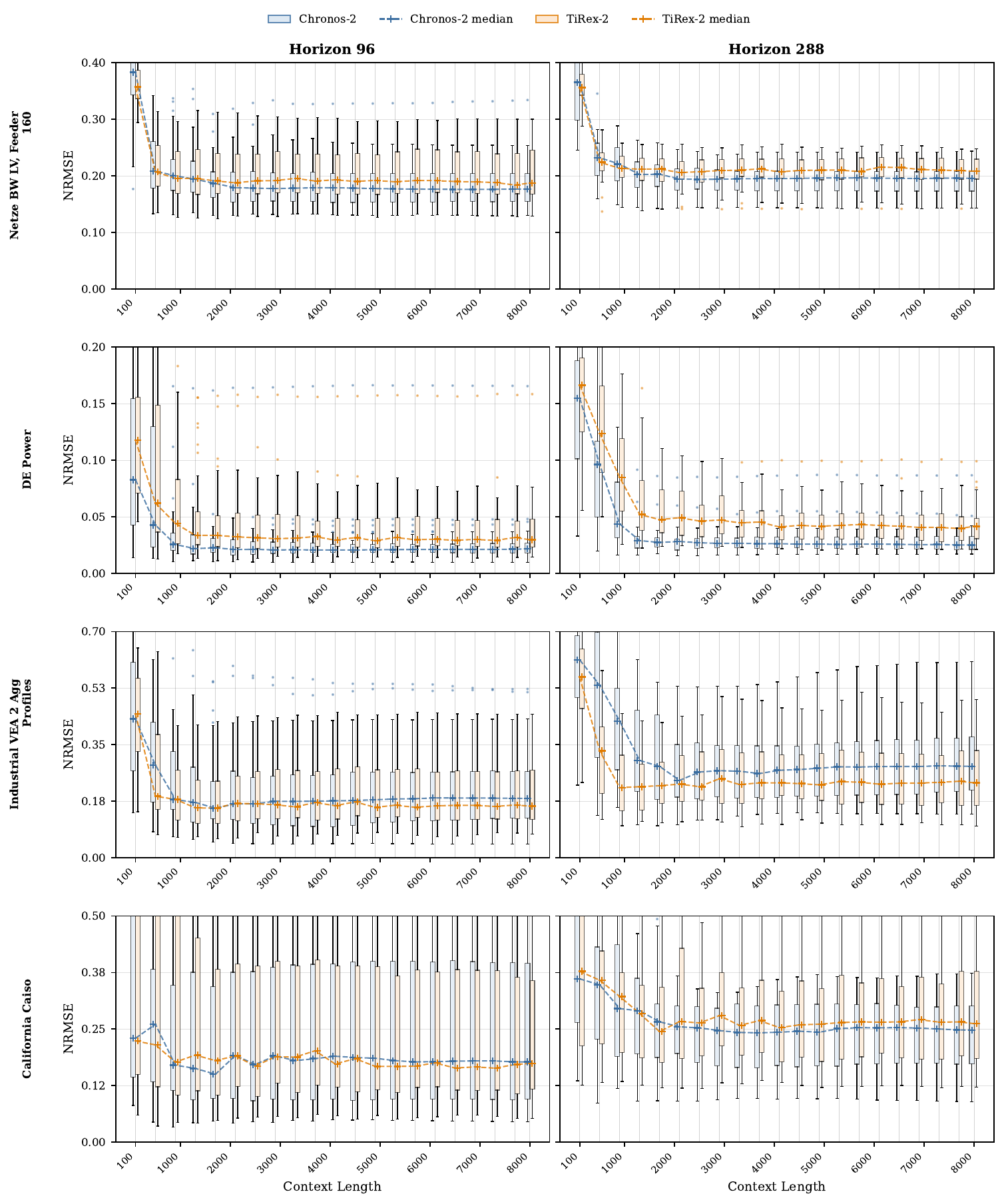}
    \caption{Context length sensitivity analysis for Chronos-2 and TiRex-2 (both covariate-informed) across four representative datasets (DE Power~\cite{entsoe_transparency_2025}, Industrial VEA~\cite{tiemann_industrial_2024}, California CAISO~\cite{CAISO2025OASIS}, Netze BW LV~\cite{treutlein_2025_zenodo}). Each box plot summarizes the NRMSE distribution across all rolling windows within the respective dataset.}
    \label{fig:context_sensitivity}
\end{figure}

First, we analyze the sensitivity to the context length, which we increase from 100 to 8{,}000 time steps (the maximum context length of Chronos-2). For TiRex-2, whose released checkpoint truncates inputs to 2{,}048 time steps, the internal context window is widened to the full input length for this analysis. Figure~\ref{fig:context_sensitivity} shows the results for four representative datasets drawn from three data categories: Load Data (country-level and industrial aggregation), Market Data, and Grid Data. While we observe consistent behavior across additional datasets within each category, similar plots for all datasets are provided in the supplementary material\footnote{\label{fn:supplementary}The supplementary material is part of the publicly released \href{https://github.com/OberMarco/Energy_Benchmark_TSFM_pub}{GitHub repository} (folder \texttt{supplementary\_material/}), archived on Zenodo~\cite{obermeier_code_fets_2026}.}. Across all datasets, forecast horizons, and both models, we observe a consistent pattern: the median NRMSE decreases as the context length increases, saturating at approximately 2{,}000 time steps, corresponding to roughly three weeks of history at the prevalent 15-minute resolution.
We further examine the influence of the forecast horizon, gradually increasing it from 40 up to 1{,}000 time steps (the maximum horizon supported by Chronos-2). For TiRex-2, horizons beyond the native 320-step prediction window of the released checkpoint are obtained by raising the model's internal prediction-window parameter to the required length and forecasting in a single forward pass. We again selected the same four datasets and show the results in Figure~\ref{fig:horizon_sensitivity}; analogous plots for all datasets are again provided in the supplementary material\footref{fn:supplementary}. Across all datasets and context lengths, the analysis reveals a decrease in predictive performance as the forecast horizon increases, reflecting the well-known effect that forecasts become increasingly uncertain the further into the future they extend. For a shorter context length, this performance decrease is somewhat more pronounced and sets in earlier. 

For analyzing the effect of the aggregation level in Figure~\ref{fig:aggregation_sensitivity}, we use the datasets \textit{HTW Berlin Households}~\cite{tjaden_repraesentative_nd}, \textit{Mobilithek}~\cite{Mobilithek2025}, and \textit{Industrial VEA Profiles}~\cite{tiemann_industrial_2024}, selected from Residential Load, Market Data, and Mobility Data as representative datasets containing multiple target variables suitable for aggregation. Here, we observe that the forecast error consistently decreases as the aggregation level increases. This decrease indicates that temporal smoothing through aggregation improves overall forecasting accuracy. This behavior is characteristic of aggregated energy time series and aligns with the concept of standard load profiles, in which larger groups of consumers tend to exhibit more predictable, standardized patterns~\cite{VDEW1999SLP,BDEW2025SLP}.

\clearpage

\clearpage
\begin{figure}[h]
    \centering
    \includegraphics[width=1.01\textwidth]{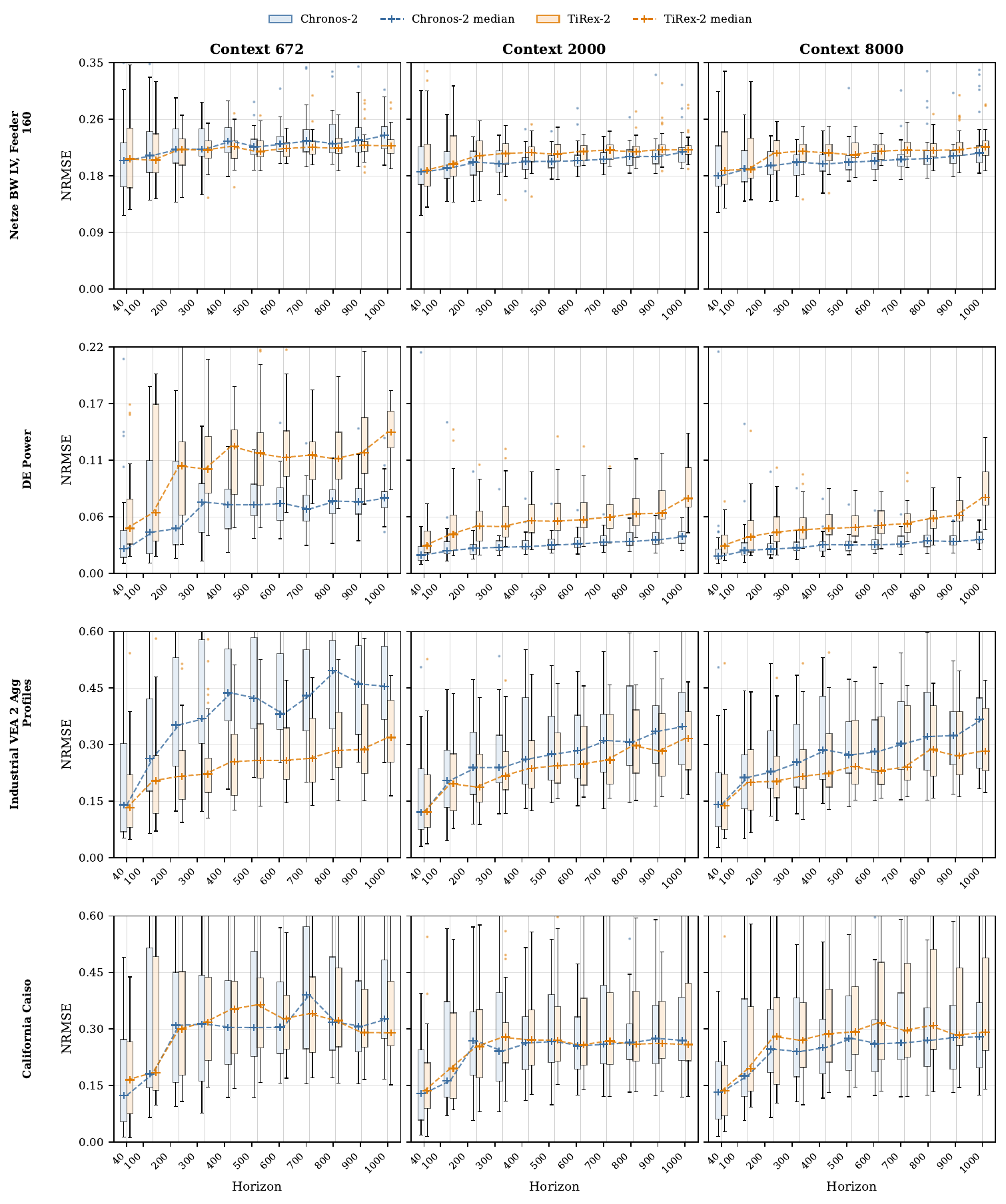}
    \caption{Forecast horizon sensitivity analysis for Chronos-2 and TiRex-2 (both covariate-informed) across four representative datasets (DE Power~\cite{entsoe_transparency_2025}, Industrial VEA~\cite{tiemann_industrial_2024}, California CAISO~\cite{CAISO2025OASIS}, Netze BW LV~\cite{treutlein_2025_zenodo}). Each box plot summarizes the NRMSE distribution across all time series within the respective dataset.}
    \label{fig:horizon_sensitivity}
\end{figure}

\clearpage

\begin{figure}[h]
    \centering
    \includegraphics[width=1.01\textwidth]{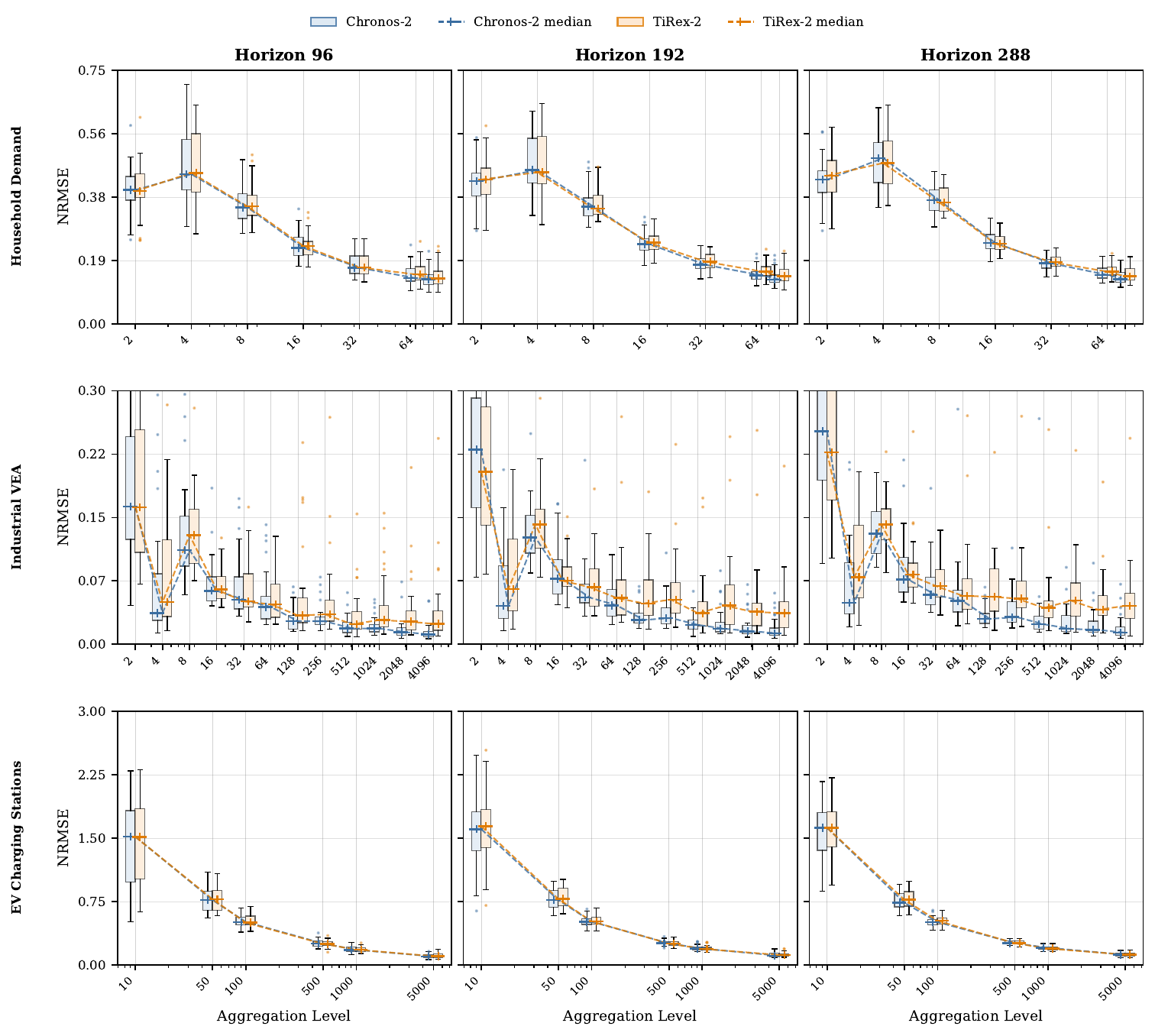}
    \caption{Aggregation level sensitivity analysis for Chronos-2 and TiRex-2 (both covariate-informed) across three representative datasets (Household Demand~\cite{tjaden_repraesentative_nd}, Industrial VEA~\cite{tiemann_industrial_2024}, EV Charging Stations~\cite{Mobilithek2025}). The x-axis represents the number of aggregated individual time series, ranging from a single profile up to the full available pool. Each box plot summarizes the NRMSE distribution.}
    \label{fig:aggregation_sensitivity}
\end{figure}

\clearpage

\section{Discussion}
\label{sec:discussion}

Figure~\ref{fig:energy_forecasts} provides an overview of the wide variation and diverse stakeholder interests within energy-related forecasting. It demonstrates the breadth of the domain and shows that general-purpose benchmarks such as GIFT-Eval~\cite{Aksu2024GIFTEVAL} and fev-bench~\cite{Shchur2025FEVBench} underrepresent energy-specific challenges by collapsing the domain into a single energy category. Our chosen categorization represents a deliberate trade-off between a dataset-level view, which would be too fragmented, and a monolithic energy data category, which obscures distinct stakeholder perspectives and forecasting requirements. Our dataset selection, though limited to available open data, aims to bridge the gap between general-purpose benchmarks and the energy-forecasting view.

\subsection{Model Performance}

Table~\ref{tab:combined_benchmark_reduced} demonstrates that TSFMs in covariate and zero-shot settings achieve the best aggregate performance across energy data categories. This ranking is robust across complementary metrics, all nine forecast-horizon and context-length settings, and the diverse signal characteristics of the collection (Section~\ref{sec:results:overview}; Appendix Tables~\ref{tab:combined_benchmark_mase}--\ref{tab:combined_benchmark_settings}). TSFMs remain competitive with or outperform specialized models, including XGBoost and random forests, even when these models are optimized task-specifically through automated hyperparameter tuning and feature engineering. Additional manual feature engineering might further improve specialized model performance for specific datasets, but similar optimization opportunities exist for TSFMs through covariate engineering and fine-tuning. Overall, this zero-shot predictive advantage of covariate-informed TSFMs over the task-specific baselines substantially reduces the need for dataset-specific model architectures and training procedures, enabling rapid deployment and scalability across the energy industry. The per-model inference times and memory requirements remain low, in parts below those of the task-specific baselines (Appendix Table~\ref{tab:computational_cost}). Furthermore, current TSFMs are comparatively memory-efficient, and for some models covariate-informed inference runs within seconds even on plain CPUs (\ref{sec:app_compute}), supporting this scaling potential. Two further properties reinforce this. First, the data requirements are low: moderate context lengths of around 2{,}000 time steps -- roughly three weeks of history at the prevalent 15-minute resolution -- are typically sufficient, as accuracy saturates beyond this length (Figure~\ref{fig:context_sensitivity}). Hence, instead of accumulating year-long training histories across organizational data silos, a few weeks of recent measurements can suffice for a competitive forecast. Second, the same property may simplify model maintenance under changing data distributions: since forecasts are conditioned only on the recent context window, distribution shifts are accounted for through the updated context, and computationally expensive model updates such as retraining may often be avoided, a potential advantage that we did not evaluate systematically.

Fine-tuning, by contrast -- performed on one individual target series per dataset in our setup -- improved only a minority of series. This is in line with the Chronos-2 technical documentation~\cite{Ansari2025Chronos2}, which still treats fine-tuning as an experimental capability recommended only for collections of roughly a hundred or more time series, as it may otherwise overfit. A small number of fine-tuned configurations even exhibit near-zero forecast scores despite high forecastability (Figure~\ref{fig:forecastability_xy}). Possible explanations include catastrophic forgetting during fine-tuning~\cite{Karaouli2025CatastrophicForgetting}, where task-specific adaptation may degrade the pretrained representation, or overfitting to the fine-tuning window (Section~\ref{sec:ft_ablation}). A full ablation study is beyond the scope of this work, and characterizing when fine-tuning helps versus harms, as well as developing specific fine-tuning strategies, remain open questions for future work.

\subsection{Forecastability and Aggregation Effects}

Figure~\ref{fig:forecastability_xy} establishes a strong relationship between forecasting performance and the Forecastability Index $\phi$ (Section~\ref{sec:results_analysis}), providing practitioners with a heuristic for assessing time series predictability prior to model deployment: aggregated categories such as country-level Load Data, Grid Data, and district Heat Data score high, while individual residential loads and balancing service activations remain the least predictable.
The aggregation sensitivity analysis in Figure~\ref{fig:aggregation_sensitivity} confirms these findings, indicating an inverse relationship between aggregation level and forecasting error. From an energy systems forecasting perspective, these results suggest prioritizing grid- or community-level coordination over individual asset optimization, as greater aggregation yields more reliable forecasts than single-residential-household or asset-level approaches. Since forecasts serve as direct inputs to energy optimization workflows, more accurate forecasts typically translate into better optimization outcomes, e.g., higher renewable usage and cost savings. This supports the concept of local energy communities, where clusters of participants (prosumers, industrial and commercial consumers) are coordinated rather than optimized in isolation, and aligns with established standard load profile concepts~\cite{VDEW1999SLP,BDEW2025SLP}, in which larger consumer aggregations exhibit more predictable consumption patterns. 

\subsection{Limitations}
\label{sec:limitations}

Despite recent advances in machine-learning methods, inherent forecast uncertainty persists in energy time series. Certain data categories remain particularly difficult, such as balancing services with their complex, stochastic system dynamics, or individual residential loads strongly shaped by individual behavior, reflecting the system's irreducible stochasticity.

While our collection already spans a broad range of data categories, countries, and market regulations (summarized per dimension in Appendix Table~\ref{tab:coverage_gaps}), a first limitation is its reliance on publicly available data. Several relevant asset types, stakeholder perspectives, and market designs remain underrepresented and should be added in future work, e.g., commercial-building and prosumer loads, balancing zones outside Germany, and markets on other continents such as South America or Africa. In addition, a finer-grained category structure could capture the heterogeneity within individual data classes more accurately. Extending coverage here is partly constrained since some of these datasets are subject to private or proprietary restrictions. As a result, our findings may not directly transfer to all regional contexts or specialized industrial applications, although the qualitative patterns -- the aggregation benefits, the predictive advantage of TSFMs, their low data requirements, and the resulting scaling potential -- are unlikely to change with extended coverage. Moreover, our evaluation remains restricted to independent data categories rather than coherent, system-wide compositions. A promising direction for future work is to study a complete energy cell -- a region with its loads, generation, storage, grid, and market signals -- forecasting all levels jointly, testing the cross-links between the individual assets, and, as part of covariate engineering, using the related series as covariates for one another, a mode that covariate-informed TSFMs natively support. Such a setting would also renew the case for the multivariate mode, which we excluded here due to its limited gains in prior evaluations (Section~\ref{sec:models}): jointly forecasting the interlinked series of an energy cell could then be compared directly with providing them as covariates.

Second, our evaluation centers on point-forecast accuracy. While probabilistic quality is assessed through nCRPS and prediction-interval coverage -- including a conformalized recalibration of the tree-based baselines (Appendix Tables~\ref{tab:combined_benchmark_ncrps} and~\ref{tab:combined_benchmark_picp}) -- we do not study asymmetric cost structures or tail risks in depth. In operational settings, operators should therefore complement forecast accuracy with scenario analysis or decision-oriented probabilistic methods that explicitly represent residual uncertainty.

Because the zero-shot TSFMs attain a lower overall median NRMSE than the task-specifically trained baselines, a further question is whether this reflects genuine generalization or whether overlap between the pretraining corpora and our evaluation data inflates performance through memorization. We therefore cross-checked every benchmark dataset against the publicly disclosed pretraining sources of each evaluated model: technical reports, model cards, and the released pretraining collections (the Chronos datasets~\cite{Ansari2024Chronos}, GIFT-Eval Pretrain~\cite{Aksu2024GIFTEVAL}, and the underlying LOTSA collection). None of our benchmark datasets appears in these corpora: the only ENTSO-E-derived series in these corpora covers the Spanish market\footnote{The subset \texttt{spain} in LOTSA/GIFT-Eval Pretrain traces back, via the ProEnFo load-forecasting archive~\cite{Wang2023ProEnFo}, to a Kaggle dataset of ENTSO-E load, generation, and price data for Spain (2015--2018).}. In addition, several models rely exclusively or predominantly on synthetic pretraining data: TabPFN is pretrained purely on synthetic tabular data~\cite{Hollmann2025TabPFN}, Toto-2.0 deliberately excludes all public time series from its pretraining mix~\cite{Khwaja2026Toto2}, and Chronos-2 acquires its multivariate capabilities entirely from synthetic data~\cite{Ansari2025Chronos2}; consistently, the synthetic-only variant of Chronos-2 performs comparably to its real-data counterpart on GIFT-Eval~\cite{Aksu2024GIFTEVAL}. Finally, repeating the comparison on the datasets least prone to contamination\footnote{The private Heat Grid Flensburg and Industrial VEA series, the HEAPO heat-pump data, and the recently released LV grid feeders; 3{,}150 experiments per model.} preserves the overall picture: the covariate-informed TSFMs remain the three most accurate models (median NRMSE: TiRex-2 0.36, Chronos-2 0.38, TabPFN-TS 0.39, versus fine-tuned Chronos-2 0.42, XGBoost 0.46, and random forest 0.53). These observations are consistent with an advantage arising from transferable structural patterns rather than from memorization of the evaluation series, although overlap with undisclosed pretraining sources cannot be entirely ruled out. However, energy data constitutes only a small fraction of the vast, predominantly cross-domain pretraining corpora of these models, so a bias of the pretrained weights towards the energy series in our benchmark is unlikely. Pre-registered live benchmarks such as TS-Arena~\cite{Meyer2025TSArena} eliminate such train--test contamination by design and are a promising direction for future evaluation, ideally extended to covariate-informed forecasting.

A further limitation concerns interpretability. In contrast to feature-engineered tree ensembles, TSFM forecasts offer limited insight into how covariates and past observations contribute to predictions. Removal-based attribution methods adapted to TSFMs -- such as the sampling-free, attention-masking SHAP of SHAPformer~\cite{Hertel2026SHAPformer} or the imputation-based explainers for tabular foundation models such as TabPFN-TS~\cite{Hoo2025TabPFNTS} -- provide a route to quantifying these contributions. A first study applying such SHAP-based attributions to covariate-informed TSFMs for load forecasting has recently demonstrated their practical feasibility~\cite{Hertel2026ExplainableTSFM}. Preliminary experiments in this direction are encouraging, but a systematic study is beyond the scope of this work.

Finally, while we compare strong baselines and state-of-the-art TSFMs, we do not exhaust the design space of model architectures, fine-tuning strategies, or feature engineering pipelines. This is a deliberate scope decision: broad architectural coverage, including deep-learning and statistical baselines, is already provided by the general-purpose benchmarks, whereas our aim is the complementary, finer-grained assessment of TSFMs against the tree-based methods most widely used in practical energy forecasting, both discussed in Section~\ref{sec:related_work}. Highly specialized deep-learning models with extensive tuning may surpass the models considered here on individual datasets; the engineering effort this requires, however, may not always be justified at the scale of heterogeneous energy portfolios. Pretrained foundation models are attractive in the same regime, as they achieve competitive accuracy without dataset-specific training; a recent tabular-data benchmark finds foundation models particularly competitive on smaller datasets~\cite{Purucker2026BeyondIID}. Consequently, our results should be interpreted as evidence for the competitiveness and practicality of TSFMs in energy forecasting, rather than as a definitive upper bound on achievable performance.

\subsection{Recommendations}
Our results indicate that the most promising practical setup is to use state-of-the-art TSFMs in the covariate-informed mode for in-context learning with typical exogenous features such as weather variables and temporal features. This recommendation is not tied to a single model: two architecturally distinct foundation models, the transformer-based Chronos-2 and the xLSTM-based TiRex-2, achieve top performance in the covariate-informed mode, with no statistically significant difference between them. In practice, this suggests that future deployments should focus on such covariate setups. Given the advantages demonstrated in this work -- the lowest overall median error, low data requirements, low inference cost, and the resulting scaling potential -- and the rapidly growing capabilities of newly released TSFMs, task-specific machine-learning models may increasingly be superseded. The engineering effort may then shift from feature engineering for classical machine-learning models to covariate engineering for foundation models. 

From a policy perspective, regulations that explicitly enable energy sharing, energy cells, and aggregation at higher grid levels appear more beneficial than a predominant focus on optimizing individual assets in isolation from a forecasting standpoint. As shown above (Figures~\ref{fig:aggregation_sensitivity} and~\ref{fig:forecastability_xy}), forecast quality improves with aggregation.  When shifting the regulatory focus towards coordinated operation and forecasting at these levels, system operators could exploit this effect, which may translate into higher overall socio-economic value. This does not imply that maximal nationwide aggregation is optimal: very high levels of aggregation forgo locally relevant grid information, while isolated single-asset optimization suffers from poor predictability. An intermediate aggregation level that considers infrastructural boundaries, therefore, appears to be the more promising direction. From a broader perspective, these findings raise the question of whether regulatory frameworks should facilitate such coordination and aggregation rather than incentivize isolated single-asset optimization. The forecasting evidence presented here supports this direction, but its broader implications go beyond the scope of this work and require further research.

\section{Conclusion}
\label{sec:conclusion}
This work demonstrates that covariate-informed TSFMs are a promising approach for diverse energy-forecasting applications, achieving the best aggregate median performance in our study. Two independently developed, architecturally distinct foundation models achieve statistically indistinguishable top performance, suggesting that this finding reflects the strength of the modeling approach rather than the properties of a single model.
Critically, TSFMs eliminate the need for extensive model training, requiring only a context window of a few weeks of historical data. This substantially reduces deployment time and directly addresses several barriers identified in the Introduction: newly commissioned renewable assets can be forecast from their first weeks of operation, portfolios fragmented into organizational data silos no longer require pooled training histories, and forecasting becomes economically viable even for niche and small-scale assets. These capabilities unlock significant scalability potential in fragmented, data-limited deployment environments, which is further supported by low inference and hardware demands and potentially lower maintenance effort, as distribution shifts are handled through the updated context rather than through retraining.

Our analysis establishes spectral entropy as an effective indicator of forecastability, enabling practitioners to assess the limits of predictability prior to deployment. Furthermore, aggregated datasets consistently show better predictive performance than individual time series, with forecast error decreasing as the level of aggregation increases.

A natural next step would be to evaluate TSFMs on a regional energy system as a whole -- jointly covering individual households and industrial sites, low- and medium-voltage distribution grids, and transmission-level loads and generation within a single geographic region -- to assess whether a shared forecasting backbone can consistently capture dependencies across aggregation levels instead of forecasting each level in isolation. Future work should investigate fine-tuning strategies for TSFMs tailored to specific energy data categories or trained across all energy time series, explore combining foundation models through simple ensembles or complementary in-context prompts, and improve the interpretability of foundation-model forecasts to support operational trust. More systematic covariate engineering, including domain-specific feature sets for weather, markets, and operational constraints, remains a further open engineering task.

The demonstrated ability of TSFMs to generalize across diverse energy time series data categories without dataset-specific retraining suggests a promising direction for future work: investigating whether such universal forecasting properties can serve as a key component of emerging dynamic network management schemes. This could be a first step towards fully dynamic, flexible connection agreements, where asset-agnostic forecasting across heterogeneous grid-connected assets is a direct operational prerequisite, and, in further steps, towards fully dynamic grid tariffs that require consistent, portfolio-wide load and generation forecasting across all connected stakeholders, including an individual assessment of each network line.

Given the rapid pace of development in this field, ongoing evaluation and further development remains essential: while this study builds on the most recent approaches available at the time, future work should continue to assess newly emerging TSFMs.
\clearpage

\section*{CRediT Author Statement}
\textbf{Marco Obermeier:} Conceptualization, Methodology, Software, Visualization, 
Formal Analysis, Data Curation, Investigation, Writing -- Original Draft.\\
\textbf{Marco Pruckner:} Supervision, Writing -- Review \& Editing.\\
\textbf{Florian Haselbeck:} Supervision, Funding Acquisition, Writing -- Review \& Editing. \\
\textbf{Andreas Zeiselmair:} Supervision, Funding Acquisition, Writing -- Review \& Editing.
\section*{Declaration of Generative AI and AI-assisted Technologies}
During the preparation of this work, the authors used AI-assisted tools for language editing and code development. After using these tools, the authors reviewed and edited the content as needed and take full responsibility for the content of the publication.
\section*{Declaration of Competing Interests}
The authors declare that they have no known competing financial interests or personal relationships that could have appeared to influence the work reported in this paper.

\section*{Acknowledgements}
The authors thank the open-source community for providing the tools and libraries on which this work builds, including Chronos, TimesFM, TiRex, FlowState, Toto, TabPFN, XGBoost, and RAPIDS cuML (random forest). We further thank all data providers for making their datasets publicly available, enabling reproducible research in the energy domain. The initial idea for this work emerged during the first author's prior collaboration with the TiRex development team at Johannes Kepler University Linz (JKU), whose pioneering work on zero-shot time series forecasting provided the key inspiration for this research. Building on this, the work was subsequently realized and further improved at Weihenstephan-Triesdorf University of Applied Sciences within the smartBattery project, funded by the Deutsche Bundesstiftung Umwelt (DBU) under grant number \texttt{40164-01} and co-funded by Bayernwerk Netz GmbH. The smartBattery project investigates AI-based methods for the grid-serving integration of large-scale battery storage systems. Within this context, the present work contributes universal, generalizable forecasting approaches -- including envelope-based predictions -- as a foundational step toward the implementation of dynamic grid tariffs.

\clearpage
\appendix
\setcounter{table}{0}\setcounter{figure}{0}
\makeatletter\@addtoreset{table}{section}\@addtoreset{figure}{section}\makeatother
\section{Data Statistics \& Model Specifications}

\begin{table}[h]
\begin{scriptsize}
\caption{Coverage of the FETS dataset collection. As an exhaustive collection is infeasible, a representative subset was selected per category and region; the table summarizes the represented scope per dimension, indicating the range within which the findings generalize. The full per-dataset list is provided in Table~\ref{tab:datasets_comprehensive}.}
\label{tab:coverage_gaps}
\centering
\begin{tabularx}{\textwidth}{@{}l X@{}}
\hline
\addlinespace[2pt]
\textbf{Dimension} & \textbf{Represented scope in FETS} \\
\hline
\addlinespace[2pt]
\multicolumn{2}{@{}l}{\textbf{Geography \& Market Regulation}} \\
\addlinespace[2pt]
\hline
\addlinespace[2pt]
Regions & Nine regions, Europe-centric (DE, FR, DK, UK, NL, CH), plus one US (California), one mainland-Chinese, and one Hong Kong dataset \\
Market designs & European zonal day-ahead and intraday, German balancing, US nodal locational marginal pricing \\
\hline
\addlinespace[2pt]
\multicolumn{2}{@{}l}{\textbf{Asset Categories}} \\
\addlinespace[2pt]
\hline
\addlinespace[2pt]
Load & Country- and system-level aggregated demand and large industrial loads \\
Residential load & Individual households and small residential portfolios, with and without behind-the-meter PV \\
Non-dispatchable gen. & Solar and wind at aggregated (country/zonal) and single-site or wind-park level \\
Dispatchable gen. & Thermal (coal and gas) generation and aggregated battery-storage dispatch \\
Grid data & Distribution-level feeders (low and medium voltage) and transmission-line flows \\
Heat data & District-heating network load and residential heat-pump electricity demand \\
Market data & Day-ahead, intraday, balancing, and nodal locational electricity prices \\
Balancing services & Secondary and tertiary reserve activation and area control error (net regulation volume) \\
Mobility data & EV charging demand, from individual sites to aggregated charging networks \\
\hline
\end{tabularx}
\end{scriptsize}
\end{table}

\clearpage

\begin{table}[htbp]
\centering
\caption{Dataset Statistics Overview: Comprehensive characteristics of all time series datasets including sample size (N), temporal resolution (Frequency), coefficient of variation (CV), percentage of negative and positive values, and Forecastability Index ($\phi$). The Forecastability Index is based on spectral entropy, where values closer to 1 indicate more predictable time series with strong periodic patterns, while values near 0 suggest higher randomness and lower predictability.}
\footnotesize
\begin{adjustbox}{max width=\textwidth}
\begin{tabular}{l l c c c c c c}
\toprule
\textbf{Dataset} & \textbf{Target} & \textbf{N} & \textbf{Frequency} & \textbf{CV} & \textbf{Neg.\%} & \textbf{Pos.\%} & $\boldsymbol{\phi}$ \\
\midrule
\multicolumn{8}{l}{\textbf{Balancing Services}} \\
\\[-1.5ex]
aFRR Germany & Germany aFRR positiv & 348,576 & 15min & 1.769 & 0.0\% & 90.2\% & 0.1428 \\
Balancing Data Germany & Area Control Error (NRV Saldo) & 364,303 & 15min & 9.550 & 45.5\% & 54.5\% & 0.1343 \\
aFRR Germany & Germany aFRR negativ & 348,576 & 15min & 1.734 & 0.0\% & 92.0\% & 0.1286 \\
mFRR Germany & TenneT positiv & 383,704 & 15min & 8.780 & 0.0\% & 2.3\% & 0.1095 \\
mFRR Germany & TenneT negativ & 383,704 & 15min & 11.994 & 0.0\% & 1.4\% & 0.0961 \\
\\[-1.5ex]
\multicolumn{8}{l}{\textbf{Dispatchable Generation}} \\
\\[-1.5ex]
California CAISO & Aggregated Batterys & 44,352 & 5min (100\%), 1.1h (0\%) & 18.011 & 51.7\% & 48.2\% & 0.8084 \\
ENTSO-E Germany & Fossil Hard Coal & 202,844 & 15min & 0.722 & 0.0\% & 100.0\% & 0.6055 \\
ENTSO-E Germany & Fossil Gas & 202,844 & 15min & 0.562 & 0.0\% & 100.0\% & 0.5916 \\
\\[-1.5ex]
\multicolumn{8}{l}{\textbf{Grid Data}} \\
\\[-1.5ex]
Netze BW LV, Feeder 37 & Active Power & 70,171 & 15min (100\%), 45.0min (0\%) & 0.343 & 0.0\% & 100.0\% & 0.7386 \\
Bayernwerk & Grid Consumption MS & 35,136 & 15min & 0.274 & 0.0\% & 100.0\% & 0.6738 \\
Bayernwerk & Grid Feed-in MS & 35,136 & 15min & 0.564 & 0.0\% & 98.5\% & 0.6706 \\
Netze BW LV, Feeder 132 & Active Power & 69,877 & 15min (100\%), 2h (0\%) & 1.642 & 18.8\% & 81.2\% & 0.5776 \\
Netze BW LV, Feeder 97 & Active Power & 70,150 & 15min (100\%), 1h (0\%) & 26.891 & 27.4\% & 72.6\% & 0.5773 \\
Netze BW LV, Feeder 63 & Active Power & 70,104 & 15min (100\%), 1h (0\%) & 4.826 & 16.5\% & 83.5\% & 0.5263 \\
Netze BW LV, Feeder 160 & Active Power & 70,169 & 15min (100\%), 2h (0\%) & 0.482 & 0.0\% & 100.0\% & 0.5205 \\
Netze BW LV, Feeder 33 & Active Power & 70,173 & 15min (100\%), 1h (0\%) & 2.007 & 17.3\% & 82.7\% & 0.4949 \\
50Hertz Line Power & UW Wieselbach UW Lauchstädt L472 & 19,729 & 1h & 0.620 & 0.0\% & 95.7\% & 0.3671 \\
50Hertz Line Power & UW Wieselbach UW Lauchstädt L471 & 19,729 & 1h & 0.652 & 0.0\% & 93.4\% & 0.3473 \\
50Hertz Line Power & UW Lubmin - OWP Baltic Eagle & 13,057 & 1h & 0.995 & 0.0\% & 97.4\% & 0.2770 \\
\\[-1.5ex]
\multicolumn{8}{l}{\textbf{Heat Data}} \\
\\[-1.5ex]
Heat Grid Flensburg & Heat Demand & 43,843 & 1h (100\%), 2h (0\%) & 0.565 & 0.0\% & 100.0\% & 0.8657 \\
HEAPO: Household Zurich Area & Heatpump 971151 & 35,616 & 15min (100\%), 1.0d (0\%) & 1.502 & 0.0\% & 99.7\% & 0.3718 \\
HEAPO: Household Zurich Area & Heatpump 818882 & 75,168 & 15min (100\%), 1.0d (0\%) & 1.569 & 0.0\% & 100.0\% & 0.3160 \\
\\[-1.5ex]
\multicolumn{8}{l}{\textbf{Load}} \\
\\[-1.5ex]
California CAISO & Aggregated Load & 44,352 & 5min (100\%), 1.1h (0\%) & 0.189 & 0.0\% & 100.0\% & 0.7487 \\
DE Power & Load DE & 202,844 & 15min & 0.177 & 0.0\% & 100.0\% & 0.7406 \\
FR Power & Load & 202,695 & 15min & 0.216 & 0.0\% & 100.0\% & 0.7240 \\
Industrial VEA 2 Agg Profiles & 2 random loads & 35,136 & 15min & 0.560 & 0.0\% & 100.0\% & 0.5909 \\
\\[-1.5ex]
\multicolumn{8}{l}{\textbf{Market Data}} \\
\\[-1.5ex]
California CAISO & LMP ZP26 & 44,352 & 5min (100\%), 1.1h (0\%) & 0.474 & 4.8\% & 95.1\% & 0.5229 \\
ENTSO-E Germany & Day-Ahead Auction & 78,812 & 15min & 0.609 & 5.5\% & 93.7\% & 0.4667 \\
California CAISO & LMP SP15 & 44,352 & 5min (100\%), 1.1h (0\%) & 0.556 & 4.6\% & 95.3\% & 0.4032 \\
California CAISO & LMP NP15 & 44,352 & 5min (100\%), 1.1h (0\%) & 0.410 & 1.1\% & 98.8\% & 0.3623 \\
ENTSO-E Germany & Intraday Continuous & 35,136 & 15min & 1.028 & 6.7\% & 93.3\% & 0.2920 \\
ENTSO-E Germany & reBAP & 416,160 & 15min & 2.827 & 18.9\% & 81.1\% & 0.1215 \\
\\[-1.5ex]
\multicolumn{8}{l}{\textbf{Mobility Data}} \\
\\[-1.5ex]
NL Office EV Park & Total Power & 143,922 & 15min & 2.354 & 0.0\% & 44.9\% & 0.6301 \\
UrbanEV China Charging Station & Aggregated Profile 107 & 17,376 & 15min & 0.387 & 0.0\% & 100.0\% & 0.4608 \\
Mobilitaet Public Charging Station & Total Power 50 Agg & 228,179 & 15min & 1.423 & 0.0\% & 66.7\% & 0.3955 \\
UK Residential Charging Station & Total Power 50 Agg & 35,058 & 15min & 1.585 & 0.0\% & 36.1\% & 0.2278 \\
\\[-1.5ex]
\multicolumn{8}{l}{\textbf{Non-Dispatchable Generation}} \\
\\[-1.5ex]
California CAISO & Aggregated Solar & 44,352 & 5min (100\%), 1.1h (0\%) & 1.295 & 48.4\% & 51.3\% & 0.8273 \\
DE Power & Solar & 202,844 & 15min & 1.537 & 0.0\% & 72.6\% & 0.7628 \\
FR Power & Solar & 202,816 & 15min & 1.458 & 0.0\% & 58.7\% & 0.7344 \\
PV Hong Kong & generation_kwh_UG Hall4 & 55,577 & 15min & 1.624 & 0.0\% & 47.9\% & 0.6574 \\
DK Power & Solar & 50,660 & 1h & 1.759 & 0.0\% & 84.6\% & 0.6138 \\
Energy Forecasting Competition PV Cluster & Solar_MWh_credit & 61,505 & 30min (100\%), 1h (0\%) & 1.631 & 0.0\% & 52.0\% & 0.5749 \\
DE Power & Wind Onshore & 202,844 & 15min & 0.802 & 0.0\% & 100.0\% & 0.4695 \\
DE Power & Wind Offshore & 202,844 & 15min & 0.686 & 0.0\% & 99.9\% & 0.4189 \\
Energy Forecasting Competition Wind+PV & total_MWh_credit & 61,505 & 30min (100\%), 1h (0\%) & 0.667 & 0.0\% & 97.7\% & 0.3973 \\
Energy Forecasting Competition Windpark & Wind_MWh_credit & 61,505 & 30min (100\%), 1h (0\%) & 0.761 & 0.0\% & 94.3\% & 0.3672 \\
Hill of Towie Wind & active_power & 58,434 & 15min (100\%), 30min (0\%) & 1.166 & 17.2\% & 82.7\% & 0.3617 \\
\\[-1.5ex]
\multicolumn{8}{l}{\textbf{Residential Load}} \\
\\[-1.5ex]
HTW Berlin Households & 10 random aggregated households & 35,040 & 15min & 0.511 & 0.0\% & 100.0\% & 0.4222 \\
Household Lower Saxony & NO_PV_SFH31_P_TOT & 88,425 & 15min & 1.231 & 0.0\% & 99.9\% & 0.2295 \\
HTW Berlin Households & random_household_1 & 35,040 & 15min & 0.879 & 0.0\% & 100.0\% & 0.2256 \\
HTW Berlin Households & random_household_0 & 35,040 & 15min & 0.938 & 0.0\% & 100.0\% & 0.2102 \\
Household Lower Saxony & WITH_PV_SFH15_P_TOT & 69,210 & 15min & 2.390 & 10.6\% & 89.4\% & 0.2020 \\
Household Lower Saxony & NO_PV_SFH10_P_TOT & 89,630 & 15min & 0.677 & 0.0\% & 99.9\% & 0.1527 \\
HTW Berlin Households & random_household_2 & 35,040 & 15min & 1.602 & 0.0\% & 100.0\% & 0.1339 \\
\\[-1.5ex]
\bottomrule
\end{tabular}
\end{adjustbox}
\label{tab:dataset_statistics}
\end{table}
\clearpage

\begin{scriptsize}
\begin{table}[htbp]
\centering
\caption{Model Specifications. Hyperparameter search spaces and configuration for trained models. For XGBoost and Random Forest, the Optuna TPE sampler is used with $n=250$ trials. SHAP-based feature selection and automatic lag selection are enabled for both tree-based models.}
\small
\begin{adjustbox}{max width=\textwidth}
\begin{tabular}{l l l l}
\toprule
\textbf{Parameter} & \textbf{Chronos-2} & \textbf{XGBoost} & \textbf{Random Forest} \\
\midrule
\multicolumn{4}{l}{\textit{General}} \\
\midrule
HPO Trials               & --          & 250             & 250 \\
Quantiles                & [0.1, 0.5, 0.9] & [0.1, 0.5, 0.9] & [0.1, 0.5, 0.9] \\
\midrule
\multicolumn{4}{l}{\textit{Lag Selection}} \\
\midrule
Auto Lag Selection       & --          & \checkmark      & \checkmark \\
Max Lags to Test         & --          & 100             & 100 \\
Top-$k$ Lags Selected    & --          & 20              & 20 \\
\midrule
\multicolumn{4}{l}{\textit{Feature Engineering}} \\
\midrule
Rolling Windows          & --          & [3, 6, 12, 24, 48, 96] & [3, 6, 12, 24, 48, 96] \\
Rolling Features         & --          & mean, std, min, max & mean, std, min, max \\
SHAP Feature Selection   & --          & \checkmark      & \checkmark \\
SHAP Top-$k$ Features    & --          & 50              & 50 \\
\midrule
\multicolumn{4}{l}{\textit{Hyperparameter Search Space}} \\
\midrule
n\_estimators         & --  & $[100, 1000]$ (int)   & $[100, 1000]$ (int) \\
max\_depth            & --  & $[3, 10]$ (int)       & $[3, 30]$ (int) \\
learning\_rate        & --  & $[0.01, 0.2]$ (log)   & -- \\
subsample             & --  & $[0.3, 1.0]$          & -- \\
colsample\_bytree     & --  & $[0.2, 1.0]$          & -- \\
min\_samples\_split   & --  & --                    & $[2, 20]$ (int) \\
min\_samples\_leaf    & --  & --                    & $[1, 10]$ (int) \\
max\_features         & --  & --                    & \{sqrt, log2, 0.5, 0.7, 1.0\} \\
bootstrap             & --  & --                    & \{true, false\} \\
\midrule
\multicolumn{4}{l}{\textit{Fine-tuning (Chronos-2 only)}} \\
\midrule
Num Steps                & 1,000      & --              & -- \\
Learning Rate            & $1 \times 10^{-6}$ & --       & -- \\
Batch Size               & 8           & --              & -- \\
Window Overlap           & 75\%        & --              & -- \\
LoRA Rank                & 8           & --              & -- \\
LoRA Alpha               & 16          & --              & -- \\
LoRA Dropout             & 0.0         & --              & -- \\
LoRA Target Modules      & \{q,\,k,\,v,\,o,\,\texttt{output\_patch\_embedding}\} & --              & -- \\
Checkpoint               & chronos-2 & --      & -- \\
\bottomrule
\end{tabular}
\end{adjustbox}
\label{tab:model_specifications}
\end{table}
\end{scriptsize}
\clearpage
\section{Computational Costs}
\label{sec:app_compute}

\textbf{Hardware and measurement:} All experiments ran on a server with 2$\times$ AMD EPYC 9334 CPUs, 1.1\,TiB of memory, and 4$\times$ NVIDIA RTX~6000 Ada Generation GPUs (49{,}140\,MiB $\approx$ 48\,GiB VRAM each). Per model and benchmark mode, Table~\ref{tab:computational_cost} reports four wall-clock quantities: one-time initialization (model loading), inference time per rolling-window forecast (excluding loading), task-specific training or fine-tuning time where applicable, and peak main and GPU memory. Each worker runs on an exclusively pinned CPU set (tree-based training: 6--10 cores; fine-tuning: 8; zero-shot inference: 1--8) and one GPU, shared by up to twelve concurrent zero-shot workers: the reported inference times are conservative. Memory peaks are sampled from the full process tree (main) and the CUDA allocator (GPU); measured values are decimal GB ($10^9$\,B), hardware capacities binary (GiB/TiB). Inference is timed per single forecast; with batching, which some foundation models support, the ordering can differ: on fev-bench~\cite{Shchur2025FEVBench}, TiRex-2 is reported substantially faster than Chronos-2~\cite{Podest2026TiRex2}.

\textbf{Results:} The complete benchmark took about one week of wall-clock time, dominated by tree-baseline training and Chronos-2 fine-tuning. Some TSFMs, e.g., Chronos-2 and TiRex-2, are faster in inference than XGBoost and random forest; Chronos-2 also trains faster at similar memory. The \texttt{---} entries mark not-applicable cells: the foundation models (including TabPFN-TS) require no per-series training, and tree-baseline initialization was not measured separately. The table thus compares zero-shot inference with training-based deployment: the practically relevant view, since practitioners train the tree ensembles themselves, whereas foundation models require only covariate engineering. Pure inference memory: XGBoost 0.30--0.32\,GB across single runs; random forest 0.9\,GB for typical models up to about 7\,GB for the largest tuned forests. For some models including Chronos-2 and TiRex-2, inference also runs without any GPU: across all 54 datasets on eight cores of a desktop CPU (AMD Ryzen~9 9950X; reduced replication: context length 2{,}000, horizon 96, eight of the 35 rolling windows per dataset), covariate-informed forecasts take a median of 1.4\,s for Chronos-2 (0.3--7.8\,s, 1.1--3.3\,GB) and 1.1\,s for TiRex-2 (0.2--48\,s, 1.0--2.4\,GB), 6 to 9 times slower in the median than the identical GPU runs and up to roughly 70 times for TiRex-2 on the largest covariate stacks. Overall, TSFMs incur no computational drawbacks compared to feature-based machine learning.

\begin{table}[htbp]
\centering
\caption{Computational cost per model and benchmark mode over all 17{,}010 experiments: times as median (Q10--Q90), memory as median (min--max) in decimal GB ($10^9$\,B); \texttt{---} = not applicable.}
\small
\begin{adjustbox}{max width=\textwidth}
\begin{tabular}{l l c c c c c}
\toprule
\textbf{Mode} & \textbf{Model} & \textbf{Init [s]} & \textbf{Inference [s]} & \textbf{Training [min]} & \textbf{Peak RAM [GB]} & \textbf{Peak GPU Mem [GB]} \\
\midrule
Univariate & Toto-2.0 & 15.80 (15.49--16.23) & 0.47 (0.43--0.50) & --- & 1.6 (1.2--5.1) & 10.6 (10.6--10.6) \\
 & TimesFM & 2.03 (1.98--2.11) & 0.92 (0.84--1.22) & --- & 1.3 (0.8--4.1) & 1.7 (1.7--1.7) \\
 & Chronos-2 & 0.19 (0.19--0.22) & 0.23 (0.21--0.41) & --- & 1.4 (1.0--4.6) & 1.0 (1.0--1.1) \\
 & FlowState & 1.69 (1.44--1.80) & 0.37 (0.20--0.41) & --- & 1.5 (1.0--4.2) & 0.8 (0.7--1.5) \\
 & TiRex-2 & 3.06 (2.60--5.70) & 0.68 (0.58--0.81) & --- & 1.7 (1.5--4.2) & 0.9 (0.9--0.9) \\
\midrule
Covariate & Chronos-2 & 0.24 (0.21--0.55) & 0.47 (0.41--1.11) & --- & 1.5 (0.9--5.8) & 1.3 (1.0--5.7) \\
 & TabPFN-TS & 0.30 (0.22--0.35) & 2.16 (1.18--4.81) & --- & 2.4 (1.3--7.2) & 1.7 (0.4--4.6) \\
 & TimesFM & 2.17 (1.99--2.57) & 1.04 (0.92--2.45) & --- & 1.7 (1.1--6.9) & 1.7 (1.7--1.7) \\
 & TiRex-2 & 2.21 (1.64--6.93) & 0.69 (0.56--0.96) & --- & 1.8 (1.4--4.4) & 1.0 (0.9--1.6) \\
\midrule
Training / Fine-tuning & Chronos-2 & 0.23 (0.21--0.24) & 0.46 (0.39--1.20) & 1.3 (1.2--8.6) & 2.8 (1.1--26.7) & 2.2 (1.7--49.7) \\
 & RandomForest & --- & 1.13 (0.76--1.58) & 37.9 (23.5--56.3) & 3.2 (0.8--36.4) & 0.5 (0.4--3.5) \\
 & XGBoost & --- & 1.13 (0.67--2.13) & 8.1 (4.4--14.4) & 1.9 (0.6--6.3) & --- \\
 & XGBoost (multi-output) & --- & 0.53 (0.43--0.79) & 19.5 (9.6--41.8) & 4.7 (1.2--68.8) & --- \\
\bottomrule
\end{tabular}
\end{adjustbox}
\label{tab:computational_cost}
\end{table}
\clearpage
\section{Extended Results}
This section extends the benchmark results:
Figure~\ref{fig:einspeisung_vorgl_netz_ms_21_03_2025_kw_forecast} shows an example prediction for one specific experimental setting, representative of the many configurations evaluated as described above. Tables~\ref{tab:combined_benchmark_mase}--\ref{tab:combined_benchmark_picp} complement this overview with five additional metrics\footnote{In these tables, blank cells mark experiments where the respective metric is undefined: MASE where the seasonal-naive baseline reproduces the evaluation window exactly (zero denominator, e.g., a window and its preceding season block that are both constantly zero); nCRPS where the quantile loss of the seasonal-naive baseline over the evaluation window is zero (the same degenerate windows; since the point-error and quantile-loss criteria differ slightly, the blank cells do not coincide exactly); and CORR where the ground truth or the point forecast is constant within the evaluation window (zero variance), e.g., all-zero windows or flat forecasts on sparse series -- because a constant forecast alone is sufficient, more cells are blank than for MASE and nCRPS. For MASE and nCRPS, ratios above 100, which arise from near-zero denominators, are likewise discarded.}; Table~\ref{tab:metric_overview} summarizes their definitions, interpretation, and relevance for energy applications. Table~\ref{tab:combined_benchmark_settings} further reports the overall NRMSE per model separately for each of the nine forecast-horizon/context-length settings.

The PICP results (Table~\ref{tab:combined_benchmark_picp}) show that the quantile intervals of the tree-based baselines are far too narrow (overall median deviation of $-49$~percentage points for XGBoost and $-79$ for random forest). This is a calibration issue rather than a limitation of the models themselves. We therefore repeated both training campaigns with conformalized quantile regression (CQR)~\cite{Romano2019CQR}: the most recent 15\% of each training set is held out, signed residuals of recursive multi-step forecasts from several origins in this slice are pooled per quantile and lead-time bin, and the resulting additive corrections are applied at forecast time with monotone rearrangement to prevent quantile crossing. This recalibration restores near-nominal coverage (overall median PICP of 0.78 for XGBoost and 0.79 for random forest at a nominal level of 0.8) and improves probabilistic accuracy (overall median nCRPS 0.567 versus 0.672 for XGBoost and 0.645 versus 0.948 for random forest), while point accuracy remains virtually unchanged (overall median NRMSE 0.613 versus 0.611 for XGBoost and 0.697 versus 0.696 for random forest). The multi-output XGBoost variant is excluded from this recalibration, as XGBoost does not support the quantile objective for multi-target models. The recalibrated variants appear as additional rows in the PICP and nCRPS tables (Tables~\ref{tab:combined_benchmark_picp} and~\ref{tab:combined_benchmark_ncrps}); all other tables report the uncalibrated variants used in the main benchmark.

\begin{table}[htbp]
\centering
\caption{Additional evaluation metrics reported in Tables~\ref{tab:combined_benchmark_mase}--\ref{tab:combined_benchmark_picp}. MASE and NCRPS are normalized by the same seasonal naive forecast, which repeats the last observed seasonal cycle (period $m$, here one day) across the forecast window of length $n$; $\hat{y}_i^{(q)}$ denotes the predicted $q$-quantile and $\rho_q(y, \hat{y}) = 2\left(q - \mathbb{1}\{y < \hat{y}\}\right)\left(y - \hat{y}\right)$ the pinball loss. PICP calibration refers to the 80\% prediction interval (quantile band $0.1$--$0.9$).}
\small
\begin{adjustbox}{max width=\textwidth}
\begin{tabular}{>{\raggedright\arraybackslash}m{2.7cm} l m{3.9cm} m{4.9cm}}
\toprule
\textbf{Metric} & \textbf{Definition} & \textbf{Interpretation} & \textbf{Relevance for Energy Applications} \\
\midrule
Mean absolute scaled error (MASE) &
$\dfrac{\frac{1}{n}\sum_{i=1}^{n} |y_i - \hat{y}_i|}{\frac{1}{n}\sum_{i=1}^{n} |y_i - y_{i-m}|}$ &
Scale-free point forecast error, well-defined for near-zero means; $<1$: better than seasonal naive (e.g., $0.8$: 20\% lower error), $>1$: repeating the previous day performs better. &
Robust model comparison across heterogeneous signals, including signed quantities such as battery dispatch or balancing energy. \\
\addlinespace
Normalized continuous ranked probability score (NCRPS) &
$\dfrac{\sum_{q \in \mathcal{Q}} \sum_{i=1}^{n} \rho_q\!\left(y_i, \hat{y}_i^{(q)}\right)}{\sum_{q \in \mathcal{Q}} \sum_{i=1}^{n} \rho_q\!\left(y_i, y_{i-m}\right)}$ &
Probabilistic forecast quality across all quantile levels $q \in \mathcal{Q}$; same baseline as MASE, hence directly comparable: $<1$ better, $>1$ worse than seasonal naive. &
Assesses the full predictive distribution required for stochastic optimization and scenario-based planning. \\
\addlinespace
Correlation (CORR) &
$\operatorname{corr}\!\left(y, \hat{y}\right)$ &
Reproduction of the temporal shape, independent of amplitude and offset errors; $1$: perfect pattern, $0$: no tracking, negative: inverted pattern. &
Correct timing of load or photovoltaic feed-in peaks, decisive for congestion management and peak shaving in distribution grid operation. \\
\addlinespace
Sum ratio &
$\dfrac{\sum_{i=1}^{n}\hat{y}_i}{\sum_{i=1}^{n} y_i}$ &
Bias in the total predicted quantity, independent of its temporal distribution; $1$: exact match, $0.9$/$1.1$: 10\% under-/overestimation. &
Total energy over the horizon, e.g., whether a heat storage suffices to cover district heating demand, or volume-based day-ahead procurement. \\
\addlinespace
Prediction interval coverage probability ($\Delta$PICP) &
$(\text{PICP} - 0.80)\cdot 100$ &
Calibration in percentage points; $0$: perfect, $<0$: overconfident, too narrow (e.g., $-13$: 67\% instead of 80\% coverage), $>0$: underconfident, too wide. &
Risk-aware decisions such as dimensioning balancing reserves or safety margins; overconfidence underestimates required reserves, overly wide intervals are unnecessarily costly. \\
\bottomrule
\end{tabular}
\end{adjustbox}
\label{tab:metric_overview}
\end{table}

\clearpage
\input{A11_plot_example1_worse}

\begin{sidewaystable}[htbp]
\centering
\caption{Results Univariate Benchmark (NRMSE): Statistics (Min, Q25, Median, Q75, Max) over multiple forecast horizons and context lengths; best median per row highlighted. Overall row aggregates globally across all datasets, targets, horizons, and context lengths. Values $>\!10^{6}$ are treated as diverged, unusable forecasts and displayed as $>\!10^{6}$. All 102,060 experiments are evaluated.}
\small
\begin{adjustbox}{max width=\linewidth}
\begin{tabular}{l l c c c c c c c c c c c c c c c c c c c c c c c c c }
\toprule
\textbf{Dataset} & \textbf{Target} & \multicolumn{5}{c}{\textbf{Toto-2.0}} & \multicolumn{5}{c}{\textbf{TimesFM}} & \multicolumn{5}{c}{\textbf{Chronos-2}} & \multicolumn{5}{c}{\textbf{FlowState}} & \multicolumn{5}{c}{\textbf{TiRex-2}} \\
\cmidrule(lr){3-7} \cmidrule(lr){8-12} \cmidrule(lr){13-17} \cmidrule(lr){18-22} \cmidrule(lr){23-27}
& & Min & Q25 & Med & Q75 & Max & Min & Q25 & Med & Q75 & Max & Min & Q25 & Med & Q75 & Max & Min & Q25 & Med & Q75 & Max & Min & Q25 & Med & Q75 & Max \\
\midrule
\multicolumn{27}{l}{\textbf{Balancing Services}} \\
\\[-1.5ex]
Balancing Data Germany & Area Control Error (NRV Saldo) & \val{1.0884} & \val{1.2030} & \val{1.2555} & \val{1.2987} & \val{1.7188} & \val{1.1012} & \val{1.2124} & \val{1.2533} & \val{1.2934} & \val{1.5940} & \val{1.1112} & \val{1.2110} & \cellcolor{green!20}\textbf{\val{1.2490}} & \val{1.2912} & \val{1.6334} & \val{1.0254} & \val{1.2002} & \val{1.2610} & \val{1.3107} & \val{1.9254} & \val{1.0931} & \val{1.2069} & \val{1.2549} & \val{1.2926} & \val{1.6258} \\
aFRR Germany & Germany aFRR negativ & \val{1.3962} & \val{1.9459} & \val{2.2609} & \val{2.6025} & \val{3.9463} & \val{1.5004} & \val{1.9190} & \cellcolor{green!20}\textbf{\val{2.2437}} & \val{2.5947} & \val{3.9192} & \val{1.4674} & \val{1.9179} & \val{2.2559} & \val{2.5909} & \val{3.9434} & \val{1.3761} & \val{1.9080} & \val{2.2475} & \val{2.6100} & \val{3.9422} & \val{1.5133} & \val{1.9540} & \val{2.2702} & \val{2.6270} & \val{3.9303} \\
aFRR Germany & Germany aFRR positiv & \val{1.6501} & \val{2.0135} & \cellcolor{green!20}\textbf{\val{2.1700}} & \val{2.3543} & \val{3.0390} & \val{1.6232} & \val{2.0210} & \val{2.2147} & \val{2.3711} & \val{3.0295} & \val{1.5813} & \val{2.0153} & \val{2.1776} & \val{2.3586} & \val{2.9942} & \val{1.5732} & \val{1.9847} & \val{2.1786} & \val{2.3292} & \val{3.0791} & \val{1.6064} & \val{2.0277} & \val{2.2164} & \val{2.3797} & \val{3.0680} \\
\bottomrule
\end{tabular}
\end{adjustbox}
\label{tab:univariate_benchmark}
\end{sidewaystable}
\clearpage

\begin{sidewaystable}[htbp]
\centering
\caption{Results Covariate Benchmark (NRMSE): Statistics (Min, Q25, Median, Q75, Max) over multiple forecast horizons and context lengths; best median per row highlighted. Overall row aggregates globally across all datasets, targets, horizons, and context lengths. Values $>\!10^{6}$ are treated as diverged, unusable forecasts and displayed as $>\!10^{6}$. All 68,040 experiments are evaluated.}
\footnotesize
\begin{adjustbox}{max width=\linewidth}
\begin{tabular}{l l l c c c c c c c c c c c c c c c c c c c c }
\toprule
\textbf{Dataset} & \textbf{Target} & \textbf{Covariates} & \multicolumn{5}{c}{\textbf{Chronos-2}} & \multicolumn{5}{c}{\textbf{TabPFN}} & \multicolumn{5}{c}{\textbf{TimesFM}} & \multicolumn{5}{c}{\textbf{TiRex-2}} \\
\cmidrule(lr){4-8} \cmidrule(lr){9-13} \cmidrule(lr){14-18} \cmidrule(lr){19-23}
& & & Min & Q25 & Med & Q75 & Max & Min & Q25 & Med & Q75 & Max & Min & Q25 & Med & Q75 & Max & Min & Q25 & Med & Q75 & Max \\
\midrule
\multicolumn{23}{l}{\textbf{Balancing Services}} \\
\\[-1.5ex]
Balancing Data Germany & Area Control Error (NRV Saldo) & \textit{DE Weather + Calendar} & \val{1.0335} & \val{1.1938} & \cellcolor{green!20}\textbf{\val{1.2546}} & \val{1.3114} & \val{2.3706} & \val{1.0389} & \val{1.2193} & \val{1.2567} & \val{1.2989} & \val{1.7632} & \val{1.0550} & \val{1.2959} & \val{1.4427} & \val{1.7956} & \val{8.0192} & \val{1.0990} & \val{1.2222} & \val{1.2702} & \val{1.3176} & \val{2.1500} \\
aFRR Germany & Germany aFRR negativ & \textit{DE Weather + Calendar} & \val{1.3545} & \val{1.8730} & \cellcolor{green!20}\textbf{\val{2.2284}} & \val{2.6071} & \val{3.9175} & \val{1.4444} & \val{1.9465} & \val{2.2788} & \val{2.6396} & \val{3.9287} & \val{1.2101} & \val{2.0741} & \val{2.6432} & \val{3.2893} & \val{15.0331} & \val{1.3762} & \val{1.9183} & \val{2.2485} & \val{2.6087} & \val{3.9211} \\
aFRR Germany & Germany aFRR positiv & \textit{DE Weather + Calendar} & \val{1.5171} & \val{2.0122} & \val{2.2122} & \val{2.3561} & \val{3.0062} & \val{1.6566} & \val{2.0589} & \val{2.2568} & \val{2.3961} & \val{3.0661} & \val{1.5009} & \val{2.0850} & \val{2.3835} & \val{3.0773} & \val{8.8459} & \val{1.4922} & \val{2.0013} & \cellcolor{green!20}\textbf{\val{2.2004}} & \val{2.3410} & \val{2.9713} \\
\bottomrule
\end{tabular}
\end{adjustbox}
\label{tab:covariate_benchmark}
\end{sidewaystable}
\clearpage

\begin{sidewaystable}[htbp]
\centering
\caption{Results Training Benchmark (NRMSE): Statistics (Min, Q25, Median, Q75, Max) over multiple forecast horizons and context lengths; best median per row highlighted. Overall row aggregates globally across all datasets, targets, horizons, and context lengths. Values $>\!10^{6}$ are treated as diverged, unusable forecasts and displayed as $>\!10^{6}$. All 68,040 experiments are evaluated.}
\footnotesize
\begin{adjustbox}{max width=\linewidth}
\begin{tabular}{l l l c c c c c c c c c c c c c c c c c c c c }
\toprule
\textbf{Dataset} & \textbf{Target} & \textbf{Covariates} & \multicolumn{5}{c}{\textbf{Chronos-2}} & \multicolumn{5}{c}{\textbf{RandomForest}} & \multicolumn{5}{c}{\textbf{XGBoost}} & \multicolumn{5}{c}{\textbf{XGBoost-MO}} \\
\cmidrule(lr){4-8} \cmidrule(lr){9-13} \cmidrule(lr){14-18} \cmidrule(lr){19-23}
& & & Min & Q25 & Med & Q75 & Max & Min & Q25 & Med & Q75 & Max & Min & Q25 & Med & Q75 & Max & Min & Q25 & Med & Q75 & Max \\
\midrule
\multicolumn{23}{l}{\textbf{Balancing Services}} \\
\\[-1.5ex]
Balancing Data Germany & Area Control Error (NRV Saldo) & \textit{DE Weather + Calendar} & \val{0.9857} & \val{1.1717} & \cellcolor{green!20}\textbf{\val{1.2185}} & \val{1.2627} & \val{1.3635} & \val{1.0557} & \val{1.2036} & \val{1.2541} & \val{1.2984} & \val{1.4084} & \val{1.1406} & \val{1.2284} & \val{1.2552} & \val{1.2807} & \val{1.3877} & \val{1.1215} & \val{1.2323} & \val{1.2600} & \val{1.2924} & \val{1.3917} \\
aFRR Germany & Germany aFRR negativ & \textit{DE Weather + Calendar} & \val{1.3388} & \val{1.8805} & \cellcolor{green!20}\textbf{\val{2.2722}} & \val{2.6232} & \val{3.9414} & \val{1.1999} & \val{2.2903} & \val{3.2709} & \val{4.9951} & \val{14.2135} & \val{1.1019} & \val{2.2574} & \val{3.3557} & \val{5.1545} & \val{16.0201} & \val{1.1845} & \val{1.8099} & \val{2.2821} & \val{3.2212} & \val{8.4744} \\
aFRR Germany & Germany aFRR positiv & \textit{DE Weather + Calendar} & \val{1.5247} & \val{2.0247} & \val{2.2240} & \val{2.3782} & \val{2.9627} & \val{1.3734} & \val{1.9884} & \val{2.2579} & \val{2.6630} & \val{7.3153} & \val{1.4335} & \val{2.0452} & \val{2.3309} & \val{2.9255} & \val{11.2569} & \val{1.4403} & \val{1.8460} & \cellcolor{green!20}\textbf{\val{2.0758}} & \val{2.3034} & \val{4.3949} \\
\bottomrule
\end{tabular}
\end{adjustbox}
\label{tab:training_benchmark}
\end{sidewaystable}

\begin{table}[htbp]
\centering
\caption{Overall results for MASE; layout and experimental setup identical to Table~\ref{tab:combined_benchmark_reduced}.}
\small
\renewcommand{\arraystretch}{1.5}
\begin{adjustbox}{max width=\textwidth}
%
\par\vspace{4pt}
\begin{minipage}{\textwidth}\footnotesize\emph{Note:} For Best Count, All = Total - Ties - NaN exp.\ = 17,010 - 20 - 492 = 16,498.\end{minipage}
\label{tab:combined_benchmark_mase}
\end{table}
\begin{table}[htbp]
\centering
\caption{Overall results for Normalized CRPS; layout and experimental setup identical to Table~\ref{tab:combined_benchmark_reduced}.}
\small
\renewcommand{\arraystretch}{1.5}
\begin{adjustbox}{max width=\textwidth}
%
\par\vspace{4pt}
\begin{minipage}{\textwidth}\footnotesize\emph{Note:} For Best Count, All = Total - Ties - NaN exp.\ = 17,010 - 20 - 503 = 16,487.\end{minipage}
\label{tab:combined_benchmark_ncrps}
\end{table}
\begin{table}[htbp]
\centering
\caption{Overall results for CORR; higher is better; layout and experimental setup identical to Table~\ref{tab:combined_benchmark_reduced}.}
\small
\renewcommand{\arraystretch}{1.5}
\begin{adjustbox}{max width=\textwidth}
%
\par\vspace{4pt}
\begin{minipage}{\textwidth}\footnotesize\emph{Note:} For Best Count, All = Total - Ties - NaN exp.\ = 17,010 - 0 - 1,123 = 15,887.\end{minipage}
\label{tab:combined_benchmark_corr}
\end{table}
\begin{table}[htbp]
\centering
\caption{Overall results for SUM RATIO (\%); closest to 0 is best; layout and experimental setup identical to Table~\ref{tab:combined_benchmark_reduced}.}
\small
\renewcommand{\arraystretch}{1.5}
\begin{adjustbox}{max width=\textwidth}
%
\par\vspace{4pt}
\begin{minipage}{\textwidth}\footnotesize\emph{Note:} For Best Count, All = Total - Ties - NaN exp.\ = 17,010 - 0 - 378 = 16,632.\end{minipage}
\label{tab:combined_benchmark_sum_ratio}
\end{table}
\begin{table}[htbp]
\centering
\caption{Overall results for PICP $\Delta$ (\%); closest to 0 is best; layout and experimental setup identical to Table~\ref{tab:combined_benchmark_reduced}.}
\small
\renewcommand{\arraystretch}{1.5}
\begin{adjustbox}{max width=\textwidth}
%
\par\vspace{4pt}
\begin{minipage}{\textwidth}\footnotesize\emph{Note:} $\Delta = (\mathrm{PICP} - 0.80)\cdot 100$ denotes the deviation of the empirical 80\% prediction interval coverage from its nominal level in percentage points: $\Delta < 0$ intervals too narrow (overconfident), $\Delta > 0$ too wide; closest to 0 is best. For Best Count, All = Total - Ties = 17,010 - 1,631 = 15,379.\end{minipage}
\label{tab:combined_benchmark_picp}
\end{table}
\begin{table}[htbp]
\centering
\caption{Results overview by experimental setting (NRMSE), with models as rows (grouped by benchmark mode) and the nine combinations of forecast horizon and context length (Ctx) as columns. Cells show the median NRMSE with (Q10, Q90) interval over each model's pooled set of all experiments with that setting (datasets, targets, and rolling windows); layout, color coding, and experimental setup otherwise identical to Table~\ref{tab:combined_benchmark_reduced}.}
\small
\renewcommand{\arraystretch}{1.5}
\begin{adjustbox}{max width=\textwidth}
\begin{tabular}{>{\raggedright\arraybackslash}m{2.6cm} c c c c c c c c c }
\toprule
\textbf{Model} & \multicolumn{3}{c}{\textbf{Horizon 96}} & \multicolumn{3}{c}{\textbf{Horizon 192}} & \multicolumn{3}{c}{\textbf{Horizon 288}} \\
\cmidrule(lr){2-4}\cmidrule(lr){5-7}\cmidrule(lr){8-10}
 & \textbf{Ctx 672} & \textbf{Ctx 2{,}000} & \textbf{Ctx 8{,}000} & \textbf{Ctx 672} & \textbf{Ctx 2{,}000} & \textbf{Ctx 8{,}000} & \textbf{Ctx 672} & \textbf{Ctx 2{,}000} & \textbf{Ctx 8{,}000} \\
\midrule
\multicolumn{10}{l}{\textit{Univariate}} \\
Toto-2.0 & $\substack{\rule{0pt}{10pt}\textstyle 0.544 \\[1pt] \scriptstyle\langle 0.099,\,1.75\rangle\rule[-5pt]{0pt}{5pt}}$ & \cellcolor{orange!10}$\substack{\rule{0pt}{10pt}\textstyle\mathbf{0.478} \\[1pt] \scriptstyle\langle 0.081,\,1.76\rangle\rule[-5pt]{0pt}{5pt}}$ & $\substack{\rule{0pt}{10pt}\textstyle 0.487 \\[1pt] \scriptstyle\langle 0.081,\,1.77\rangle\rule[-5pt]{0pt}{5pt}}$ & $\substack{\rule{0pt}{10pt}\textstyle 0.614 \\[1pt] \scriptstyle\langle 0.133,\,1.89\rangle\rule[-5pt]{0pt}{5pt}}$ & \cellcolor{orange!10}$\substack{\rule{0pt}{10pt}\textstyle\mathbf{0.542} \\[1pt] \scriptstyle\langle 0.106,\,1.75\rangle\rule[-5pt]{0pt}{5pt}}$ & \cellcolor{orange!10}$\substack{\rule{0pt}{10pt}\textstyle\mathbf{0.549} \\[1pt] \scriptstyle\langle 0.111,\,1.77\rangle\rule[-5pt]{0pt}{5pt}}$ & $\substack{\rule{0pt}{10pt}\textstyle 0.671 \\[1pt] \scriptstyle\langle 0.153,\,1.84\rangle\rule[-5pt]{0pt}{5pt}}$ & \cellcolor{orange!10}$\substack{\rule{0pt}{10pt}\textstyle\mathbf{0.589} \\[1pt] \scriptstyle\langle 0.119,\,1.69\rangle\rule[-5pt]{0pt}{5pt}}$ & $\substack{\rule{0pt}{10pt}\textstyle 0.611 \\[1pt] \scriptstyle\langle 0.123,\,1.73\rangle\rule[-5pt]{0pt}{5pt}}$ \\
TimesFM & $\substack{\rule{0pt}{10pt}\textstyle 0.562 \\[1pt] \scriptstyle\langle 0.099,\,1.86\rangle\rule[-5pt]{0pt}{5pt}}$ & $\substack{\rule{0pt}{10pt}\textstyle 0.532 \\[1pt] \scriptstyle\langle 0.085,\,1.84\rangle\rule[-5pt]{0pt}{5pt}}$ & $\substack{\rule{0pt}{10pt}\textstyle 0.513 \\[1pt] \scriptstyle\langle 0.086,\,1.72\rangle\rule[-5pt]{0pt}{5pt}}$ & $\substack{\rule{0pt}{10pt}\textstyle 0.618 \\[1pt] \scriptstyle\langle 0.119,\,1.88\rangle\rule[-5pt]{0pt}{5pt}}$ & $\substack{\rule{0pt}{10pt}\textstyle 0.594 \\[1pt] \scriptstyle\langle 0.113,\,1.76\rangle\rule[-5pt]{0pt}{5pt}}$ & $\substack{\rule{0pt}{10pt}\textstyle 0.560 \\[1pt] \scriptstyle\langle 0.107,\,1.70\rangle\rule[-5pt]{0pt}{5pt}}$ & $\substack{\rule{0pt}{10pt}\textstyle 0.655 \\[1pt] \scriptstyle\langle 0.131,\,1.86\rangle\rule[-5pt]{0pt}{5pt}}$ & $\substack{\rule{0pt}{10pt}\textstyle 0.636 \\[1pt] \scriptstyle\langle 0.124,\,1.77\rangle\rule[-5pt]{0pt}{5pt}}$ & \cellcolor{orange!10}$\substack{\rule{0pt}{10pt}\textstyle\mathbf{0.611} \\[1pt] \scriptstyle\langle 0.120,\,1.71\rangle\rule[-5pt]{0pt}{5pt}}$ \\
Chronos-2 & \cellcolor{orange!10}$\substack{\rule{0pt}{10pt}\textstyle\mathbf{0.540} \\[1pt] \scriptstyle\langle 0.095,\,1.86\rangle\rule[-5pt]{0pt}{5pt}}$ & $\substack{\rule{0pt}{10pt}\textstyle 0.493 \\[1pt] \scriptstyle\langle 0.080,\,1.76\rangle\rule[-5pt]{0pt}{5pt}}$ & \cellcolor{orange!10}$\substack{\rule{0pt}{10pt}\textstyle\mathbf{0.487} \\[1pt] \scriptstyle\langle 0.080,\,1.68\rangle\rule[-5pt]{0pt}{5pt}}$ & \cellcolor{orange!10}$\substack{\rule{0pt}{10pt}\textstyle\mathbf{0.598} \\[1pt] \scriptstyle\langle 0.121,\,1.86\rangle\rule[-5pt]{0pt}{5pt}}$ & $\substack{\rule{0pt}{10pt}\textstyle 0.562 \\[1pt] \scriptstyle\langle 0.107,\,1.75\rangle\rule[-5pt]{0pt}{5pt}}$ & $\substack{\rule{0pt}{10pt}\textstyle 0.569 \\[1pt] \scriptstyle\langle 0.108,\,1.68\rangle\rule[-5pt]{0pt}{5pt}}$ & \cellcolor{orange!10}$\substack{\rule{0pt}{10pt}\textstyle\mathbf{0.652} \\[1pt] \scriptstyle\langle 0.133,\,1.86\rangle\rule[-5pt]{0pt}{5pt}}$ & $\substack{\rule{0pt}{10pt}\textstyle 0.609 \\[1pt] \scriptstyle\langle 0.117,\,1.78\rangle\rule[-5pt]{0pt}{5pt}}$ & $\substack{\rule{0pt}{10pt}\textstyle 0.615 \\[1pt] \scriptstyle\langle 0.115,\,1.71\rangle\rule[-5pt]{0pt}{5pt}}$ \\
FlowState & $\substack{\rule{0pt}{10pt}\textstyle 0.602 \\[1pt] \scriptstyle\langle 0.101,\,1.97\rangle\rule[-5pt]{0pt}{5pt}}$ & $\substack{\rule{0pt}{10pt}\textstyle 0.553 \\[1pt] \scriptstyle\langle 0.096,\,1.85\rangle\rule[-5pt]{0pt}{5pt}}$ & $\substack{\rule{0pt}{10pt}\textstyle 0.546 \\[1pt] \scriptstyle\langle 0.094,\,1.87\rangle\rule[-5pt]{0pt}{5pt}}$ & $\substack{\rule{0pt}{10pt}\textstyle 0.646 \\[1pt] \scriptstyle\langle 0.132,\,1.93\rangle\rule[-5pt]{0pt}{5pt}}$ & $\substack{\rule{0pt}{10pt}\textstyle 0.609 \\[1pt] \scriptstyle\langle 0.121,\,1.75\rangle\rule[-5pt]{0pt}{5pt}}$ & $\substack{\rule{0pt}{10pt}\textstyle 0.597 \\[1pt] \scriptstyle\langle 0.120,\,1.80\rangle\rule[-5pt]{0pt}{5pt}}$ & $\substack{\rule{0pt}{10pt}\textstyle 0.688 \\[1pt] \scriptstyle\langle 0.140,\,1.82\rangle\rule[-5pt]{0pt}{5pt}}$ & $\substack{\rule{0pt}{10pt}\textstyle 0.649 \\[1pt] \scriptstyle\langle 0.132,\,1.69\rangle\rule[-5pt]{0pt}{5pt}}$ & $\substack{\rule{0pt}{10pt}\textstyle 0.640 \\[1pt] \scriptstyle\langle 0.129,\,1.68\rangle\rule[-5pt]{0pt}{5pt}}$ \\
TiRex-2 & $\substack{\rule{0pt}{10pt}\textstyle 0.558 \\[1pt] \scriptstyle\langle 0.099,\,1.84\rangle\rule[-5pt]{0pt}{5pt}}$ & $\substack{\rule{0pt}{10pt}\textstyle 0.518 \\[1pt] \scriptstyle\langle 0.084,\,1.76\rangle\rule[-5pt]{0pt}{5pt}}$ & $\substack{\rule{0pt}{10pt}\textstyle 0.514 \\[1pt] \scriptstyle\langle 0.086,\,1.76\rangle\rule[-5pt]{0pt}{5pt}}$ & $\substack{\rule{0pt}{10pt}\textstyle 0.608 \\[1pt] \scriptstyle\langle 0.125,\,1.84\rangle\rule[-5pt]{0pt}{5pt}}$ & $\substack{\rule{0pt}{10pt}\textstyle 0.580 \\[1pt] \scriptstyle\langle 0.115,\,1.78\rangle\rule[-5pt]{0pt}{5pt}}$ & $\substack{\rule{0pt}{10pt}\textstyle 0.579 \\[1pt] \scriptstyle\langle 0.116,\,1.76\rangle\rule[-5pt]{0pt}{5pt}}$ & $\substack{\rule{0pt}{10pt}\textstyle 0.657 \\[1pt] \scriptstyle\langle 0.140,\,1.79\rangle\rule[-5pt]{0pt}{5pt}}$ & $\substack{\rule{0pt}{10pt}\textstyle 0.623 \\[1pt] \scriptstyle\langle 0.125,\,1.71\rangle\rule[-5pt]{0pt}{5pt}}$ & $\substack{\rule{0pt}{10pt}\textstyle 0.627 \\[1pt] \scriptstyle\langle 0.126,\,1.71\rangle\rule[-5pt]{0pt}{5pt}}$ \\
\midrule
\multicolumn{10}{l}{\textit{Covariate}} \\
Chronos-2 & $\substack{\rule{0pt}{10pt}\textstyle 0.461 \\[1pt] \scriptstyle\langle 0.092,\,1.74\rangle\rule[-5pt]{0pt}{5pt}}$ & \cellcolor{green!15}$\substack{\rule{0pt}{10pt}\textstyle\mathbf{0.416} \\[1pt] \scriptstyle\langle 0.077,\,1.61\rangle\rule[-5pt]{0pt}{5pt}}$ & $\substack{\rule{0pt}{10pt}\textstyle 0.421 \\[1pt] \scriptstyle\langle 0.078,\,1.60\rangle\rule[-5pt]{0pt}{5pt}}$ & $\substack{\rule{0pt}{10pt}\textstyle 0.520 \\[1pt] \scriptstyle\langle 0.118,\,1.78\rangle\rule[-5pt]{0pt}{5pt}}$ & \cellcolor{green!15}$\substack{\rule{0pt}{10pt}\textstyle\mathbf{0.453} \\[1pt] \scriptstyle\langle 0.099,\,1.64\rangle\rule[-5pt]{0pt}{5pt}}$ & \cellcolor{green!15}$\substack{\rule{0pt}{10pt}\textstyle\mathbf{0.467} \\[1pt] \scriptstyle\langle 0.101,\,1.64\rangle\rule[-5pt]{0pt}{5pt}}$ & $\substack{\rule{0pt}{10pt}\textstyle 0.586 \\[1pt] \scriptstyle\langle 0.135,\,1.82\rangle\rule[-5pt]{0pt}{5pt}}$ & \cellcolor{green!15}$\substack{\rule{0pt}{10pt}\textstyle\mathbf{0.483} \\[1pt] \scriptstyle\langle 0.114,\,1.63\rangle\rule[-5pt]{0pt}{5pt}}$ & \cellcolor{green!15}$\substack{\rule{0pt}{10pt}\textstyle\mathbf{0.491} \\[1pt] \scriptstyle\langle 0.112,\,1.61\rangle\rule[-5pt]{0pt}{5pt}}$ \\
TabPFN & $\substack{\rule{0pt}{10pt}\textstyle 0.486 \\[1pt] \scriptstyle\langle 0.116,\,1.74\rangle\rule[-5pt]{0pt}{5pt}}$ & $\substack{\rule{0pt}{10pt}\textstyle 0.448 \\[1pt] \scriptstyle\langle 0.105,\,1.79\rangle\rule[-5pt]{0pt}{5pt}}$ & $\substack{\rule{0pt}{10pt}\textstyle 0.432 \\[1pt] \scriptstyle\langle 0.112,\,1.76\rangle\rule[-5pt]{0pt}{5pt}}$ & $\substack{\rule{0pt}{10pt}\textstyle 0.527 \\[1pt] \scriptstyle\langle 0.134,\,1.71\rangle\rule[-5pt]{0pt}{5pt}}$ & $\substack{\rule{0pt}{10pt}\textstyle 0.479 \\[1pt] \scriptstyle\langle 0.123,\,1.66\rangle\rule[-5pt]{0pt}{5pt}}$ & $\substack{\rule{0pt}{10pt}\textstyle 0.469 \\[1pt] \scriptstyle\langle 0.123,\,1.64\rangle\rule[-5pt]{0pt}{5pt}}$ & $\substack{\rule{0pt}{10pt}\textstyle 0.568 \\[1pt] \scriptstyle\langle 0.146,\,1.62\rangle\rule[-5pt]{0pt}{5pt}}$ & $\substack{\rule{0pt}{10pt}\textstyle 0.526 \\[1pt] \scriptstyle\langle 0.130,\,1.57\rangle\rule[-5pt]{0pt}{5pt}}$ & $\substack{\rule{0pt}{10pt}\textstyle 0.504 \\[1pt] \scriptstyle\langle 0.133,\,1.58\rangle\rule[-5pt]{0pt}{5pt}}$ \\
TimesFM & $\substack{\rule{0pt}{10pt}\textstyle 0.645 \\[1pt] \scriptstyle\langle 0.171,\,2.89\rangle\rule[-5pt]{0pt}{5pt}}$ & $\substack{\rule{0pt}{10pt}\textstyle 0.503 \\[1pt] \scriptstyle\langle 0.127,\,2.10\rangle\rule[-5pt]{0pt}{5pt}}$ & $\substack{\rule{0pt}{10pt}\textstyle 0.453 \\[1pt] \scriptstyle\langle 0.111,\,1.85\rangle\rule[-5pt]{0pt}{5pt}}$ & $\substack{\rule{0pt}{10pt}\textstyle 0.766 \\[1pt] \scriptstyle\langle 0.219,\,3.17\rangle\rule[-5pt]{0pt}{5pt}}$ & $\substack{\rule{0pt}{10pt}\textstyle 0.537 \\[1pt] \scriptstyle\langle 0.154,\,1.92\rangle\rule[-5pt]{0pt}{5pt}}$ & $\substack{\rule{0pt}{10pt}\textstyle 0.480 \\[1pt] \scriptstyle\langle 0.128,\,1.70\rangle\rule[-5pt]{0pt}{5pt}}$ & $\substack{\rule{0pt}{10pt}\textstyle 0.875 \\[1pt] \scriptstyle\langle 0.234,\,3.63\rangle\rule[-5pt]{0pt}{5pt}}$ & $\substack{\rule{0pt}{10pt}\textstyle 0.601 \\[1pt] \scriptstyle\langle 0.164,\,1.88\rangle\rule[-5pt]{0pt}{5pt}}$ & $\substack{\rule{0pt}{10pt}\textstyle 0.529 \\[1pt] \scriptstyle\langle 0.139,\,1.63\rangle\rule[-5pt]{0pt}{5pt}}$ \\
TiRex-2 & \cellcolor{green!15}$\substack{\rule{0pt}{10pt}\textstyle\mathbf{0.460} \\[1pt] \scriptstyle\langle 0.099,\,1.69\rangle\rule[-5pt]{0pt}{5pt}}$ & $\substack{\rule{0pt}{10pt}\textstyle 0.421 \\[1pt] \scriptstyle\langle 0.086,\,1.62\rangle\rule[-5pt]{0pt}{5pt}}$ & \cellcolor{green!15}$\substack{\rule{0pt}{10pt}\textstyle\mathbf{0.419} \\[1pt] \scriptstyle\langle 0.086,\,1.58\rangle\rule[-5pt]{0pt}{5pt}}$ & \cellcolor{green!15}$\substack{\rule{0pt}{10pt}\textstyle\mathbf{0.517} \\[1pt] \scriptstyle\langle 0.122,\,1.74\rangle\rule[-5pt]{0pt}{5pt}}$ & $\substack{\rule{0pt}{10pt}\textstyle 0.473 \\[1pt] \scriptstyle\langle 0.109,\,1.63\rangle\rule[-5pt]{0pt}{5pt}}$ & $\substack{\rule{0pt}{10pt}\textstyle 0.469 \\[1pt] \scriptstyle\langle 0.110,\,1.61\rangle\rule[-5pt]{0pt}{5pt}}$ & \cellcolor{green!15}$\substack{\rule{0pt}{10pt}\textstyle\mathbf{0.553} \\[1pt] \scriptstyle\langle 0.143,\,1.80\rangle\rule[-5pt]{0pt}{5pt}}$ & $\substack{\rule{0pt}{10pt}\textstyle 0.499 \\[1pt] \scriptstyle\langle 0.124,\,1.67\rangle\rule[-5pt]{0pt}{5pt}}$ & $\substack{\rule{0pt}{10pt}\textstyle 0.495 \\[1pt] \scriptstyle\langle 0.124,\,1.67\rangle\rule[-5pt]{0pt}{5pt}}$ \\
\midrule
\multicolumn{10}{l}{\textit{Training / Fine-tuning}} \\
Chronos-2 & \cellcolor{orange!10}$\substack{\rule{0pt}{10pt}\textstyle\mathbf{0.526} \\[1pt] \scriptstyle\langle 0.102,\,2.09\rangle\rule[-5pt]{0pt}{5pt}}$ & $\substack{\rule{0pt}{10pt}\textstyle 0.524 \\[1pt] \scriptstyle\langle 0.105,\,2.07\rangle\rule[-5pt]{0pt}{5pt}}$ & \cellcolor{orange!10}$\substack{\rule{0pt}{10pt}\textstyle\mathbf{0.531} \\[1pt] \scriptstyle\langle 0.105,\,2.09\rangle\rule[-5pt]{0pt}{5pt}}$ & $\substack{\rule{0pt}{10pt}\textstyle 0.640 \\[1pt] \scriptstyle\langle 0.166,\,2.06\rangle\rule[-5pt]{0pt}{5pt}}$ & $\substack{\rule{0pt}{10pt}\textstyle 0.641 \\[1pt] \scriptstyle\langle 0.163,\,2.09\rangle\rule[-5pt]{0pt}{5pt}}$ & \cellcolor{orange!10}$\substack{\rule{0pt}{10pt}\textstyle\mathbf{0.633} \\[1pt] \scriptstyle\langle 0.167,\,2.08\rangle\rule[-5pt]{0pt}{5pt}}$ & \cellcolor{orange!10}$\substack{\rule{0pt}{10pt}\textstyle\mathbf{0.611} \\[1pt] \scriptstyle\langle 0.158,\,1.81\rangle\rule[-5pt]{0pt}{5pt}}$ & \cellcolor{orange!10}$\substack{\rule{0pt}{10pt}\textstyle\mathbf{0.612} \\[1pt] \scriptstyle\langle 0.156,\,1.80\rangle\rule[-5pt]{0pt}{5pt}}$ & \cellcolor{orange!10}$\substack{\rule{0pt}{10pt}\textstyle\mathbf{0.611} \\[1pt] \scriptstyle\langle 0.157,\,1.81\rangle\rule[-5pt]{0pt}{5pt}}$ \\
RandomForest & $\substack{\rule{0pt}{10pt}\textstyle 0.604 \\[1pt] \scriptstyle\langle 0.141,\,2.28\rangle\rule[-5pt]{0pt}{5pt}}$ & $\substack{\rule{0pt}{10pt}\textstyle 0.625 \\[1pt] \scriptstyle\langle 0.140,\,2.03\rangle\rule[-5pt]{0pt}{5pt}}$ & $\substack{\rule{0pt}{10pt}\textstyle 0.635 \\[1pt] \scriptstyle\langle 0.138,\,1.95\rangle\rule[-5pt]{0pt}{5pt}}$ & $\substack{\rule{0pt}{10pt}\textstyle 0.720 \\[1pt] \scriptstyle\langle 0.170,\,2.26\rangle\rule[-5pt]{0pt}{5pt}}$ & $\substack{\rule{0pt}{10pt}\textstyle 0.675 \\[1pt] \scriptstyle\langle 0.168,\,1.88\rangle\rule[-5pt]{0pt}{5pt}}$ & $\substack{\rule{0pt}{10pt}\textstyle 0.719 \\[1pt] \scriptstyle\langle 0.178,\,2.12\rangle\rule[-5pt]{0pt}{5pt}}$ & $\substack{\rule{0pt}{10pt}\textstyle 0.743 \\[1pt] \scriptstyle\langle 0.199,\,2.27\rangle\rule[-5pt]{0pt}{5pt}}$ & $\substack{\rule{0pt}{10pt}\textstyle 0.729 \\[1pt] \scriptstyle\langle 0.168,\,2.02\rangle\rule[-5pt]{0pt}{5pt}}$ & $\substack{\rule{0pt}{10pt}\textstyle 0.766 \\[1pt] \scriptstyle\langle 0.209,\,1.99\rangle\rule[-5pt]{0pt}{5pt}}$ \\
XGBoost & $\substack{\rule{0pt}{10pt}\textstyle 0.530 \\[1pt] \scriptstyle\langle 0.110,\,1.91\rangle\rule[-5pt]{0pt}{5pt}}$ & \cellcolor{orange!10}$\substack{\rule{0pt}{10pt}\textstyle\mathbf{0.522} \\[1pt] \scriptstyle\langle 0.111,\,1.92\rangle\rule[-5pt]{0pt}{5pt}}$ & $\substack{\rule{0pt}{10pt}\textstyle 0.581 \\[1pt] \scriptstyle\langle 0.134,\,2.05\rangle\rule[-5pt]{0pt}{5pt}}$ & \cellcolor{orange!10}$\substack{\rule{0pt}{10pt}\textstyle\mathbf{0.604} \\[1pt] \scriptstyle\langle 0.147,\,2.55\rangle\rule[-5pt]{0pt}{5pt}}$ & \cellcolor{orange!10}$\substack{\rule{0pt}{10pt}\textstyle\mathbf{0.574} \\[1pt] \scriptstyle\langle 0.132,\,2.20\rangle\rule[-5pt]{0pt}{5pt}}$ & $\substack{\rule{0pt}{10pt}\textstyle 0.701 \\[1pt] \scriptstyle\langle 0.187,\,2.68\rangle\rule[-5pt]{0pt}{5pt}}$ & $\substack{\rule{0pt}{10pt}\textstyle 0.661 \\[1pt] \scriptstyle\langle 0.150,\,2.43\rangle\rule[-5pt]{0pt}{5pt}}$ & $\substack{\rule{0pt}{10pt}\textstyle 0.620 \\[1pt] \scriptstyle\langle 0.139,\,2.12\rangle\rule[-5pt]{0pt}{5pt}}$ & $\substack{\rule{0pt}{10pt}\textstyle 0.685 \\[1pt] \scriptstyle\langle 0.187,\,1.84\rangle\rule[-5pt]{0pt}{5pt}}$ \\
XGBoost-MO & $\substack{\rule{0pt}{10pt}\textstyle 0.557 \\[1pt] \scriptstyle\langle 0.120,\,1.88\rangle\rule[-5pt]{0pt}{5pt}}$ & $\substack{\rule{0pt}{10pt}\textstyle 0.568 \\[1pt] \scriptstyle\langle 0.133,\,1.89\rangle\rule[-5pt]{0pt}{5pt}}$ & $\substack{\rule{0pt}{10pt}\textstyle 0.695 \\[1pt] \scriptstyle\langle 0.166,\,1.83\rangle\rule[-5pt]{0pt}{5pt}}$ & $\substack{\rule{0pt}{10pt}\textstyle 0.613 \\[1pt] \scriptstyle\langle 0.152,\,1.80\rangle\rule[-5pt]{0pt}{5pt}}$ & $\substack{\rule{0pt}{10pt}\textstyle 0.614 \\[1pt] \scriptstyle\langle 0.150,\,1.87\rangle\rule[-5pt]{0pt}{5pt}}$ & $\substack{\rule{0pt}{10pt}\textstyle 0.731 \\[1pt] \scriptstyle\langle 0.199,\,1.87\rangle\rule[-5pt]{0pt}{5pt}}$ & $\substack{\rule{0pt}{10pt}\textstyle 0.640 \\[1pt] \scriptstyle\langle 0.148,\,1.75\rangle\rule[-5pt]{0pt}{5pt}}$ & $\substack{\rule{0pt}{10pt}\textstyle 0.649 \\[1pt] \scriptstyle\langle 0.156,\,1.82\rangle\rule[-5pt]{0pt}{5pt}}$ & $\substack{\rule{0pt}{10pt}\textstyle 0.752 \\[1pt] \scriptstyle\langle 0.204,\,1.76\rangle\rule[-5pt]{0pt}{5pt}}$ \\
\bottomrule
\end{tabular}
\end{adjustbox}
\vspace{4pt}
\noindent\begin{tikzpicture}
\node[inner sep=3pt, minimum width=1.2em, minimum height=1.2em, fill=orange!10, draw=gray!50] at (0.65\textwidth,0) (A) {};
\node[right=4pt of A, anchor=west] {\small Mode Winner};
\node[inner sep=3pt, minimum width=1.2em, minimum height=1.2em, fill=green!15, draw=gray!50] at (0.9\textwidth,0) (B) {};
\node[right=4pt of B, anchor=west] {\small Setting Winner};
\end{tikzpicture}
\label{tab:combined_benchmark_settings}
\end{table}
\begin{table}[t]
\centering
\caption{Model ranking by median NRMSE (left block, point-forecast accuracy) and, recomputed independently, by median normalized CRPS (right block, probabilistic accuracy) across $N=54$ and $N=54$ (dataset, target) series, respectively. Friedman tests confirm significant rank differences for both metrics (NRMSE: $\chi^2=176.4$, $p=2.3\times10^{-31}$, CD$=2.48$; nCRPS: $\chi^2=361.2$, $p=6.1\times10^{-70}$, CD$=2.48$; Nemenyi post-hoc, $\alpha=0.05$). A homogeneous group is a maximal set of models that are pairwise \emph{not} significantly different ($|\Delta\bar R|<\mathrm{CD}$); since this relation is not transitive, groups may overlap, and two models differ significantly iff their group sets are disjoint. Group letters are defined \emph{within each block only} and are not comparable across the two rankings. A lower average rank is better.}
\label{tab:significance_nrmse}
\begin{adjustbox}{max width=\textwidth}
\begin{tabular}{rlcc @{\hspace{2.5em}} rlcc}
\toprule
\multicolumn{4}{c}{\emph{Point-forecast ranking (median NRMSE)}} & \multicolumn{4}{c}{\emph{Probabilistic ranking (median nCRPS)}} \\
\cmidrule(lr){1-4} \cmidrule(lr){5-8}
\# & Model configuration & Avg.\ rank & Group & \# & Model configuration & Avg.\ rank & Group \\
\midrule
1 & Chronos-2 (covariate) & 3.63 & $\{a\}$ & 1 & Chronos-2 (covariate) & 3.56 & $\{a\}$ \\
2 & Chronos-2 (univariate) & 5.19 & $\{a,b\}$ & 2 & Chronos-2 (univariate) & 4.44 & $\{a,b\}$ \\
3 & TiRex-2 (covariate) & 5.47 & $\{a,b\}$ & 3 & TiRex-2 (covariate) & 4.65 & $\{a,b\}$ \\
4 & FlowState (univariate) & 6.02 & $\{a,b\}$ & 4 & Toto-2.0 (univariate) & 4.83 & $\{a,b\}$ \\
5 & TabPFN (covariate) & 6.15 & $\{b\}$ & 5 & TabPFN (covariate) & 5.17 & $\{a,b\}$ \\
6 & TimesFM (univariate) & 6.15 & $\{b\}$ & 6 & TiRex-2 (univariate) & 6.00 & $\{a,b\}$ \\
7 & Toto-2.0 (univariate) & 6.30 & $\{b\}$ & 7 & Chronos-2 (fine-tuned) & 6.11 & $\{b\}$ \\
8 & TiRex-2 (univariate) & 6.82 & $\{b,c\}$ & 8 & FlowState (univariate) & 6.28 & $\{b\}$ \\
9 & Chronos-2 (fine-tuned) & 6.97 & $\{b,c\}$ & 9 & TimesFM (univariate) & 6.31 & $\{b\}$ \\
10 & XGBoost (multi-output) & 8.82 & $\{c,d\}$ & 10 & TimesFM (covariate) & 9.43 & $\{c\}$ \\
11 & TimesFM (covariate) & 9.07 & $\{c,d\}$ & 11 & XGBoost & 10.17 & $\{c,d\}$ \\
12 & XGBoost & 9.56 & $\{d\}$ & 12 & XGBoost (multi-output) & 11.70 & $\{c,d\}$ \\
13 & RandomForest & 10.84 & $\{d\}$ & 13 & RandomForest & 12.35 & $\{d\}$ \\
\bottomrule
\end{tabular}
\end{adjustbox}
\end{table}

\clearpage
\section{Extended Sensitivity Analyses}

\definecolor{c2blue}{HTML}{3A6DA0}
\definecolor{t2amber}{HTML}{E07B00}
\begin{table}[t]
\centering
\caption{Weather-input sensitivity: median NRMSE per dataset under four weather scenarios A--D (defined below the table). Each cell reports \textcolor{c2blue}{\textbf{Chronos-2}}\,/\,\textcolor{t2amber}{\textbf{TiRex-2}}; the lower (better) scenario per model and dataset is in \textbf{bold}.}
\label{tab:weather_sensitivity}
\footnotesize\setlength{\tabcolsep}{8pt}
\begin{tabular}{l cccc}
\toprule
Dataset & \textbf{A} & \textbf{B} & \textbf{C} & \textbf{D} \\
\midrule
\multicolumn{5}{l}{\textit{Non-Dispatchable Generation}} \\
\quad Solar (DE) & \textcolor{c2blue}{0.227}\,/\,\textcolor{t2amber}{0.209} & \textcolor{c2blue}{0.264}\,/\,\textcolor{t2amber}{0.239} & \textcolor{c2blue}{0.245}\,/\,\textcolor{t2amber}{0.208} & \textcolor{c2blue}{\textbf{0.195}}\,/\,\textcolor{t2amber}{\textbf{0.202}} \\
\quad Wind onshore (DE) & \textcolor{c2blue}{\textbf{0.255}}\,/\,\textcolor{t2amber}{\textbf{0.237}} & \textcolor{c2blue}{0.258}\,/\,\textcolor{t2amber}{0.250} & \textcolor{c2blue}{0.292}\,/\,\textcolor{t2amber}{0.257} & \textcolor{c2blue}{0.261}\,/\,\textcolor{t2amber}{0.266} \\
\quad Wind offshore (DE) & \textcolor{c2blue}{\textbf{0.464}}\,/\,\textcolor{t2amber}{\textbf{0.423}} & \textcolor{c2blue}{0.501}\,/\,\textcolor{t2amber}{0.445} & \textcolor{c2blue}{0.519}\,/\,\textcolor{t2amber}{0.432} & \textcolor{c2blue}{0.530}\,/\,\textcolor{t2amber}{0.512} \\
\addlinespace
\multicolumn{5}{l}{\textit{Dispatchable Generation}} \\
\quad Fossil gas (DE) & \textcolor{c2blue}{0.265}\,/\,\textcolor{t2amber}{0.346} & \textcolor{c2blue}{\textbf{0.263}}\,/\,\textcolor{t2amber}{\textbf{0.339}} & \textcolor{c2blue}{0.291}\,/\,\textcolor{t2amber}{0.341} & \textcolor{c2blue}{0.281}\,/\,\textcolor{t2amber}{0.360} \\
\addlinespace
\multicolumn{5}{l}{\textit{Load}} \\
\quad Load (DE) & \textcolor{c2blue}{0.025}\,/\,\textcolor{t2amber}{0.042} & \textcolor{c2blue}{0.025}\,/\,\textcolor{t2amber}{0.041} & \textcolor{c2blue}{0.025}\,/\,\textcolor{t2amber}{0.049} & \textcolor{c2blue}{\textbf{0.023}}\,/\,\textcolor{t2amber}{\textbf{0.038}} \\
\quad Load (FR) & \textcolor{c2blue}{0.025}\,/\,\textcolor{t2amber}{0.031} & \textcolor{c2blue}{0.026}\,/\,\textcolor{t2amber}{0.032} & \textcolor{c2blue}{0.025}\,/\,\textcolor{t2amber}{0.036} & \textcolor{c2blue}{\textbf{0.023}}\,/\,\textcolor{t2amber}{\textbf{0.030}} \\
\addlinespace
\multicolumn{5}{l}{\textit{Grid Data}} \\
\quad Bayernwerk feed-in & \textcolor{c2blue}{\textbf{0.115}}\,/\,\textcolor{t2amber}{\textbf{0.146}} & \textcolor{c2blue}{0.138}\,/\,\textcolor{t2amber}{0.159} & \textcolor{c2blue}{0.143}\,/\,\textcolor{t2amber}{0.167} & \textcolor{c2blue}{0.125}\,/\,\textcolor{t2amber}{0.157} \\
\quad 50Hertz line L472 & \textcolor{c2blue}{\textbf{0.491}}\,/\,\textcolor{t2amber}{\textbf{0.540}} & \textcolor{c2blue}{0.654}\,/\,\textcolor{t2amber}{0.757} & \textcolor{c2blue}{0.648}\,/\,\textcolor{t2amber}{0.760} & \textcolor{c2blue}{0.645}\,/\,\textcolor{t2amber}{0.753} \\
\quad 50Hertz line Baltic Eagle & \textcolor{c2blue}{\textbf{0.602}}\,/\,\textcolor{t2amber}{\textbf{0.619}} & \textcolor{c2blue}{0.972}\,/\,\textcolor{t2amber}{1.024} & \textcolor{c2blue}{0.963}\,/\,\textcolor{t2amber}{1.033} & \textcolor{c2blue}{0.975}\,/\,\textcolor{t2amber}{1.031} \\
\addlinespace
\multicolumn{5}{l}{\textit{Heat Data}} \\
\quad Flensburg heat & \textcolor{c2blue}{\textbf{0.110}}\,/\,\textcolor{t2amber}{\textbf{0.107}} & \textcolor{c2blue}{0.128}\,/\,\textcolor{t2amber}{0.123} & \textcolor{c2blue}{0.128}\,/\,\textcolor{t2amber}{0.124} & \textcolor{c2blue}{0.124}\,/\,\textcolor{t2amber}{0.124} \\
\bottomrule
\end{tabular}
\vspace{4pt}
\begin{minipage}{0.95\linewidth}
\footnotesize\raggedright
\textbf{Weather scenarios} (data source in context $\le t_0$ $\to$ forecast horizon $> t_0$):\\[1pt]
\textbf{A}~~Historical Forecast $\to$ Historical Forecast \emph{(near-idealized upper bound)}\\
\textbf{B}~~Historical Forecast $\to$ NWP \emph{(operational)}\\
\textbf{C}~~Historical Forecast\,+\,ERA5 $\to$ NWP\\
\textbf{D}~~ERA5 $\to$ NWP\\[4pt]
NWP = ECMWF-IFS run issued $\geq$9\,h before $t_0$ (no leakage); Historical Forecast = best-match Open-Meteo forecast; ERA5 = observed reanalysis.
\end{minipage}
\end{table}

\clearpage

\begin{figure}
  \centering
  \resizebox{\textwidth}{!}{\input{seasonality_per_dataset.pgf}}
  \caption{Rolling 5-week median NRMSE per dataset over the year; background shading marks meteorological seasons, the embedded table lists per-season and annual medians per model.}
  \label{fig:seasonality}
\end{figure}

\clearpage

\section{Preliminary Fine-tuning Ablation Analysis}
\label{sec:ft_ablation}
We fine-tune Chronos-2 using LoRA and compare the resulting models to the covariate-informed zero-shot baseline on the same set of 35 rolling windows. Fine-tuning yields improvements primarily for the most challenging, least predictable time series, while degrading performance on the more easily predictable aggregated load series, for which prediction-interval coverage collapses. Overall, fine-tuning benefits 16 of the 54 series, with the per-series effect exhibiting a moderate correlation with zero-shot difficulty and intrinsic forecastability, but not with dataset length or the number of covariates (Tables~\ref{tab:ft_ablation},~\ref{tab:ft_helps_binned}). A hyperparameter sweep over LoRA rank, learning rate, number of optimization steps, and the set of adapted modules indicates that model performance is far more sensitive to dataset characteristics than to these configuration choices (Table~\ref{tab:ft_hyperparam}). Fine-tuned Chronos-2 attains the highest single-experiment win rate of any model (best count 2{,}995, 18\%; Table~\ref{tab:combined_benchmark_reduced}), yet its aggregate median regresses (overall NRMSE 0.599 versus 0.472 in covariate mode) and it wins only one category outright (Heat Data), collapsing on the easy aggregated loads whose zero-shot forecasts are already near-optimal (e.g.\ DE load interval coverage 0.94 to 0.27). Both training and validation loss continue to decrease monotonically, whereas accuracy deteriorates (Figure~\ref{fig:ft_learning_curves}); since the validation split reflects the fine-tuning regime rather than the evaluation distribution, this does not, by itself, rule out overfitting to the fine-tuning window.

\begin{table}[htbp]\centering\small
\caption{Per-dataset fine-tuning effect (35 rolling windows), sorted by zero-shot difficulty: zero-shot (ZS) and fine-tuned (FT) NRMSE and PICP. Fine-tuning helps only the hardest series and collapses coverage on the easy aggregated loads; 16 of 54 series improve overall.}
\label{tab:ft_ablation}
\begin{tabular}{l r r r r r}
\toprule
Dataset & ZS NRMSE & FT NRMSE & $\Delta$NRMSE\% & ZS PICP & FT PICP \\
\midrule
DE load & 0.027 & 0.160 & -492.4 & 0.94 & 0.27 \\
FR load & 0.027 & 0.085 & -211.6 & 0.94 & 0.45 \\
CAISO load & 0.029 & 0.128 & -336.2 & 0.89 & 0.24 \\
Flensburg heat & 0.128 & 0.150 & -17.9 & 0.73 & 0.84 \\
PV Hong Kong & 0.198 & 0.688 & -248.3 & 0.86 & 0.88 \\
NetzeBW feeder (grid) & 0.535 & 0.483 & +9.9 & 0.82 & 0.95 \\
DK Solar & 0.648 & 0.555 & +14.3 & 0.77 & 0.85 \\
reBAP (balancing) & 0.763 & 0.650 & +14.7 & 0.90 & 0.90 \\
Lower Saxony (household) & 0.788 & 0.716 & +9.2 & 0.80 & 0.86 \\
\bottomrule
\end{tabular}
\end{table}

\begin{table}[t]\centering\small
\caption{Fine-tuning effect over all 54 series, binned by zero-shot difficulty, with mean forecastability $\phi$ per bin. Improvements are concentrated in the hardest, least-structured bin; 16 of 54 improve overall. The per-series effect correlates with zero-shot difficulty (Spearman $r=0.39$) and forecastability ($r=-0.36$, both $p<0.01$), but not with dataset length or covariate count ($|r|<0.1$).}
\label{tab:ft_helps_binned}
\begin{tabular}{l r r r r}
\toprule
Zero-shot NRMSE & \#series & \#FT improves & median $\Delta$NRMSE\% & mean $\phi$ \\
\midrule
$<0.05$ & 3 & 0 & $-336$ & 0.74 \\
$0.05$--$0.2$ & 6 & 0 & $-30$ & 0.69 \\
$0.2$--$0.5$ & 19 & 3 & $-18$ & 0.53 \\
$\geq 0.5$ & 26 & 13 & $-0.2$ & 0.30 \\
\bottomrule
\end{tabular}
\end{table}

\begin{table}[t]\centering\footnotesize\setlength{\tabcolsep}{3.5pt}
\caption{Default-anchored fine-tuning hyperparameter sensitivity on the nine ablation datasets, each at one operating point from the main benchmark (context/horizon and target under the name), 35 rolling windows: median NRMSE change vs.\ the covariate-informed zero-shot Chronos-2 at that point (positive = better). \emph{Default} is the deployed configuration of Table~\ref{tab:model_specifications} (lr~$10^{-6}$, 1000 steps, rank~8, $\alpha=16$, $q,k,v,o$+emb); each remaining column changes a \emph{single} hyperparameter relative to this default. Within a dataset the result is essentially invariant to the configuration -- it stays negative for the easily predicted, aggregated series (loads, PV) and positive only for the hardest, least predictable series -- so whether fine-tuning helps is governed by the dataset, not the configuration. Bottom row: median over the nine datasets (these span the difficulty range and are not a representative sample, hence not comparable to the overall 16/54 figure of Table~\ref{tab:ft_helps_binned}).}
\label{tab:ft_hyperparam}
\resizebox{\textwidth}{!}{%
\begin{tabular}{l r r r r r r r r r r r}
\toprule
 & & \multicolumn{4}{c}{Learning rate} & \multicolumn{2}{c}{Steps} & \multicolumn{2}{c}{Rank} & \multicolumn{2}{c}{Modules} \\
\cmidrule(lr){3-6}\cmidrule(lr){7-8}\cmidrule(lr){9-10}\cmidrule(lr){11-12}
Dataset {\scriptsize(ctx/hor\,\textperiodcentered\,target)} & Default & $10^{-8}$ & $10^{-7}$ & $10^{-5}$ & $10^{-4}$ & 500 & 1500 & 4 & 16 & $q$ & $qkvo$ \\
\midrule
\shortstack[l]{PV Hong Kong\\\scriptsize 2000/192\,\textperiodcentered\,UG Hall4} & -304 & -304 & -303 & -305 & -317 & -318 & -303 & -320 & -320 & -307 & -314 \\
\shortstack[l]{FR load\\\scriptsize 2000/96\,\textperiodcentered\,Load} & -141 & -142 & -142 & -132 & -133 & -124 & -132 & -139 & -138 & -134 & -131 \\
\shortstack[l]{DE load\\\scriptsize 672/96\,\textperiodcentered\,Load} & -74 & -74 & -74 & -70 & -109 & -73 & -70 & -82 & -82 & -66 & -82 \\
\shortstack[l]{CAISO load\\\scriptsize 8000/288\,\textperiodcentered\,load\_mw} & -48 & -48 & -48 & -50 & -41 & -53 & -27 & -30 & -36 & -40 & -36 \\
\shortstack[l]{DK Solar\\\scriptsize 2000/96\,\textperiodcentered\,Solar} & -27 & -27 & -27 & -28 & -18 & -19 & -21 & -21 & -21 & -20 & -28 \\
\shortstack[l]{Flensburg heat\\\scriptsize 2000/96\,\textperiodcentered\,Waermel.} & -17 & -17 & -18 & -17 & -13 & -19 & -18 & -15 & -16 & -17 & -19 \\
\shortstack[l]{NetzeBW feeder\\\scriptsize 8000/192\,\textperiodcentered\,act.\ power} & 8 & 7 & 7 & 10 & 14 & 2 & 10 & 3 & 3 & 3 & 4 \\
\shortstack[l]{reBAP\\\scriptsize 672/96\,\textperiodcentered\,EUR/MWh} & 22 & 22 & 22 & 20 & 19 & 21 & 21 & 22 & 22 & 21 & 21 \\
\shortstack[l]{Lower Saxony\\\scriptsize 672/96\,\textperiodcentered\,SFH15} & 23 & 23 & 23 & 21 & 18 & 16 & 21 & 21 & 21 & 22 & 20 \\
\midrule
\textbf{Median} & \textbf{-27} & \textbf{-27} & \textbf{-27} & \textbf{-28} & \textbf{-18} & \textbf{-19} & \textbf{-21} & \textbf{-21} & \textbf{-21} & \textbf{-20} & \textbf{-28} \\
\bottomrule
\end{tabular}}
\end{table}

\begin{figure}[t]
\centering
\resizebox{\textwidth}{!}{\input{ft_learning_curves.pgf}}
\caption{Chronos-2 fine-tuning training and validation losses (raw and rolling mean) for four datasets at their best hyperparameter configuration. The loss decreases monotonically even where predictive accuracy degrades on the easy, well-predictable series (bottom row); since the validation split shares the fine-tuning regime, the decreasing losses do not reflect accuracy on the evaluation windows.}
\label{fig:ft_learning_curves}
\end{figure}

\clearpage

\clearpage

\end{document}